\documentclass[twoside,11pt]{article}

\usepackage[preprint,nohyperref]{jmlr2e}
\usepackage{amsmath}
\usepackage{xcolor}

\usepackage[T1]{fontenc}
\usepackage[utf8]{inputenc}
\usepackage{amssymb}
\usepackage{array}
\usepackage{booktabs}
\usepackage{graphicx}
\usepackage{float}
\usepackage{tabularx}
\usepackage{hyperref}
\usepackage{rotating}
\usepackage{xurl}

\ShortHeadings{Koopman-Type Latent Data Assimilation}{Zhang et al.}

\hypersetup{
    colorlinks=true,
    citecolor=black,
    filecolor=black,
    linkcolor=black,
    urlcolor=black
}

\begin{document}

\title{When Latent Dynamics Meet Data Assimilation: A Benchmark of
Koopman-Type Representations}

\author{\name Zhiwei Zhang\textsuperscript{1} \email zhiwei.zhang@enpc.fr \\
\name Wenxuan Yuan\textsuperscript{2} \email YUAN0186@e.ntu.edu.sg \\
\name Xiaoyuan Cheng\textsuperscript{3} \email ucesxc4@ucl.ac.uk \\
\name Pouria Behnoudfar\textsuperscript{4} \email behnoudfar@wisc.edu \\
\name Chuanqi Chen\textsuperscript{5} \email chuanqichen@uchicago.edu \\
\name Charlotte Moser\textsuperscript{4} \email crmoser2@wisc.edu \\
\name Yuta Tarumi\textsuperscript{7,9} \email yuta.tarumi@riken.jp \\
\name Jin-Long Wu\textsuperscript{6} \email jinlong.wu@wisc.edu \\
\name Rossella Arcucci\textsuperscript{8} \email r.arcucci@imperial.ac.uk \\
\name Takemasa Miyoshi\textsuperscript{7,9} \email takemasa.miyoshi@riken.jp \\
\name Nan Chen\textsuperscript{4} \email chennan@math.wisc.edu \\
\name Marc Bocquet\textsuperscript{1} \email marc.bocquet@enpc.fr \\
\name Sibo Cheng\textsuperscript{1}\thanks{Corresponding author.} \email sibo.cheng@enpc.fr \\[1ex]
\addr
\textsuperscript{1}CEREA, ENPC, EDF R\&D, Institut Polytechnique de Paris, \^{I}le-de-France, France\\
\textsuperscript{2}College of Computing and Data Science, Nanyang Technological University, Singapore\\
\textsuperscript{3}Dynamic Systems Lab, University College London, London, UK\\
\textsuperscript{4}Department of Mathematics, University of Wisconsin--Madison, Madison, WI, USA\\
\textsuperscript{5}Department of the Geophysical Sciences, University of Chicago, Chicago, IL, USA\\
\textsuperscript{6}Department of Mechanical Engineering, University of Wisconsin--Madison, Madison, WI, USA\\
\textsuperscript{7}RIKEN Center for Interdisciplinary Theoretical and Mathematical Sciences (iTHEMS), Kobe, Japan\\
\textsuperscript{8}Department of Earth Science \& Engineering, Data Science Imperial, Imperial College London, London, UK\\
\textsuperscript{9}RIKEN Center for Computational Science (R-CCS), Kobe, Japan}

\maketitle

\begin{abstract}%
    Koopman representations describe nonlinear dynamics through linear evolution in a transformed space, offering an attractive framework for their linear simplicity and generality. Yet these advantages do not necessarily translate into improved performance in deep latent data assimilation (LDA). We systematically compare seven representative methods and structural baselines across five systems spanning approximately periodic to strongly chaotic dynamics and a broad range of state dimensions. We evaluate forecast and assimilation accuracy, physical state recovery, latent-trajectory behavior, and dynamical consistency, and assess Gaussian error assumptions, objective convexity, and uniqueness of analysis solutions. Our results reveal a trade-off between representational flexibility and optimization tractability. Linear evolution of the latent state can stabilize short-term forecasts but restrict expressiveness in strongly chaotic and high-dimensional systems, without consistently improving assimilation accuracy. Parameterized linear operators offer greater flexibility, yet their joint optimization with the encoder--decoder remains nonconvex. Methods with closed-form analysis updates generally perform better and exhibit greater robustness in our experiments, although matrix inversion remains a computational bottleneck. These findings show that the practical benefits of Koopman structure depend on both sufficient observational information and the choice of assimilation procedure.
\end{abstract}


\section{Introduction}
\label{sec:introduction}
Data assimilation (DA) combines dynamical model forecasts with available observations to correct the estimated system state and improve subsequent forecasts~\citep{carrassi2018data}. For high-dimensional systems, performing the forecast and analysis directly in the physical space can be computationally expensive. Latent data assimilation (LDA) addresses this problem by encoding the physical state into a lower-dimensional latent representation, where the forecast and analysis can be performed at a reduced computational cost~\citep{Mathis2021,cheng2022wildfire,LI2024112953,behnoudfar2026rl,mousavi2026sequential}. By encoding different physical fields into a common latent representation, LDA captures their spatial relationships and allows partial observations to improve the full-state estimate~\citep{cheng2023gla,cheng2024multidomain,Wang2025,behnoudfar2026bridging}.

Although LDA reduces the dimensionality of the assimilation problem, it does not ensure that the learned latent representation retains the statistical properties required by DA~\citep{carrassi2018data,sarkka2023bayesian}. In many LDA frameworks, a neural network defines the latent transition by mapping the current latent state to the next and is applied autoregressively for forecasting~\citep{Mathis2021,cheng2022wildfire}. As a result, the latent dynamics are often highly nonlinear, particularly as complex Transformer architectures become increasingly common in recent research~\citep{Wen2023Transformers,He2025ChaosAttention,Bi2023Pangu,cheng2025unstructured}. After several forecast steps, this nonlinearity can yield non-Gaussian forecast distributions~\citep{Miller1999,bocquet2010beyond,chen2025taming} and non-convex assimilation objectives~\citep{Pires1996,Cox2015,yang2025tensorvar}, complicating uncertainty propagation and latent state estimation. To address this issue, some methods impose additional constraints on the latent representation to encourage an approximately Gaussian latent distribution~\citep{MelincZaplotnik2024,Pasmans2026}. However, these approaches do not explicitly consider how subsequent forecast propagation can shift the latent distribution. These challenges have motivated latent dynamical models that impose explicit structural constraints on the latent transition~\citep{yang2025tensorvar,otto2019linearly}.

Among these structural constraints, linearity is particularly useful for DA because it simplifies forecast propagation and provides a tractable structure for the analysis update when the observation coupling is linear and the errors follow Gaussian distributions~\citep{kalman1960new,sarkka2023bayesian,law2015data,yang2025tensorvar}. Koopman operator theory provides a systematic way to impose this linear structure on latent transitions. Specifically, the finite-dimensional physical state is lifted through a set of observable functions into an infinite-dimensional observable space. Although the original physical dynamics are nonlinear, the resulting observable state evolves linearly under the Koopman operator~\citep{koopman1931hamiltonian,brunton2021modernkoopmantheorydynamical}. Because the exact infinite-dimensional representation is computationally intractable, neural networks are used to learn a finite-dimensional latent approximation of the observable space, with the latent dynamics constrained to be approximately linear~\citep{Lusch2018,otto2019linearly}. Within the learned latent space, the state is propagated by a finite-dimensional linear operator, which supports computationally efficient prediction over long horizons and improves the interpretability of the learned dynamics~\citep{KORDA2018,brunton2021modernkoopmantheorydynamical}.
\begin{figure}
    \centering
    \includegraphics[width=0.8\linewidth]{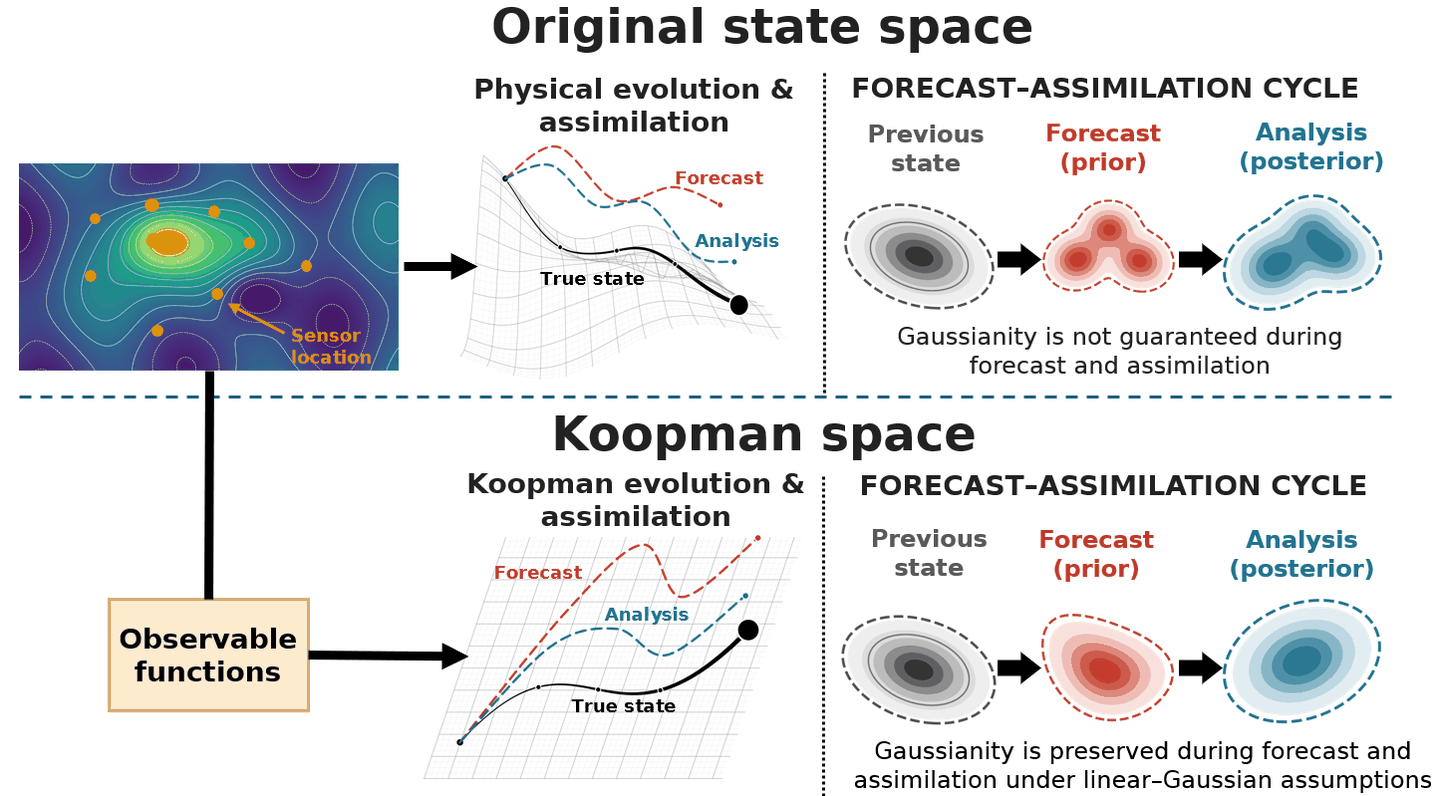}
\caption{Schematic comparison of forecast--analysis cycles in the physical and Koopman latent spaces. The Koopman-space cycle illustrates the linear--Gaussian case.}
    \label{fig:koopman_da_overview}
\end{figure}

A linear latent transition offers potential advantages for uncertainty propagation and latent state estimation in data assimilation. Figure~\ref{fig:koopman_da_overview} compares the forecast--analysis cycles in the physical and Koopman latent spaces. Nonlinear physical dynamics can distort the forecast distribution and make Gaussian-based analysis updates no longer exact. By contrast, under Gaussian error assumptions, a linear latent transition allows the forecast mean and covariance to be computed directly~\citep{kalman1960new,sarkka2023bayesian}. When the observation coupling is also linear, Gaussianity is preserved throughout the forecast--analysis cycle, and the analysis update can be obtained in closed form~\citep{jazwinski1970stochastic,anderson1979optimal}. For variational assimilation, combining linear latent dynamics with linear observation coupling yields a convex quadratic objective that can be solved globally~\citep{law2015data,yang2025tensorvar}.

Building on these advantages, Koopman representations have increasingly been used for state estimation and data assimilation. DMDEnKF combines DMD with an ensemble Kalman filter to track and forecast noisy, time-varying systems, using a linear observation operator for the analysis updates~\citep{Falconer2023DMDEnKF}. However, the linear representation obtained by DMD has limited flexibility for describing complex nonlinear dynamics. KAE--EnKF employs a nonlinear encoder--decoder to learn a more flexible latent representation while retaining a linear latent transition. It encodes new observations into latent space and performs ensemble Kalman updates with linear observation coupling~\citep{falconer2025tracking}, although the nonlinear encoder does not guarantee that observation errors remain Gaussian in latent space. Using a similar learned Koopman representation, Neural Koopman Prior employs pretrained linear latent dynamics as a variational prior to reconstruct and predict system trajectories from sparse and noisy observations~\citep{frion2024neural}. It couples observations to the latent state through a nonlinear decoder in physical space, so the linear latent transition alone does not guarantee a convex assimilation objective or a Gaussian analysis distribution. Tensor-Var instead maps observations into the same latent space as the forecast state through a learned inverse-observation encoder, yielding linear observation coupling when optimizing the latent trajectory. Under Gaussian error assumptions, this structure produces a convex quadratic assimilation objective and preserves Gaussianity in the latent analysis distribution~\citep{yang2025tensorvar}.

These studies seek to exploit linear latent dynamics for uncertainty propagation and analysis updates. However, a fixed linear latent transition may not be expressive enough to describe complex dynamics~\citep{sholokhov2023physics}, while nonlinear observation coupling can leave the analysis objective non-convex. Analysis corrections may also disrupt consistency with Koopman dynamics and the linear evolution of the learned latent representation. Differences in latent representations, observation couplings, and assimilation procedures across existing studies make it difficult to separate their effects on assimilation. This leads to the central question:
\vspace{0.4em}
\begin{center}
\emph{What do different Koopman-type LDA structures contribute, and when are they beneficial?}
\end{center}
\vspace{0.4em}

Motivated by this question, we propose a unified evaluation framework to examine how Koopman representations affect data assimilation. We first impose progressively stronger constraints on the latent transition within a common encoder--decoder architecture to assess whether simplifying forecast propagation also simplifies the analysis update. We next compare decoder-based observation coupling with linear observation coupling in convex variational formulations. To assess the role of the assimilation procedure, we compare variational optimization over an assimilation window with sequential filtering that propagates and updates the latent state and its uncertainty. Finally, we examine how replacing a fixed linear latent transition with an observation-dependent transition affects assimilation performance and the structure of the analysis updates. We do not consider ensemble-based approaches in the current benchmark to avoid additional differences from finite-ensemble sampling and ensemble-specific tuning when comparing Koopman representations. The methods considered here could be easily coupled with an ensemble DA framework.

Within this framework, we conduct experiments across five systems spanning approximately periodic to strongly chaotic dynamics and a broad range of state dimensions. We consider three aspects of latent data assimilation: assimilation accuracy, dynamical consistency, and assimilation tractability. We first evaluate forecast and assimilation accuracy across observation ratios and assimilation frequency. Complementary diagnostics in the physical, latent, and spectral spaces focus on reconstructed spatial structures, latent-trajectory patterns, and the recovery of structures at different spatial scales, respectively. To connect analysis corrections with subsequent forecast propagation, we examine whether analysis trajectories remain consistent with Koopman dynamics and the linear evolution of the learned latent representation. Our evaluation of assimilation tractability covers consistency with Gaussian error assumptions, objective convexity, and solution uniqueness during optimization. Specifically, we assess whether analysis errors remain consistent with Gaussian error assumptions and identify the sources of non-convexity in variational objectives. For convex quadratic objectives, we further examine how strongly the latent trajectory is constrained.

To summarize, the main contributions of this work are as follows:
\begin{itemize}
    \item We provide a systematic evaluation framework that links forecast and assimilation accuracy with dynamical consistency and assimilation tractability to examine the benefits and limitations of Koopman structure in LDA.

    \item We demonstrate that Koopman dynamics can improve forecast stability but do not by themselves ensure effective assimilation; their benefits depend on observation coupling, uncertainty propagation, and how analysis updates are performed, particularly in strongly chaotic and high-dimensional systems.

    \item We perform extensive numerical experiments across five dynamical systems to characterize how the trade-off among forecast stability, representational flexibility, and assimilation tractability changes with dynamical complexity and state dimension.
\end{itemize}

The rest of this paper is organized as follows. Section~\ref{sec:lda} introduces the LDA framework, and Section~\ref{sec:deep-koopman} presents deep-learning-based Koopman representations. Section~\ref{sec:koopman-type-latent-da} describes the evaluated methods and their structural differences, followed by the experimental setup in Section~\ref{sec:experimental-setup}. Section~\ref{sec:numerical-results} presents the numerical results and discusses their implications. Finally, Section~\ref{sec:discussion-future-work} concludes the paper and outlines its limitations and future directions.

\section{Latent Data Assimilation}
\label{sec:lda}
Latent Data Assimilation (LDA) aims to perform data assimilation in a reduced-dimensional latent space, where high-dimensional physical states are represented by compact latent variables~\citep{Mathis2021,cheng2023gla,cheng2025torchda}. This reduced representation lowers the computational cost of the assimilation step and allows the latent state distribution to be constrained more effectively, supporting efficient real-time forecast correction for dynamical systems. Let $\mathbf{x}_{0:T}$ denote a trajectory of physical states up to time $T$, and let $\{\mathbf{y}_{t}\}_{t \in \mathcal{T}}$ denote the available observations, where $\mathcal{T} \subseteq \{0,\ldots,T\}$ is the observation-time index set. The neural network encoder and decoder are denoted by $\mathcal{E}_{x}$ and $\mathcal{D}_{x}$, respectively. Each physical state $\mathbf{x}_{t}$ is mapped to a latent state $\mathbf{z}_{t}$ and reconstructed back to the physical space as
\begin{equation}
    \mathbf{z}_{t} = \mathcal{E}_{x}(\mathbf{x}_{t}),
    \qquad
    \hat{\mathbf{x}}_{t} = \mathcal{D}_{x}(\mathbf{z}_{t}),
    \quad t=0,\ldots,T .
\end{equation}
Here, $\widehat{\mathbf{x}}_{t}$ denotes an approximation of $\mathbf{x}_{t}$ reconstructed from the latent state $\mathbf{z}_{t}$. For simplicity, existing latent DA methods often assume Gaussian background errors directly in the latent space~\citep{cheng2023gla,MelincZaplotnik2024,Tarumi2025}. In this space, the dynamics are described by a latent dynamical model $\mathcal{M}_{z}$, while observations are obtained by applying a time-dependent observation operator $\mathcal{H}_{t}$ to the decoded physical state:
\begin{equation}
\begin{aligned}
    \mathbf{z}_{t}
    &=
    \mathcal{M}_{z}(\mathbf{z}_{t-1}) + \boldsymbol{\epsilon}^{z}_{t},
    &
    \boldsymbol{\epsilon}^{z}_{t}
    &\sim \mathcal{N}(\mathbf{0}, \mathbf{Q}_{z}), \\
    \mathbf{y}_{t}
    &=
    \mathcal{H}_t\circ\mathcal{D}_{x}(\mathbf{z}_{t}) + \boldsymbol{\epsilon}^{y}_{t},
    &
    \boldsymbol{\epsilon}^{y}_{t}
    &\sim \mathcal{N}(\mathbf{0}, \mathbf{R}) .
\end{aligned}
\label{eq:linear-gaussian-lda}
\end{equation}
Here, $\mathbf{Q}_{z}$ and $\mathbf{R}$ denote the latent model error and observation error covariance matrices, respectively. The latent dynamical manifold describes the geometric structure of latent trajectories consistent with the underlying dynamics. Based on how observations are incorporated, LDA can be formulated using either variational optimization or sequential filtering. Variational LDA estimates the latent state or trajectory by minimizing an objective that balances the background, model, and observation errors over an assimilation window. Filtering-based LDA instead propagates the previous latent state distribution to form a forecast distribution and updates it recursively as new observations become available.

In the variational formulation, the latent analysis trajectory can be estimated using the following weak-constraint latent 4D-Var objective~\citep{tremolet2006}:
\begingroup
\small
\begin{equation}
J(\mathbf{z}_{0:T})
=
\underbrace{
\frac{1}{2}
\left\|
\mathbf{z}_{0}-\mathbf{z}_{0}^{\mathrm{b}}
\right\|_{\mathbf{B}^{-1}}^{2}
}_{\text{background error}}
+
\underbrace{
\frac{1}{2}
\sum_{t\in\mathcal{T}}
\left\|
\mathbf{y}_{t}
-
\mathcal{H}_{t}\circ\mathcal{D}_{x}(\mathbf{z}_{t})
\right\|_{\mathbf{R}^{-1}}^{2}
}_{\text{observation error}}
+
\underbrace{
\frac{1}{2}
\sum_{t=1}^{T}
\left\|
\mathbf{z}_{t}
-
\mathcal{M}_{z}(\mathbf{z}_{t-1})
\right\|_{\mathbf{Q}_{z}^{-1}}^{2}
}_{\text{latent model error}},
\label{eq:weak-4d-var}
\end{equation}
\endgroup
Here, $\|\mathbf{v}\|_{\mathbf{A}}^{2}=\mathbf{v}^{\top}\mathbf{A}\mathbf{v}$ denotes the squared weighted norm, and $\mathbf{z}_{0}^{\mathrm{b}}$ denotes the background latent state. The matrix $\mathbf{B}$ is the latent background error covariance. The covariance matrices $\mathbf{Q}_{z}$ and $\mathbf{R}$ are often assumed to be time-independent and remain fixed across all assimilation cycles. Assuming that the latent dynamical model $\mathcal{M}_{z}$ is differentiable, the latent analysis trajectory $\mathbf{z}_{0:T}^{\mathrm{a}}$ is obtained by minimizing the objective function using gradient-based optimization:

\begin{equation}
    \mathbf{z}_{0:T}^{\mathrm{a}}
    =
    \arg\min_{\mathbf{z}_{0:T}}
    J(\mathbf{z}_{0:T}) .
\end{equation}
In this formulation, strong-constraint 4D-Var~\citep{le1986variational} and 3D-Var~\citep{3dvar} are two special cases: the former assumes a perfect latent dynamical model and optimizes only the initial latent state, whereas the latter updates the current latent state using the background and current observations.

For filtering-based LDA, the latent state is updated recursively as observations become available. Under linear--Gaussian assumptions, the forecast and analysis updates admit analytical expressions. Specifically, let $\widetilde{\mathbf{M}}$ and $\widetilde{\mathbf{H}}_{t}$ denote linear approximations of the latent transition and the decoder-based observation coupling, respectively:
\begin{equation}
\mathcal{M}_{z}(\mathbf{z}_{t-1})
\approx
\widetilde{\mathbf{M}}\mathbf{z}_{t-1},
\qquad
\mathcal{H}_{t}\circ\mathcal{D}_{x}(\mathbf{z}_{t})
\approx
\widetilde{\mathbf{H}}_{t}\mathbf{z}_{t}.
\end{equation}
With these approximations, the latent dynamics and observations can be written as
\begin{equation}
\begin{aligned}
    \mathbf{z}_{t}
    &=
    \widetilde{\mathbf{M}}\mathbf{z}_{t-1}
    +
    \boldsymbol{\epsilon}^{z}_{t},
    &
    \boldsymbol{\epsilon}^{z}_{t}
    &\sim \mathcal{N}(\mathbf{0}, \mathbf{Q}_{z}), \\
    \mathbf{y}_{t}
    &=
    \widetilde{\mathbf{H}}_t\mathbf{z}_{t}
    +
    \boldsymbol{\epsilon}^{y}_{t},
    &
    \boldsymbol{\epsilon}^{y}_{t}
    &\sim \mathcal{N}(\mathbf{0}, \mathbf{R}) .
\end{aligned}
\end{equation}
In the linear--Gaussian case, minimizing the variational objective in Eq.~\eqref{eq:weak-4d-var} yields the same latent trajectory estimate as fixed-interval Kalman smoothing~\citep{carrassi2018data,sarkka2023bayesian}. When only past and current observations are used, the Kalman filter applies the following sequential updates:
\begin{equation}
\label{eq:kalman-forecast}
\begin{aligned}
\mathbf{z}_{t}^{f}
&=
\widetilde{\mathbf{M}}\mathbf{z}_{t-1}^{\mathrm{a}},
\qquad
\boldsymbol{\Sigma}_{t}^{f}
=
\widetilde{\mathbf{M}}
\boldsymbol{\Sigma}_{t-1}^{\mathrm{a}}
\widetilde{\mathbf{M}}^{\top}
+
\mathbf{Q}_{z}.
\end{aligned}
\end{equation}

\begin{equation}
\label{eq:kalman-analysis}
\renewcommand{\arraystretch}{1}
\begin{array}{@{}c@{\qquad\qquad}c@{}}
\text{No observation }(t\notin\mathcal{T})
&
\text{With observation }(t\in\mathcal{T})
\\[0.5em]
\begin{aligned}
\mathbf{z}_{t}^{\mathrm{a}}
&=
\mathbf{z}_{t}^{f}, \\
\boldsymbol{\Sigma}_{t}^{\mathrm{a}}
&=
\boldsymbol{\Sigma}_{t}^{f}
\end{aligned}
&
\begin{aligned}
\mathbf{G}_{t}
&=
\boldsymbol{\Sigma}_{t}^{f}
\widetilde{\mathbf{H}}_{t}^{\top}
\left(
\widetilde{\mathbf{H}}_{t}
\boldsymbol{\Sigma}_{t}^{f}
\widetilde{\mathbf{H}}_{t}^{\top}
+
\mathbf{R}
\right)^{-1}, \\
\mathbf{z}_{t}^{\mathrm{a}}
&=
\mathbf{z}_{t}^{f}
+
\mathbf{G}_{t}
\left(
\mathbf{y}_{t}
-
\widetilde{\mathbf{H}}_{t}\mathbf{z}_{t}^{f}
\right), \\
\boldsymbol{\Sigma}_{t}^{\mathrm{a}}
&=
\left(
\mathbf{I}
-
\mathbf{G}_{t}\widetilde{\mathbf{H}}_{t}
\right)
\boldsymbol{\Sigma}_{t}^{f}.
\end{aligned}
\end{array}
\end{equation}
Here, $\mathbf{z}_{t}^{f}$ denotes the forecast latent state before assimilating $\mathbf{y}_{t}$, and $\mathbf{z}_{t}^{\mathrm{a}}$ denotes the analysis latent state after the update. 
The matrix $\mathbf{G}_{t}$ is the Kalman gain, while $\boldsymbol{\Sigma}_{t}^{f}$ and $\boldsymbol{\Sigma}_{t}^{\mathrm{a}}$ are the forecast error and analysis error covariance matrices of the latent state.

Variational and filtering-based data assimilation methods are often formulated with Gaussian error models, as specified in Eq.~\eqref{eq:linear-gaussian-lda}. In variational data assimilation, these assumptions lead to a quadratic objective; when the dynamical and observation operators are linear, the resulting problem is convex. In filtering-based data assimilation, the same linear--Gaussian assumptions make the Kalman update exact and preserve the Gaussianity of the latent state distribution through the forecast and analysis steps. However, in practice, the dynamical model and the observation operator are often strongly nonlinear. Such nonlinear mappings generally do not preserve Gaussianity, leading to non-Gaussian filtering distributions and non-convex variational objectives~\citep{sarkka2023bayesian}. This motivates the use of Koopman-based latent representations, which seek to approximate nonlinear dynamics by linear evolution in latent space.

\section{Deep Learning Koopman Operator}
\label{sec:deep-koopman}

For a nonlinear dynamical model $\mathcal{M}$ acting on the physical state $\mathbf{x}$, the transition from time $t$ to time $t+1$ is given by
\begin{equation}
    \mathbf{x}_{t+1}
    =
    \mathcal{M}(\mathbf{x}_{t}) .
\end{equation}
Koopman theory represents these nonlinear dynamics through an infinite-dimensional observable space~\citep{koopman1931hamiltonian,budisic2012applied,mezic2013analysis,brunton2021modernkoopmantheorydynamical}. Let $g$ denote an observable in this space. The Koopman operator $\mathcal{K}$ acts linearly on $g$ according to
\begin{equation}
    \left(\mathcal{K}g\right)(\mathbf{x}_{t})
    =
    g\!\left(\mathcal{M}(\mathbf{x}_{t})\right)
    =
    g(\mathbf{x}_{t+1}) .
    \label{eq:koopman-observable-evolution}
\end{equation}
However, because the infinite-dimensional observable space is computationally intractable, a finite set of observables is used to construct a finite-dimensional approximation of the relevant dynamics. Let
\begin{equation}
    \mathcal{F}_{d}
    =
    \operatorname{span}
    \left\{
        e_{1},
        e_{2},
        \ldots,
        e_{d}
    \right\}
\end{equation}
be a finite-dimensional subspace of observable functions, where each $e_i$ is an observable of the physical state. This subspace is Koopman invariant if
\begin{equation}
    \mathcal{K}\mathcal{F}_{d}
    \subseteq
    \mathcal{F}_{d}.
\end{equation}
Under this condition, the evolution of these observables can be described exactly by a finite-dimensional Koopman operator~\citep{Brunton_2016}. Otherwise, a finite-dimensional linear operator only approximates their Koopman evolution~\citep{Williams2015}.

Deep-learning Koopman models~\citep{Li2017extended,takeishi2017learning,Lusch2018,otto2019linearly,azencot2020forecasting,frion2024neural} parameterize these observables using an encoder $\mathcal{E}_{x}$, whose output defines the latent state
\begin{equation}
    \mathbf{z}_{t}
    =
    \mathcal{E}_{x}(\mathbf{x}_{t}) .
\end{equation}
The corresponding Koopman evolution is then approximated by a finite-dimensional linear operator $\mathbf{K}$:
\begin{equation}
    \mathbf{z}_{t+1}
    \approx
    \mathbf{K}\mathbf{z}_{t},
    \qquad
    \widehat{\mathbf{x}}_{t+1}
    =
    \mathcal{D}_{x}(\mathbf{K}\mathbf{z}_{t}) .
\end{equation}
Here, $\mathcal{E}_{x}$ learns the observable functions, while $\mathcal{D}_{x}$ maps the latent state back to the physical space~\citep{Lusch2018,otto2019linearly}. The operator $\mathbf{K}$ can be parameterized as a trainable matrix and optimized jointly with the encoder and decoder through reconstruction and prediction losses. Enforcing a linear latent transition does not by itself ensure that the learned observables span a Koopman-invariant subspace. In practice, the learned latent representation generally only approximates the Koopman dynamics~\citep{Brunton_2016,Lusch2018,otto2019linearly}. The approximation error can accumulate during prediction, causing the latent trajectory to drift into regions where the decoder is less reliable and degrading the reconstructed forecast.

Overall, the Koopman operator allows the latent dynamics to be described by a linear transition. This structure could make data assimilation more tractable under linear--Gaussian assumptions. For variational methods, the Koopman operator replaces the nonlinear latent transition with a linear model constraint, although the decoder-based observation coupling may remain nonlinear. For filtering-based methods, the linear latent transition allows the forecast mean and covariance to be propagated analytically. The Koopman-based latent representation therefore provides a common dynamical structure for both variational and filtering-based LDA methods.

\section{Koopman-Type Latent Data Assimilation}
\label{sec:koopman-type-latent-da}

The previous sections show that linear latent evolution has the potential to improve assimilation accuracy under linear-Gaussian assumptions. However, linearizing the dynamics alone is insufficient because the observation coupling may still introduce nonlinearity into the assimilation problem. This section therefore examines Koopman-type latent data assimilation methods according to how they structure the latent dynamics and incorporate observational information into the assimilation procedure. These choices determine whether assimilation requires nonlinear optimization, can be formulated as a convex problem, or can be performed through closed-form filtering updates.

\begin{figure}[htbp]
    \centering
    \includegraphics[width=\linewidth]{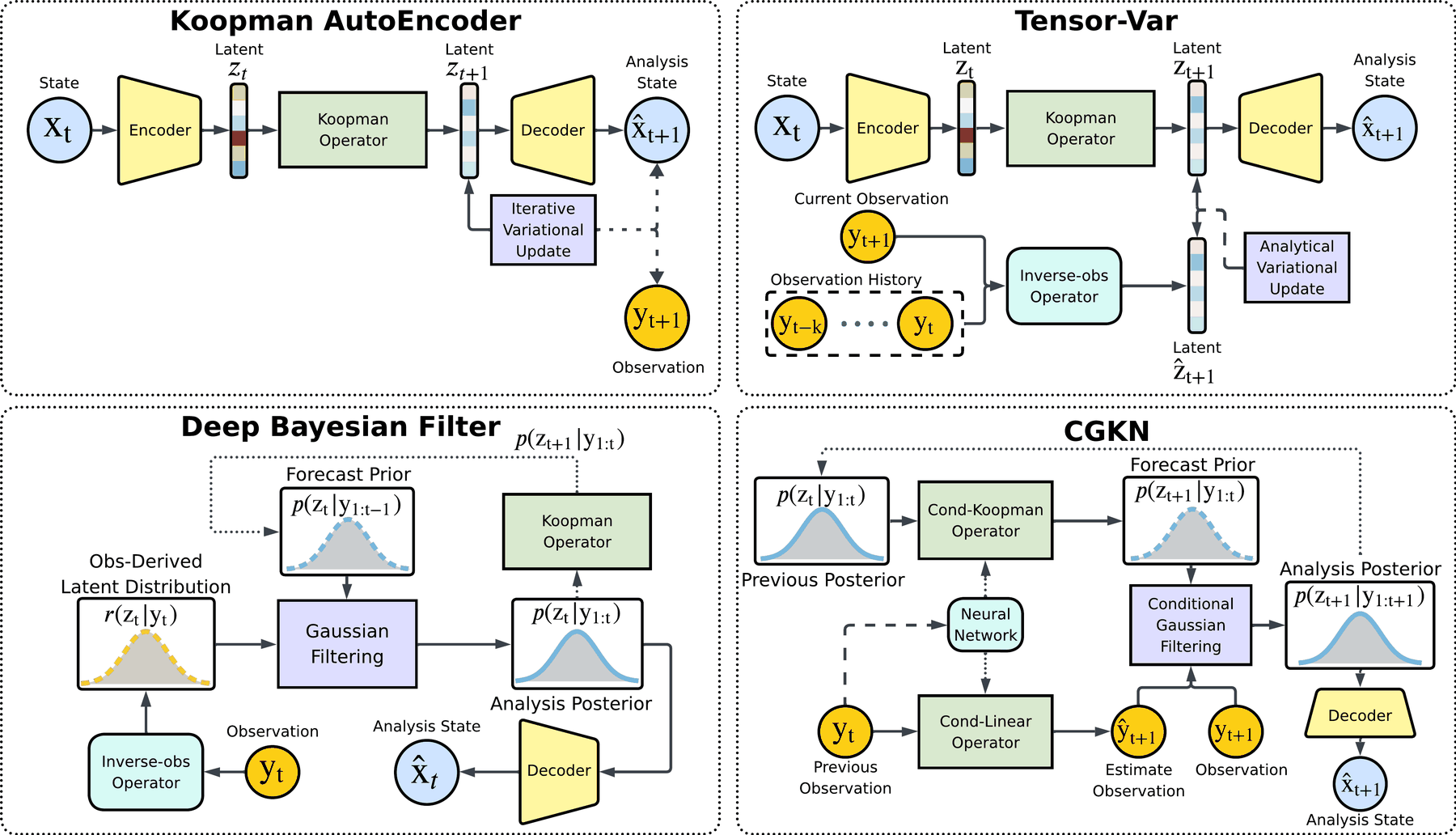}
    \caption{Schematic comparison of the inference pipelines for KAE (top left), Tensor-Var (top right), DBF (bottom left), and CGKN (bottom right). DBF and CGKN perform filtering-based assimilation, whereas Tensor-Var and KAE use variational assimilation.}
    \label{fig:methodology}
\end{figure}

We first consider variational methods, including DMD~\citep{Schmid2010,tu2014dynamic}, the autoencoder variants, and Tensor-Var~\citep{yang2025tensorvar}, which differ in the structures used for latent dynamics and observation coupling. We then turn to the filtering-based methods DBF~\citep{Tarumi2025} and CGKN~\citep{Chen_2025}, which provide recursive latent inference with uncertainty propagation. Figure~\ref{fig:methodology} compares the representative inference pipelines of these models.

\subsection{Autoencoder Variants (NL-AE, WL-AE, KAE)}
\label{sec:kae-family}

DMD provides a traditional data-driven approach based on a low-rank linear representation of the dynamics. However, some state trajectories cannot be represented accurately within a low-dimensional linear subspace~\citep{LEE2020108973}. Autoencoders (AEs) parameterized by neural networks have been widely used to encode high-dimensional physical states into low-dimensional latent representations, allowing such trajectories to be represented more compactly and flexibly~\citep{LEE2020108973,Alford_Lago_2022}.

Existing LDA methods commonly use autoencoders to construct latent representations while modeling the latent transition with a general nonlinear neural network, without explicitly constraining the latent transition to be linear~\citep{Mathis2021,cheng2022wildfire,LI2024112953,Wang2025Latent4DVar}. To examine the role of linearity in the latent dynamics, we consider three autoencoder-based variants with progressively stronger constraints: the nonlinear autoencoder (NL-AE), weak-linear autoencoder (WL-AE), and Koopman autoencoder (KAE). NL-AE uses a nonlinear neural network transition. WL-AE retains this nonlinear transition but adds latent-space regularization to promote approximately linear evolution, whereas KAE explicitly imposes a linear latent transition through a learned Koopman operator:

\begin{equation}
    \mathbf{z}_{t+1}
    =
    \mathcal{M}_{z}(\mathbf{z}_{t})
    =
    \begin{cases}
        \mathcal{M}_{\theta}(\mathbf{z}_{t}),
        & \text{NL-AE and WL-AE},\\
        \mathbf{K}\mathbf{z}_{t},
        & \text{KAE}.
    \end{cases}
\end{equation}
Here, \(\mathcal{M}_{\theta}\) denotes a nonlinear neural network transition parameterized by \(\theta\), while \(\mathbf{K}\) denotes the learned linear Koopman operator. Both approximate the evolution of the encoded physical states; the encoder--decoder also introduces reconstruction error. Combining the autoencoder with the nonlinear neural network transition yields our baseline NL-AE model, whose training objective consists of physical state reconstruction and multi-step prediction losses:
\begin{equation}
    \mathcal{L}_{\mathrm{rec}}
    =
    \sum_{t}
    \left\|
        \mathcal{D}_{x}(\mathbf{z}_{t})
        -
        \mathbf{x}_{t}
    \right\|^{2},
    \qquad
    \mathcal{L}_{\mathrm{pred}}
    =
    \sum_{t,\tau}
    \left\|
        \mathcal{D}_{x}\!\left(
            \mathcal{M}_{z}^{(\tau)}(\mathbf{z}_{t})
        \right)
        -
        \mathbf{x}_{t+\tau}
    \right\|^{2}.
\end{equation}
Here, $\tau$ denotes the prediction horizon in model steps, and $\mathcal{M}_{z}^{(\tau)}$ denotes $\tau$ successive applications of the latent transition. Since both losses are evaluated in the physical state space, they do not directly regularize the latent trajectories. WL-AE therefore extends NL-AE with two latent-space objectives. The latent-consistency loss $\mathcal{L}_{\mathrm{latent}}$ aligns multi-step latent predictions with the encoded future states, encouraging consistency between latent forecast propagation and the encoded trajectory, whereas the temporal-straightening regularization $\mathcal{L}_{\mathrm{straight}}$~\citep{wang2026temporalstraighteninglatentplanning} penalizes changes in direction between consecutive latent displacements, thereby encouraging locally straighter latent trajectories that are easier to approximate with a linear transition model:
\begingroup
\small
\begin{equation}
\mathcal{L}_{\mathrm{latent}}
=
\sum_{t,\tau}
\left\|
\mathcal{M}_{z}^{(\tau)}(\mathbf{z}_{t})
-
\mathbf{z}_{t+\tau}
\right\|^{2},
\quad
\mathcal{L}_{\mathrm{straight}}
=
\sum_{t}
\left(
1-
\frac{
(\mathbf{z}_{t+1}-\mathbf{z}_{t})^{\top}
(\mathbf{z}_{t+2}-\mathbf{z}_{t+1})
}{
\|\mathbf{z}_{t+1}-\mathbf{z}_{t}\|
\|\mathbf{z}_{t+2}-\mathbf{z}_{t+1}\|
}
\right),
\end{equation}
\endgroup
Despite these latent-space constraints, WL-AE still models the latent evolution using the nonlinear transition $\mathcal{M}_{\theta}$. KAE imposes linearity explicitly by replacing the nonlinear transition network with a single bias-free linear layer without an activation function, $\mathbf{z}_{t+1}=\mathbf{K}\mathbf{z}_{t}$, where the trainable weight matrix $\mathbf{K}$ provides a finite-dimensional approximation of Koopman evolution. This linear structure does not guarantee that the encoded physical states evolve exactly under $\mathbf{K}$. The latent-consistency loss $\mathcal{L}_{\mathrm{latent}}$ constrains the latent prediction error without restricting the eigenvalue magnitudes of $\mathbf{K}$. Repeated propagation can therefore lead to exponential growth in the latent state and degrade long-horizon predictions. To suppress this instability, we adopt the spectral-radius penalty introduced in \citet{ElHussieny2026}:
\begin{equation}
    \mathcal{L}_{\mathrm{spec}}
    =
    \left[
        \max\!\left(
            0,
            \widehat{\rho}(\mathbf{K})-1
        \right)
    \right]^2,
\end{equation}
where $\widehat{\rho}(\mathbf{K})$ denotes an estimate of the largest eigenvalue magnitude of $\mathbf{K}$. All three autoencoder variants use the reconstruction and prediction losses. For NL-AE, these two terms define the training objective,
$\mathcal{L}_{\mathrm{NL\text{-}AE}}
=\lambda_{\mathrm{rec}}\mathcal{L}_{\mathrm{rec}}
+\lambda_{\mathrm{pred}}\mathcal{L}_{\mathrm{pred}}$,
where each $\lambda$ denotes the weight of the corresponding loss term. WL-AE and KAE extend this objective with the latent-space regularization terms described above:
\begin{equation}
\begin{aligned}
    \mathcal{L}_{\mathrm{WL\text{-}AE}}
    &=
    \mathcal{L}_{\mathrm{NL\text{-}AE}}
    +\lambda_{\mathrm{latent}}\mathcal{L}_{\mathrm{latent}}
    +\lambda_{\mathrm{straight}}\mathcal{L}_{\mathrm{straight}},\\
    \mathcal{L}_{\mathrm{KAE}}
    &=
    \mathcal{L}_{\mathrm{NL\text{-}AE}}
    +\lambda_{\mathrm{latent}}\mathcal{L}_{\mathrm{latent}}
    +\lambda_{\mathrm{spec}}\mathcal{L}_{\mathrm{spec}}.
\end{aligned}
\end{equation}
During assimilation, the learned transitions are incorporated into the weak-constraint latent 4D-Var objective introduced in Section~\ref{sec:lda}. These transition choices directly affect the structure of the variational assimilation objective. In NL-AE and WL-AE, the nonlinear transitions generally make the latent model term non-convex. By contrast, the linear latent transition in KAE makes the model error term convex quadratic. The complete objective can still be non-convex because of the nonlinear decoder-based observation coupling.

\subsection{Tensor-Var}
Tensor-Var~\citep{yang2025tensorvar} addresses the remaining nonlinearity in the observation term while retaining the state encoder $\mathcal{E}_{x}$ and decoder $\mathcal{D}_{x}$ used by the autoencoder variants. Rather than evaluating the observation residual through the nonlinear decoder during assimilation, it introduces an approximate inverse-observation encoder $\mathcal{E}_{y}$, parameterized by a neural network, to map observational information into the same latent space as the encoded physical state. The observation residual can therefore be evaluated directly in the latent space without applying the decoder $\mathcal{D}_{x}$ during assimilation. Similar observation encoders have also been adopted in other latent DA frameworks to obtain state-compatible latent representations from partial observations~\citep{Pasmans2026,cheng2024multidomain}.

Although evaluating the observation residual directly in the latent space avoids the decoder-based observation coupling, the latent state inferred from the observations must still retain sufficient information to represent the original observation constraint. Tensor-Var addresses this question by constructing the inverse-observation operator through conditional mean embeddings (CMEs) in a reproducing kernel Hilbert space (RKHS). Under the embedding and stability conditions associated with Kazantzis--Kravaris--Luenberger (KKL) observer theory~\citep{AndrieuPraly2006,TranBernard2023}, the resulting latent-space observation term can be shown to be consistent with the corresponding physical-space observation term~\citep{yang2025tensorvar}. To supplement the information provided by the current observation, Tensor-Var combines it with a history of $m$ past observations. The available observation history $\mathbf{h}_{t}$ and the observation-sequence feature vector $\mathbf{v}_{t}$ are defined as
\begin{equation}
\begin{gathered}
    \mathbf{h}_{t}=(\mathbf{y}_{t-m},\ldots,\mathbf{y}_{t-1}),
    \qquad
    \mathbf{v}_{t}=\boldsymbol{\Phi}(\mathbf{y}_{t},\mathbf{h}_{t}).
\end{gathered}
\label{eq:inverse-observation-definitions}
\end{equation}
Here, $\boldsymbol{\Phi}$ maps the available observation sequence to the feature vector $\mathbf{v}_{t}$ in the RKHS feature space. For the training samples, the encoded latent snapshots and their corresponding feature vectors are collected as $\mathbf{Z}=\left[\mathbf{z}_{1},\ldots,\mathbf{z}_{N-1}\right]$ and $\mathbf{V}=\left[\mathbf{v}_{1},\ldots,\mathbf{v}_{N-1}\right]$. These paired matrices define the conditional mean embedding (CME) operator through ridge regression:
\begin{equation}
\begin{gathered}
    \widehat{\mathbf{C}}_{y}
    =
    \mathbf{Z}\mathbf{V}^{\top}
    \left(\mathbf{V}\mathbf{V}^{\top}+\lambda\mathbf{I}\right)^{-1},
    \qquad
    \widehat{\mathbf{z}}_{t}
    =
    \mathcal{E}_{y}(\mathbf{y}_{t},\mathbf{h}_{t})
    =
    \widehat{\mathbf{C}}_{y}\mathbf{v}_{t}.
\end{gathered}
\label{eq:inverse-observation-map}
\end{equation}
Here, $\widehat{\mathbf{C}}_{y}$ is the CME operator, and $\lambda$ controls the ridge regularization strength. Applying this operator to the feature vector produced from the available observation sequence gives the latent state estimate $\widehat{\mathbf{z}}_{t}$, which approximates the encoded physical state $\mathcal{E}_{x}(\mathbf{x}_{t})$; the resulting map is denoted by the inverse-observation operator $\mathcal{E}_{y}$. Following the deep-feature approach of Tensor-Var~\citep{yang2025tensorvar}, we use neural networks to approximate the RKHS feature maps in a finite-dimensional latent space.

For the latent dynamics, Tensor-Var retains the linear transition used by KAE to approximate the evolution of the encoded physical states. The state encoder $\mathcal{E}_{x}$ and decoder $\mathcal{D}_{x}$ are pretrained using the reconstruction loss $\mathcal{L}_{\mathrm{rec}}$ and prediction loss $\mathcal{L}_{\mathrm{pred}}$. Pairs of consecutive encoded states are collected in the snapshot matrices $\mathbf{Z}$ and $\mathbf{Z}^{+}$, containing the current and corresponding next-step latent states, respectively. The Koopman operator is estimated via dual ridge regression:
\begin{equation}
\mathbf{Z}^{+}=\left[\mathbf{z}_{2},\ldots,\mathbf{z}_{N}\right],
\qquad
\mathbf{K}=\mathbf{Z}^{+}
\left(\mathbf{Z}^{\top}\mathbf{Z}+\lambda_{K}\mathbf{I}\right)^{-1}
\mathbf{Z}^{\top},
\qquad
\mathcal{L}_{\mathrm{pred}}
=
\sum_{t}
\left\|
\mathcal{D}_{x}\!\left(\mathbf{K}\mathbf{z}_{t}\right)
-\mathbf{x}_{t+1}
\right\|^{2}.
\label{eq:latent-snapshot-matrices}
\end{equation}
where $\lambda_{K}$ denotes the ridge regularization parameter. During training, $\mathbf{K}$ is estimated from encoded state pairs within each minibatch. After training, the encoder and decoder are held fixed, and $\mathbf{K}$ is re-estimated from encoded state pairs drawn from the training data set.

The inverse-observation encoder $\mathcal{E}_{y}$ is trained to align the observation-derived latent state with the encoded physical state and recover the physical state through the fixed decoder:
\begin{equation}
\begin{aligned}
\mathcal{L}_{\mathrm{inv}}
={}&
\sum_{t}
\left\|
\mathcal{E}_{y}(\mathbf{y}_{t},\mathbf{h}_{t})
-
\mathcal{E}_{x}(\mathbf{x}_{t})
\right\|^{2}
+
\lambda_{\mathrm{rec}}\sum_{t}
\left\|
\mathcal{D}_{x}\!\left(
\mathcal{E}_{y}(\mathbf{y}_{t},\mathbf{h}_{t})
\right)
-
\mathbf{x}_{t}
\right\|^{2},
\end{aligned}
\end{equation}
where $\lambda_{\mathrm{rec}}$ weights the physical state reconstruction term.

With $\mathcal{E}_{y}$ and $\mathbf{K}$ fixed during inference, Tensor-Var performs variational assimilation entirely in the latent space. The analysis trajectory is obtained by minimizing
\begin{equation}
\begin{aligned}
    J(\mathbf{z}_{0:T})
    =
    \frac{1}{2}
    \left\|
        \mathbf{z}_{0}
        -
        \mathbf{z}_{0}^{\mathrm{b}}
    \right\|_{\mathbf{B}^{-1}}^{2}
    +
    \frac{1}{2}
    \sum_{t=1}^{T}
    \left\|
        \mathbf{z}_{t}
        -
        \mathbf{K}\mathbf{z}_{t-1}
    \right\|_{\mathbf{Q}_{z}^{-1}}^{2}
    +
    \frac{1}{2}
    \sum_{t\in\mathcal{T}}
    \left\|
        \mathbf{z}_{t}
        -
        \widehat{\mathbf{z}}_{t}
    \right\|_{\mathbf{R}_{z}^{-1}}^{2}.
\end{aligned}
\end{equation}
Here, $\mathbf{B}$, $\mathbf{Q}_{z}$, and $\mathbf{R}_{z}$ denote the covariance matrices for background errors, latent model errors, and observation errors, respectively. The three terms enforce consistency with the latent background, learned dynamics, and observation-derived latent states. By replacing the decoder-based observation residual in KAE with the latent-space residual $\mathbf{z}_{t}-\widehat{\mathbf{z}}_{t}$, Tensor-Var makes the observation term quadratic in the latent state. Together with the linear transition $\mathbf{K}$, this yields a convex quadratic objective in the latent trajectory, but the additional hyperparameters require tuning and can affect assimilation performance. DMD has the same linear-transition and fixed-target structure, so both methods admit closed-form solutions that exactly minimize their respective latent quadratic objectives without iterative optimization, which can substantially reduce the computational cost compared with conventional iterative solvers. The corresponding derivation is provided in Appendix~\ref{app:closed-form-4dvar}.

\subsection{Deep Bayesian Filter (DBF)}
Although Tensor-Var yields a convex variational objective, it still requires solving a trajectory optimization problem over each assimilation window. A linear latent transition makes the forecast step analytically tractable, but obtaining a closed-form recursive analysis update under Gaussian assumptions also requires preserving the Gaussian latent state distribution when incorporating new observations.

DBF~\citep{Tarumi2025} addresses this limitation by introducing a Gaussian inverse-observation operator. Whereas Tensor-Var maps the current observation and its history to a latent target $\widehat{\mathbf{z}}_{t}$, DBF uses a Gaussian distribution with a learned mean and covariance to approximate the latent state distribution conditioned on the current observation. This operator is combined with a linear Gaussian transition that approximates the one-step evolution of the latent state:
\begingroup
\small
\begin{equation}
(\boldsymbol{\mu}_{t}^{o},\boldsymbol{\Sigma}_{t}^{o})
=
\mathcal{G}_{\theta}(\mathbf{y}_{t}),
\quad
r(\mathbf{z}_{t}\mid\mathbf{y}_{t})
=
\mathcal{N}\!\left(
\mathbf{z}_{t};
\boldsymbol{\mu}_{t}^{o},
\boldsymbol{\Sigma}_{t}^{o}
\right),
\quad
p(\mathbf{z}_{t}\mid\mathbf{z}_{t-1})
=
\mathcal{N}\!\left(
\mathbf{z}_{t};
\mathbf{K}\mathbf{z}_{t-1},
\mathbf{Q}_{z}
\right),
\end{equation}
\endgroup
The inverse-observation network $\mathcal{G}_{\theta}$ maps the current observation $\mathbf{y}_{t}$ to the mean $\boldsymbol{\mu}_{t}^{o}$ and covariance $\boldsymbol{\Sigma}_{t}^{o}$ of a Gaussian distribution over the latent state, denoted by $r(\mathbf{z}_{t}\mid\mathbf{y}_{t})$. The distribution $p(\mathbf{z}_{t}\mid\mathbf{z}_{t-1})$ denotes the linear Gaussian latent transition, where $\mathbf{K}$ is the learned linear Koopman operator and $\mathbf{Q}_{z}$ is a predefined latent model error covariance, held fixed during training and assimilation.

DBF applies the latent transition and inverse-observation distribution sequentially within a forecast--analysis cycle, beginning with the forecast step. Let $q(\mathbf{z}_{t-1}\mid\mathbf{y}_{1:t-1})$ denote the previous Gaussian analysis posterior, with mean $\boldsymbol{\mu}_{t-1}^{\mathrm{a}}$ and covariance $\boldsymbol{\Sigma}_{t-1}^{\mathrm{a}}$. Propagating this posterior through the linear Gaussian transition preserves Gaussianity, so the forecast prior can be computed directly as in Eq.~\eqref{eq:kalman-forecast}:
\begin{equation}
    p(\mathbf{z}_{t}\mid\mathbf{y}_{1:t-1})
    =
    \mathcal{N}\!\left(
        \mathbf{z}_{t};
        \boldsymbol{\mu}_{t}^{f},
        \boldsymbol{\Sigma}_{t}^{f}
    \right),
    \qquad
    \boldsymbol{\mu}_{t}^{f}
    =
    \mathbf{K}\boldsymbol{\mu}_{t-1}^{\mathrm{a}},
    \qquad
    \boldsymbol{\Sigma}_{t}^{f}
    =
    \mathbf{K}\boldsymbol{\Sigma}_{t-1}^{\mathrm{a}}\mathbf{K}^{\top}
    +
    \mathbf{Q}_{z}.
\end{equation}
The forecast step propagates the previous analysis mean and uncertainty through $\mathbf{K}$ and incorporates the model uncertainty represented by $\mathbf{Q}_{z}$.

To incorporate the current observation, DBF introduces a Gaussian virtual prior
$\rho(\mathbf{z}_{t})
=
\mathcal{N}(\mathbf{z}_{t};\mathbf{0},\lambda\mathbf{I})$
with $\lambda \gg 1$.
By construction,
$r(\mathbf{z}_{t}\mid\mathbf{y}_{t})
\propto
p(\mathbf{y}_{t}\mid\mathbf{z}_{t})\rho(\mathbf{z}_{t})$,
so dividing by $\rho(\mathbf{z}_{t})$ recovers the observation likelihood required for the Bayesian update. The analysis posterior is therefore given by
\begin{equation}
    q(\mathbf{z}_{t}\mid\mathbf{y}_{1:t})
    =
    \mathcal{N}\!\left(
        \mathbf{z}_{t};
        \boldsymbol{\mu}_{t}^{\mathrm{a}},
        \boldsymbol{\Sigma}_{t}^{\mathrm{a}}
    \right)
    \propto
    \frac{
        r(\mathbf{z}_{t}\mid\mathbf{y}_{t})
    }{
        \rho(\mathbf{z}_{t})
    }
    p(\mathbf{z}_{t}\mid\mathbf{y}_{1:t-1}).
\end{equation}

Because all terms in this update have Gaussian form, the analysis posterior remains Gaussian, and its mean and covariance are obtained in closed form:
\begin{equation}
    \left(\boldsymbol{\Sigma}_{t}^{\mathrm{a}}\right)^{-1}
    =
    \left(\boldsymbol{\Sigma}_{t}^{f}\right)^{-1}
    +
    \left(\boldsymbol{\Sigma}_{t}^{o}\right)^{-1}
    -
    \lambda^{-1}\mathbf{I},
    \qquad
    \boldsymbol{\mu}_{t}^{\mathrm{a}}
    =
    \boldsymbol{\Sigma}_{t}^{\mathrm{a}}
    \left[
        \left(\boldsymbol{\Sigma}_{t}^{f}\right)^{-1}\boldsymbol{\mu}_{t}^{f}
        +
        \left(\boldsymbol{\Sigma}_{t}^{o}\right)^{-1}\boldsymbol{\mu}_{t}^{o}
    \right].
\end{equation}

These closed-form updates complete the recursive forecast--analysis cycle. When no observation is available, the forecast prior is retained as the analysis posterior.

DBF learns the transition, decoder, and inverse-observation operator using a physical state reconstruction term and the Kullback--Leibler (KL) divergence~\citep{kullback1951information} between the analysis posterior and the forecast prior:
\begin{equation}
    \mathcal{L}_{\mathrm{DBF}}
    =
    \sum_{t}
    \left[
        -\mathbb{E}_{q(\mathbf{z}_{t}\mid\mathbf{y}_{1:t})}
        \log p(\mathbf{x}_{t}\mid\mathbf{z}_{t})
        +
        \mathrm{KL}\!\left(
            q(\mathbf{z}_{t}\mid\mathbf{y}_{1:t})
            \,\middle\|\,
            p(\mathbf{z}_{t}\mid\mathbf{y}_{1:t-1})
        \right)
    \right],
\label{eq:dbf-training-loss}
\end{equation}
where $p(\mathbf{x}_{t}\mid\mathbf{z}_{t})$ is parameterized by the decoder $\mathcal{D}_{x}$. The expanded loss in terms of the learned operators is given in Appendix~\ref{app:dbf-update}. During inference, the analysis mean $\boldsymbol{\mu}_{t}^{\mathrm{a}}$ is passed through $\mathcal{D}_{x}$ to obtain the physical state estimate.

While its Gaussian forecast--analysis cycle enables closed-form recursive inference, DBF represents the latent evolution using a fixed Koopman operator. Consequently, changes in the latent dynamics across different observed conditions must be represented by the same global linear transition, which may limit its expressiveness.

\subsection{Conditional Gaussian Koopman Network (CGKN)}
CGKN~\citep{CHEN2025113950,Chen_2025} relaxes the globally linear latent transition constraint. It is motivated by the conditional Gaussian nonlinear system (CGNS) framework~\citep{liptser2013statistics,chen2022conditional}, which describes a broad class of nonlinear stochastic systems whose unresolved variables evolve linearly conditional on the observed variables~\citep{chen2018conditional}. Such structures arise in a range of multiscale physical models and permit closed-form conditional statistics. CGKN combines this conditional-Gaussian structure with learned latent representations to approximate the joint dynamics of the observations and latent states. Specifically, the observation $\mathbf{y}_{t}$ and the latent representation $\mathbf{z}_{t}=\mathcal{E}_{x}(\mathbf{x}_{t})$ of the physical state are modeled by two distinct but coupled dynamics:
\begin{equation}
\begin{aligned}
    \mathbf{y}_{t+1}
    &=
    \mathcal{M}_{1}(\mathbf{y}_{t},\mathbf{z}_{t})
    =
    \mathbf{d}_{t+1}
    +
    \mathbf{C}_{t+1}\mathbf{z}_{t}
    +
    \boldsymbol{\sigma}_{1}\boldsymbol{\epsilon}_{t}^{y},\\
    \mathbf{z}_{t+1}
    &=
    \mathcal{M}_{2}(\mathbf{y}_{t},\mathbf{z}_{t})
    =
    \mathbf{f}_{t+1}
    +
    \mathbf{K}_{t+1}\mathbf{z}_{t}
    +
    \boldsymbol{\sigma}_{2}\boldsymbol{\epsilon}_{t}^{z}.
\end{aligned}
\label{eq:cgkn-discrete}
\end{equation}
Here, $\mathcal{M}_{1}$ and $\mathcal{M}_{2}$ describe the dynamics of the observation system and the latent state system, respectively. The variables $\boldsymbol{\epsilon}_{t}^{y}$ and $\boldsymbol{\epsilon}_{t}^{z}$ denote Gaussian noise, while $\boldsymbol{\sigma}_{1}$ and $\boldsymbol{\sigma}_{2}$ are the corresponding noise coefficient matrices. The deterministic coefficients $\mathbf{d}_{t+1}$, $\mathbf{C}_{t+1}$, $\mathbf{f}_{t+1}$, and $\mathbf{K}_{t+1}$ are obtained from the current observation through a neural network $\mathcal{G}_{\theta}$:
\begin{equation}
    \left[
        \mathbf{d}_{t+1},
        \mathbf{C}_{t+1},
        \mathbf{f}_{t+1},
        \mathbf{K}_{t+1}
    \right]
    =
    \mathcal{G}_{\theta}(\mathbf{y}_{t}).
\end{equation}
Among these coefficients, $\mathbf{K}_{t+1}$ can be viewed as an observation-dependent Koopman operator. Unlike the previously described methods with fixed linear latent transitions, which use a single state-independent Koopman operator, CGKN conditions $\mathbf{K}_{t+1}$ on the current observation, increasing the expressiveness of the latent transition model.

Together with the observation-dependent term $\mathbf{f}_{t+1}$, the latent transition can be viewed as a generalized form of Koopman with control~\citep{proctor2018generalizing}, with $\mathbf{y}_{t}$ serving as the control input. This structure allows CGKN to represent nonlinear dependence on the observation while retaining conditionally linear dynamics in the latent state. As a result, starting from a Gaussian initial distribution, the latent state distribution remains Gaussian throughout the recursive filtering process:
\begin{equation}
    p(\mathbf{z}_{t}\mid\mathbf{y}_{1:t})
    =
    \mathcal{N}\!\left(
        \mathbf{z}_{t};
        \boldsymbol{\mu}_{t},
        \boldsymbol{\Sigma}_{t}
    \right),
\end{equation}
where $\boldsymbol{\mu}_{t}$ and $\boldsymbol{\Sigma}_{t}$ denote the mean and covariance of the Gaussian latent state distribution. Given $\boldsymbol{\mu}_{t-1}$ and $\boldsymbol{\Sigma}_{t-1}$, CGKN incorporates the current observation $\mathbf{y}_{t}$ through the following closed-form updates:
\begin{equation}
\begin{aligned}
\mathbf{S}_{t}
&=
\boldsymbol{\sigma}_{1}\boldsymbol{\sigma}_{1}^{\top}
+
\mathbf{C}_{t}\boldsymbol{\Sigma}_{t-1}\mathbf{C}_{t}^{\top},
\quad
\mathbf{G}_{t}
=
\mathbf{K}_{t}\boldsymbol{\Sigma}_{t-1}
\mathbf{C}_{t}^{\top}\mathbf{S}_{t}^{-1},
\\
\boldsymbol{\mu}_{t}
&=
\mathbf{f}_{t}
+
\mathbf{K}_{t}\boldsymbol{\mu}_{t-1}
+
\mathbf{G}_{t}
\left(
\mathbf{y}_{t}
-
\mathbf{d}_{t}
-
\mathbf{C}_{t}\boldsymbol{\mu}_{t-1}
\right),
\\
\boldsymbol{\Sigma}_{t}
&=
\mathbf{K}_{t}\boldsymbol{\Sigma}_{t-1}\mathbf{K}_{t}^{\top}
+
\boldsymbol{\sigma}_{2}\boldsymbol{\sigma}_{2}^{\top}
-
\mathbf{G}_{t}\mathbf{S}_{t}\mathbf{G}_{t}^{\top},
\end{aligned}
\label{eq:cgkn-filter}
\end{equation}
where $\mathbf{S}_{t}$ denotes the innovation covariance and $\mathbf{G}_{t}$ denotes the Kalman gain.

To learn the latent representation and the coefficients required by these recursive filtering updates, CGKN uses reconstruction, physical state forecast, latent state forecast, and data assimilation losses:
\begin{equation}
\begin{aligned}
\mathcal{L}_{x}
&=
\sum_{t}
\left[
    \left\|
        \mathcal{M}_{1}(\mathbf{y}_{t},\mathbf{z}_{t})
        -
        \mathbf{y}_{t+1}
    \right\|^{2}
    +
    \left\|
        \mathcal{D}_{x}\!\left(
            \mathcal{M}_{2}(\mathbf{y}_{t},\mathbf{z}_{t})
        \right)
        -
        \mathbf{x}_{t+1}
    \right\|^{2}
\right],
\\
\mathcal{L}_{z}
&=
\sum_{t}
\left\|
    \mathcal{M}_{2}(\mathbf{y}_{t},\mathbf{z}_{t})
    -
    \mathbf{z}_{t+1}
\right\|^{2},
\qquad
\mathcal{L}_{\mathrm{DA}}
=
\sum_{t}
\left\|
    \mathcal{D}_{x}(\boldsymbol{\mu}_{t})
    -
    \mathbf{x}_{t}
\right\|^{2},
\\
\mathcal{L}_{\mathrm{stage1}}
&=
\lambda_{\mathrm{rec}}\mathcal{L}_{\mathrm{rec}}
+
\lambda_{x}\mathcal{L}_{x}
+
\lambda_{z}\mathcal{L}_{z},
\qquad
\mathcal{L}_{\mathrm{stage2}}
=
\mathcal{L}_{\mathrm{stage1}}
+
\lambda_{\mathrm{DA}}\mathcal{L}_{\mathrm{DA}}.
\end{aligned}
\label{eq:cgkn-training-loss}
\end{equation}
CGKN follows a two-stage training procedure. In Stage 1, the encoder $\mathcal{E}_{x}$, decoder $\mathcal{D}_{x}$, and coefficient neural network $\mathcal{G}_{\theta}$ are optimized using $\mathcal{L}_{\mathrm{stage1}}$. The noise coefficient matrices $\boldsymbol{\sigma}_{1}$ and $\boldsymbol{\sigma}_{2}$ are either estimated from one-step forecast residuals or predefined, depending on the data set, as detailed in Appendix~\ref{app:covariance-calibration}. In Stage 2, training uses $\mathcal{L}_{\mathrm{stage2}}$, where $\mathcal{L}_{\mathrm{DA}}$ encourages accurate physical state reconstruction from the filtering mean and $\mathcal{L}_{\mathrm{rec}}$ maintains reconstruction from the encoded latent state. The loss weights are set separately for each training stage. 
DBF and CGKN maintain the latent mean and covariance recursively through filtering during assimilation. The two methods differ in how they incorporate observations: DBF uses an inverse-observation operator to produce a Gaussian latent distribution, whereas CGKN explicitly models the coupled dynamics of the observation and latent state to recursively correct the latent estimate.

\subsection{Summary of Model Properties}

Table~\ref{tab:model_properties} summarizes the structural and assimilation properties of the benchmarked models. DMD, KAE, Tensor-Var, and DBF use fixed linear latent transitions, whereas NL-AE and WL-AE use fixed nonlinear latent transitions. CGKN instead uses an observation-dependent transition that is affine in the latent state. The autoencoder variants evaluate observation residuals through the nonlinear decoder, while DMD and Tensor-Var compare the latent state directly with an observation-derived latent target. DMD constructs this target from the current observation, whereas Tensor-Var incorporates both current and past observations. DBF maps observations to a Gaussian distribution over the latent state through its inverse-observation operator, while CGKN uses coupled observed state and latent state dynamics.

These structural differences also determine whether Gaussianity is preserved under the assumed error model, as indicated in the Gauss. column. DMD and Tensor-Var combine linear latent transitions with linear observation coupling, yielding quadratic objectives corresponding to Gaussian posteriors. In the autoencoder variants, nonlinear decoder-based observation coupling does not generally preserve Gaussianity, even when the latent transition is linear. DBF preserves Gaussianity by combining Gaussian factors during the analysis update, while CGKN preserves it conditionally through its observation-dependent affine dynamics and analysis update.

The methods also differ in whether the observations used for assimilation are incorporated when learning the latent representation and Koopman dynamics. Tensor-Var pretrains its autoencoder on physical states, then fits the Koopman operator to encoded state snapshots before training the inverse-observation encoder. In contrast, DBF learns its model parameters from observation-conditioned analysis posteriors, while CGKN uses coupled observed state and latent state dynamics during training.

\begin{table}[htbp]
\centering
\resizebox{\textwidth}{!}{%
\begin{tabular}{@{}lccccll@{}}
\toprule
Model & Dynamics & Obs. coupling & Gauss. & Uncert. & Inference & Solution \\
\midrule
DMD & Fixed linear & Latent residual & Yes & No & Convex 4D-Var & Closed form \\
NL-AE & Fixed nonlinear & Decoder & No & No & Nonlinear 4D-Var & Iterative \\
WL-AE & Fixed nonlinear & Decoder & No & No & Nonlinear 4D-Var & Iterative \\
KAE & Fixed linear & Decoder & No & No & Nonlinear 4D-Var & Iterative \\
Tensor-Var & Fixed linear & Latent residual$^{*}$ & Yes & No & Convex 4D-Var & Closed form \\
DBF & Fixed linear & Inverse Obs. & Yes & Yes & Gaussian filtering & Closed form \\
CGKN & Cond. affine & Dynamic coupling & Cond. & Yes & Conditional Gaussian filtering & Closed form \\
\bottomrule
\end{tabular}%
}
\caption{Structural comparison of the benchmarked models. Gauss. and Uncert. denote preservation of Gaussianity and recursive uncertainty propagation, respectively. Solution indicates whether assimilation is computed in closed form or through iterative optimization. Cond. denotes a conditional property. $^{*}$The latent target uses both current and past observations.}
\label{tab:model_properties}
\end{table}

\section{Experimental Setup}
\label{sec:experimental-setup}

We evaluate the benchmarked methods on five different settings spanning low-dimensional chaotic dynamics, fluid flow, shallow-water dynamics, and global atmospheric fields at two spatial resolutions. All methods are assessed using consistent data splits, forecast horizons, observation settings, and evaluation metrics. This section describes the benchmark data sets and the common evaluation protocol.

\subsection{Benchmark Data Sets}
\label{sec:benchmark-datasets}

\textit{Lorenz--96.}
\label{sec:exp-lorenz96}
The Lorenz--96 system~\citep{lorenz1996predictability} provides a low-dimensional benchmark for chaotic dynamics and predictability. Its state is governed by
\begin{equation}
    \frac{\mathrm{d}x_i}{\mathrm{d}t}
    =
    \left(x_{i+1}-x_{i-2}\right)x_{i-1}
    -x_i+F,
    \qquad i=1,\ldots,D,
\end{equation}
with periodic indexing, $x_{i+D}=x_i$. We use $D=40$ state variables and forcing $F=8$, for which the system exhibits strongly chaotic dynamics. Trajectories are generated using a fourth-order Runge--Kutta method with a time step of $0.01$ model time units (MTU). After discarding the first $30$ MTU, the states are sampled every $0.05$ MTU, over which the evolution is mildly nonlinear.

\textit{Cylinder.}
The Cylinder benchmark uses the boundary-condition (BC) subset of CFDBench~\citep{luo2024cfdbenchlargescalebenchmarkmachine}, which contains two-dimensional incompressible flow past a circular cylinder with varying inlet velocities. Each state contains the horizontal and vertical velocity components $(u,v)$ on a $64\times64$ grid. To include both the transient flow and the subsequent vortex-shedding dynamics, we use frames 40--200 and sample every five stored frames.

\textit{Shallow water.}
The shallow-water benchmark uses the ShallowWater-2D data set from PDEArena~\citep{gupta2023towards}. The shallow-water equations are solved on a global spherical domain using a spherical-harmonic spectral method and leapfrog time integration~\citep{brandstetter2022clifford}. Each state contains two velocity components $(u,v)$ and pressure $p$ on a $96\times192$ spatial grid with periodicity in longitude. Its dynamics combine advective transport and wave propagation, providing a high-dimensional spatial benchmark with interactions across multiple scales.

\textit{ERA5--Coarse and ERA5--High.}
The ERA5 benchmarks use the ERA5 reanalysis~\citep{hersbach2020era5} from the WeatherBench~2 archive~\citep{rasp2024weatherbench2}. Following the variable selection used in Tensor-Var~\citep{yang2025tensorvar}, we retain 500-hPa geopotential ($Z_{500}$), 850-hPa temperature ($T_{850}$), 600-hPa specific humidity ($q_{600}$), and the zonal and meridional wind components at 850~hPa ($u_{850}$ and $v_{850}$). ERA5--Coarse uses a $5.625^\circ$ ($64\times32$) latitude--longitude grid, whereas ERA5--High uses a $1.5^\circ$ ($240\times120$) grid. The original 6-hourly records are sampled at 24-hour intervals, so each prediction step corresponds to one day. The two settings use the same variables and temporal sampling, allowing the effect of spatial resolution to be examined directly.

\subsection{Evaluation Protocol}
\label{sec:evaluation-protocol}

Table~\ref{tab:dataset-summary} summarizes the data set dimensions, latent dimensions, evaluation settings, and data splits. Details of data set preparation and data splits are provided in Appendix~\ref{app:dataset-preparation}. Prediction length, assimilation window, spinup, and observation interval are reported in model steps. ``Spinup'' denotes the duration of forecast propagation without analysis updates before assimilation begins, and ``Obs. every'' denotes the interval between consecutive observations.

\begin{table}[ht]
\centering
\footnotesize
\setlength{\tabcolsep}{2pt}
\renewcommand{\arraystretch}{1.2}
\begin{tabular*}{\linewidth}{@{\extracolsep{\fill}}lccccccccc@{}}
\toprule
Data set & \shortstack{State\\shape} & \shortstack{Latent\\dimension} & \shortstack{Pred.\\length} & \shortstack{Assim.\\window} & Spinup & \shortstack{Obs.\\every} & Train & Val. & Test \\
\midrule
Lorenz--96    & $1\times40$           & $120^{\dagger}$ & 40 & 5 & 2 & 1 & 128 & 16 & 16 \\
Cylinder      & $2\times64\times64$   & 512            & 32 & 6 & 4 & 2 & 39 & 5 & 5 \\
Shallow water & $3\times96\times192$  & 1152           & 40 & 6 & 4 & 2 & 560 & 168 & 168 \\
ERA5--Coarse  &  $5\times64\times32$   & 1024           & 14 & 3 & 2 & 1 & 1979--2015 & 2016 & 2018 \\
ERA5--High    & $5\times240\times120$ & 5400           & 14 & 3 & 2 & 1 & 1979--2015 & 2016 & 2018 \\
\bottomrule
\end{tabular*}
\par\smallskip
{\raggedright $^{\dagger}$DMD uses a latent dimension of 40 for Lorenz--96.\par}
\caption{Summary of the data sets and evaluation settings used in the main experiments. Train, Val., and Test report trajectory counts for Lorenz--96, Cylinder, and shallow water, and calendar years for the ERA5 data sets.}
\label{tab:dataset-summary}
\end{table}

All state variables are standardized channel-wise by subtracting the mean and dividing by the standard deviation, both computed from the training data. To support a fair comparison, NL-AE, WL-AE, KAE, and Tensor-Var use the same encoder--decoder architecture for each data set. Each neural model is trained independently using three random seeds. The training configurations and loss weights are summarized in Appendices~\ref{app:hyperparameters} and~\ref{app:training-loss-weights}, respectively.

The benchmark focuses on data assimilation rather than standalone forecasting performance. Although the Koopman transition is linear, the number of parameters in a dense Koopman operator scales as $\mathcal{O}(d^2)$, where $d$ is the latent dimension. For high-dimensional systems, we therefore use compact latent representations to keep the assimilation procedure computationally tractable. This choice can limit the physical state information retained in the latent representation and reduce forecast accuracy, particularly for ERA5--High.

The neural models use a 120-dimensional latent representation for Lorenz--96, whose physical state has 40 dimensions. The higher-dimensional representation accommodates additional observables for approximating nonlinear dynamics under a single linear operator. As shown in Appendix~\ref{app:koopman-latent-dimension}, the forecast error continues to decrease beyond 40 latent dimensions, with 120 dimensions producing the most stable predictions among the evaluated settings. DMD uses a 40-dimensional representation because the rank of its snapshot matrix cannot exceed the physical state dimension.

We evaluate assimilation on all test trajectories at observation ratios of 5\%, 15\%, 30\%, and 50\% of the spatial locations. The observation operator $\mathcal{H}_t$ is implemented as a linear sampling mask $\mathbf{H}_t$. The observed locations are selected randomly using a fixed random seed and vary across observation times, except for CGKN, which uses a fixed observed state subset because its coupled dynamics are defined for the same observed variables throughout assimilation. In this paper, Tensor-Var uses the available observations within a 10-frame history window, including the current frame, to construct the observation-derived latent estimate.

To examine the effects of latent representation and observation coupling on assimilation accuracy, we use observations without added measurement noise. In latent data assimilation, the latent representation may not fully capture the true physical state, and the observation coupling may introduce additional errors. The resulting mismatch between the represented physical state and the observations is referred to as representation error~\citep{Janjic2018Representation}. The observation error covariance accounts for this mismatch, while the background error and latent model error covariance matrices describe uncertainty in the initial latent state and latent dynamics, respectively. Their construction and calibration are described in Appendix~\ref{app:covariance-calibration}.

To ensure comparable initial conditions across methods, both DBF and CGKN use the full initial physical state for initialization. DBF uses it in an initial inverse-observation update, whereas CGKN encodes it to initialize the latent state.

\section{Numerical Results and Analysis}
\label{sec:numerical-results}

This section examines the three aspects of latent data assimilation introduced in the Introduction. We first evaluate forecast and assimilation performance across systems and observation ratios, supported by physical-field, spectral, and latent-trajectory comparisons. We then assess whether analysis trajectories remain consistent with Koopman dynamics and the linear evolution of the learned latent representation. Finally, we examine assimilation tractability through Gaussianity diagnostics and convexity analysis, followed by a comparison of computational cost.

\subsection{Autoregressive Prediction}
\label{sec:autoregressive-prediction}

We first evaluate the variational models using autoregressive prediction. Each model is initialized from the same physical state and recursively uses its previous prediction to advance to the next step without assimilating subsequent observations. DBF and CGKN are excluded from this comparison because their implementations are built around observation-based state estimation rather than standalone autoregressive prediction. For each forecast step, RRMSE quantifies relative numerical error, whereas SSIM measures the preservation of spatial structure. We report the mean and standard deviation of both metrics across evaluation trajectories.

\begin{figure}[htbp]
    \centering
    \includegraphics[width=\textwidth]{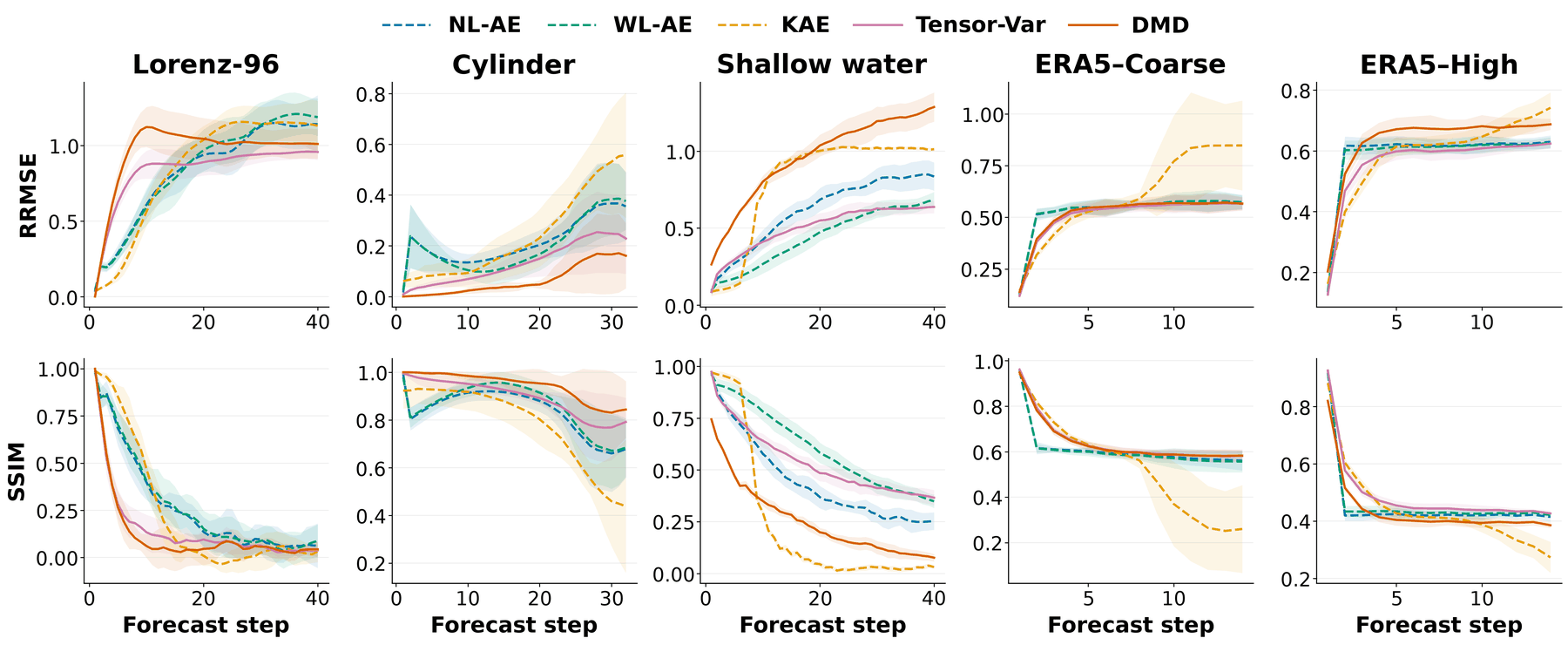}
    \caption{Stepwise forecast performance over the full prediction horizon. The top row shows RRMSE (lower is better) and the bottom row shows SSIM (higher is better) over the forecast steps; shaded bands indicate the standard deviation across evaluation trajectories.}
    \label{fig:forecast-per-step}
\end{figure}
Figure~\ref{fig:forecast-per-step} shows that forecasting difficulty arises differently across the benchmarks. Cylinder provides the simplest setting, where most models accurately propagate the structured and approximately periodic dynamics. The more challenging data sets reveal two distinct failure modes. In Lorenz--96 and shallow water, forecasts are initially accurate but deteriorate substantially during autoregressive prediction, indicating that the main difficulty lies in maintaining accurate propagation over complex dynamics. The ERA5 forecasts instead exhibit relatively large errors from the beginning, which reflect information loss associated with the high compression ratio, particularly at the higher ERA5 resolution. This setting allows us to assess whether data assimilation can improve state estimation despite poor forecast performance.

Compared with the linearly projected latent state used by DMD, the learned latent representations of KAE and Tensor-Var generally yield more accurate forecasts on complex systems, which reflect their greater flexibility in representing the dynamics and supporting physical state reconstruction. Within the autoencoder variants, stronger structural constraints on the latent transition generally improve short-term forecast accuracy, with KAE performing particularly well at early steps. On Lorenz--96, KAE achieves better short-term forecast performance than both DMD and Tensor-Var, suggesting that jointly learning the latent representation and Koopman operator could better capture the short-term evolution of complex dynamics than fitting an operator from consecutive latent state snapshots. Over longer forecast horizons, however, the nonlinear-transition models, such as NL-AE and WL-AE, outperform most Koopman models on several data sets. This contrast is consistent with the simplicity--expressivity trade-off discussed by \citet{cheng2026informationshapeskoopmanrepresentation}: a Koopman representation must retain the information required for physical state reconstruction while supporting evolution under a single linear operator. Satisfying both requirements becomes more challenging as physical state dimensionality and dynamical complexity increase.

These results show that the short-term forecast advantages of a fixed linear latent transition over nonlinear latent transitions do not consistently persist over longer forecast horizons, particularly in strongly chaotic and high-dimensional systems. These limitations motivate the use of data assimilation to incorporate observations and correct the state estimates as forecast errors accumulate.

\subsection{Data Assimilation Performance}
\label{sec:data-assimilation-performance}

To examine the effectiveness of data assimilation, we first evaluate how analysis updates reduce state estimation errors and whether these improvements persist during subsequent forecast propagation. The specification and calibration of error covariances for each method are described in Appendix~\ref{app:covariance-calibration}. Figure~\ref{fig:da-obs30-per-step} shows the stepwise RRMSE and SSIM at a representative observation ratio of 30\%.

\begin{figure}[htbp]
    \centering
    \includegraphics[width=\textwidth]{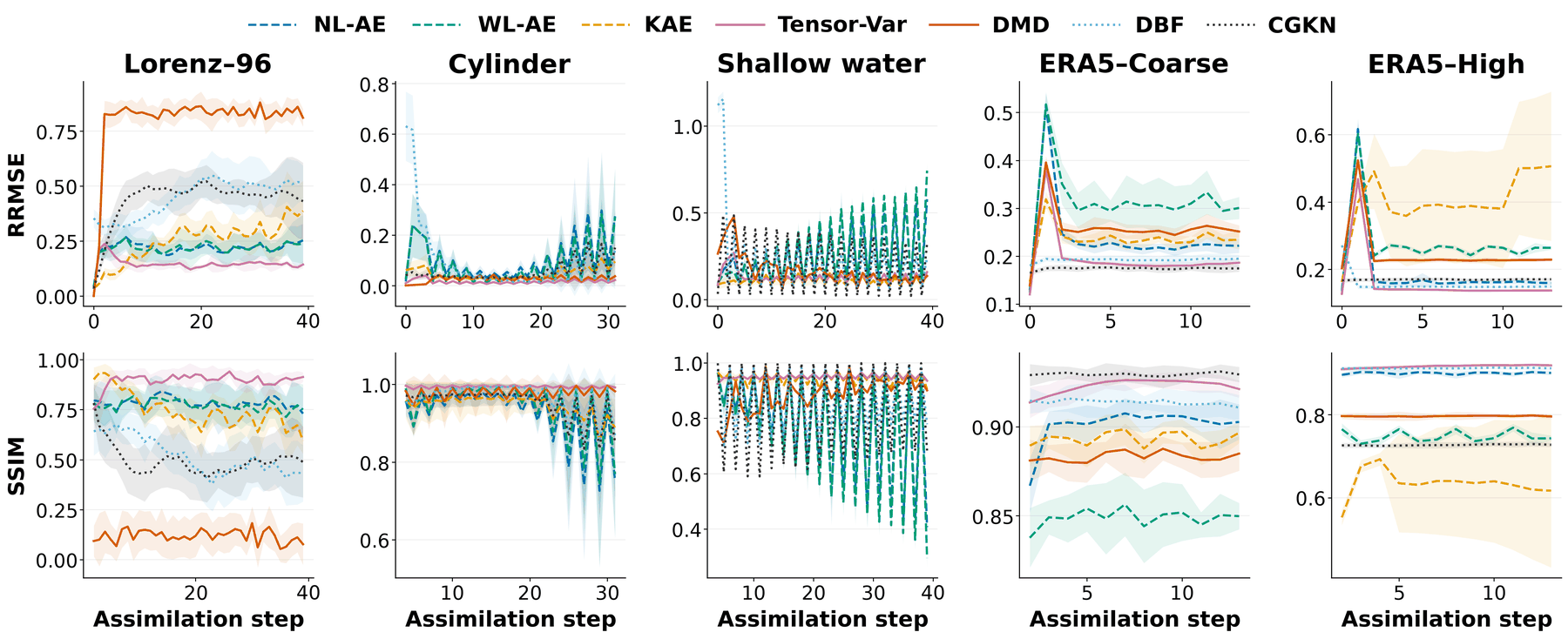}
    \caption{Stepwise data assimilation performance at a 30\% observation ratio. Curves show the mean RRMSE (top) and SSIM (bottom) across evaluation trajectories; shaded bands indicate the corresponding standard deviations.}
    \label{fig:da-obs30-per-step}
\end{figure}

The stepwise curves show that each analysis update generally reduces the current error, but this improvement does not always persist during subsequent forecast propagation. The Koopman-type models with fixed linear latent transitions exhibit relatively stable error evolution, although a globally fixed linear operator has limited expressiveness in representing the strongly chaotic Lorenz--96 dynamics. Among these models, Tensor-Var retains the benefits of analysis corrections more consistently during subsequent propagation. By contrast, the nonlinear-transition models, such as NL-AE and WL-AE, exhibit greater error variability across assimilation cycles, suggesting that their subsequent forecasts are more sensitive to the updated latent state. This behavior contrasts with their strong autoregressive prediction performance. As shown in Figure~\ref{fig:forecast-per-step}, WL-AE can outperform NL-AE during prediction, but this advantage does not consistently lead to better assimilation performance here.

Compared with variational methods, filtering-based methods generally exhibit more controlled error evolution across assimilation cycles. This behavior is consistent with the inclusion of analysis corrections during training in DBF and CGKN. In contrast, the variational methods considered here learn their latent representations and dynamics without directly incorporating analysis corrections into training. Their analysis updates can move the latent state away from the latent dynamical manifold into regions where the learned dynamics may be less accurate during subsequent forecast propagation.

RRMSE and SSIM characterize the overall error magnitude and spatial similarity but do not separately describe the bias and variance of the analysis. Table~\ref{tab:da-nmb-vr} therefore reports the normalized mean bias (NMB) and variance ratio (VR), defined in Appendix~\ref{app:evaluation-metrics}, at the representative 30\% observation ratio. NMB measures the systematic offset from the true state, while VR compares the analysis variance with the true-state variance; values below and above one indicate underestimation and overestimation, respectively.

\begin{table*}[htbp]
    \centering
    \small
    \setlength{\tabcolsep}{5pt}
    \begin{tabular}{@{}lccccc@{}}
        \toprule
        Model & Lorenz--96 & Cylinder & Shallow water & ERA5--Coarse & ERA5--High \\
        \midrule
        NL-AE      & $+0.006 / 0.943$ & $+0.012 / 0.975$ & $-0.001 / 0.920$ & $+0.001 / 0.992$ & $-0.001 / 0.995$ \\
        WL-AE      & $+0.002 / 0.974$ & $+0.003 / 0.982$ & $ 0.000 / 0.929$ & $-0.001 / 1.028$ & $+0.002 / 0.899$ \\
        KAE        & $+0.020 / 1.001$ & $+0.002 / 1.001$ & $ 0.000 / 0.986$ & $-0.001 / 1.007$ & $-0.013 / 0.813$ \\
        Tensor-Var & $-0.006 / 0.861$ & $ 0.000 / 1.002$ & $+0.005 / 0.908$ & $+0.001 / 0.959$ & $-0.003 / 0.988$ \\
        DMD        & $+0.008 / 0.299$ & $+0.001 / 0.997$ & $+0.001 / 0.979$ & $-0.002 / 1.016$ & $-0.001 / 0.967$ \\
        DBF        & $-0.018 / 0.836$ & $-0.005 / 0.993$ & $-0.003 / 0.968$ & $-0.006 / 0.951$ & $-0.002 / 0.979$ \\
        CGKN       & $-0.050 / 0.886$ & $-0.001 / 0.980$ & $-0.001 / 0.999$ & $-0.005 / 0.951$ & $+0.012 / 0.987$ \\
        \bottomrule
    \end{tabular}
    \caption{Normalized mean bias (NMB) and variance ratio (VR) at a 30\% observation ratio. Each cell reports NMB / VR, for which the ideal values are 0 / 1.}
    \label{tab:da-nmb-vr}
\end{table*}

Across most methods and data sets, NMB remains close to zero, showing that assimilation errors arise primarily from inaccurate state reconstruction rather than a uniform shift in the physical state values. VR provides a clearer distinction among the methods. DMD exhibits substantial variance collapse on Lorenz--96, while the reduced VR of KAE on ERA5--High indicates that its analysis retains insufficient physical state variability. In contrast, Tensor-Var and the filtering-based methods preserve the true-state variance more effectively on ERA5--High.

\begin{figure}[htbp]
    \centering
    \includegraphics[width=\textwidth]{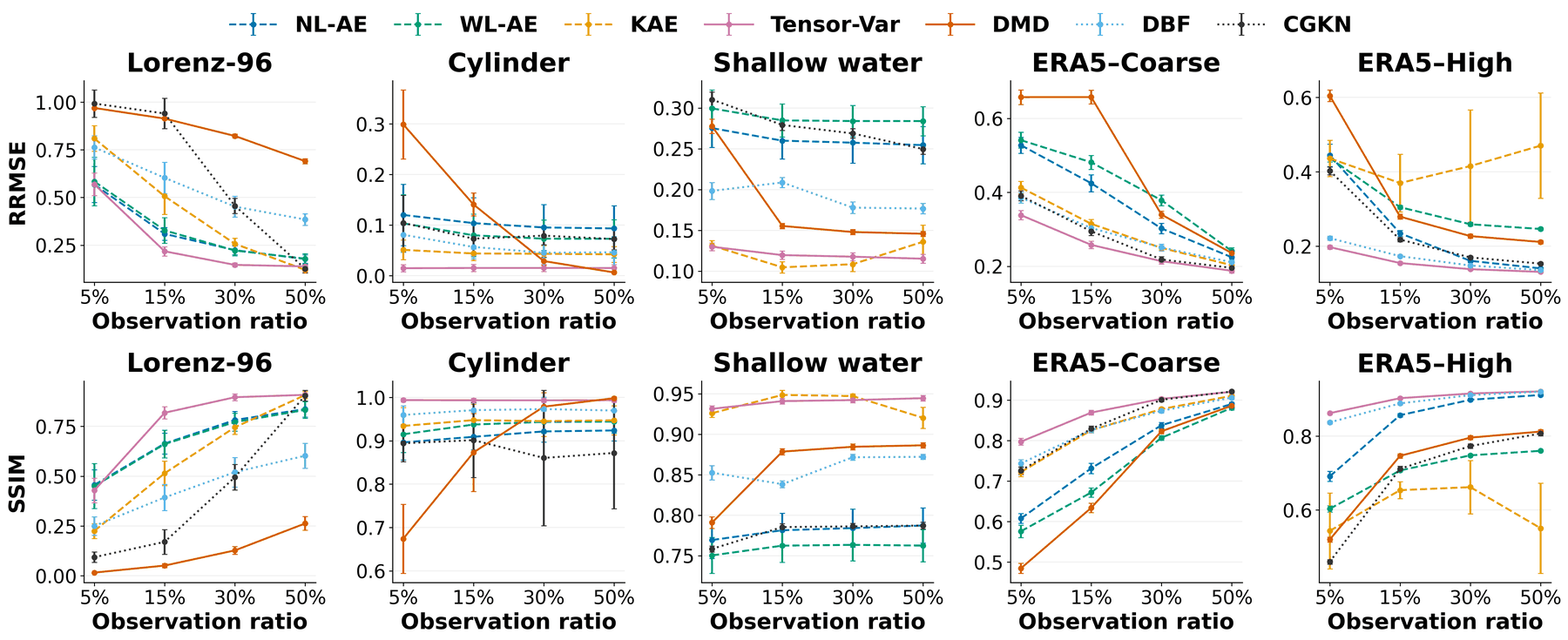}
    \caption{Data assimilation performance across observation ratios. Curves show mean RRMSE (top) and SSIM (bottom); error bars indicate the standard deviation across evaluation trajectories.}
    \label{fig:da-performance}
\end{figure}

We next examine how these behaviors change with observation availability. Figure~\ref{fig:da-performance} reports the mean RRMSE and SSIM across evaluation trajectories at each observation ratio, together with their standard deviations. Performance across training seeds is reported in Appendix~\ref{app:seed-variability}.

Increasing the observation ratio generally improves assimilation accuracy, but the improvement varies across architectures and is not always monotonic. The nonlinear-transition models, such as NL-AE and WL-AE, benefit from additional observations, although their performance saturates at higher ratios on several data sets. For methods with nonlinear decoder-based observation coupling, representation error can further limit these benefits because observations are compared with physical states reconstructed by the decoder. When the learned representation cannot fully recover the true physical state, this comparison is affected by the representation error described in Section~\ref{sec:evaluation-protocol}. Increasing the observation ratio introduces more observational constraints but does not by itself eliminate this representation error.

The effect of representation error becomes more apparent for KAE on the high-dimensional data sets, where assimilation accuracy deteriorates as the observation ratio increases. Its additional linear-dynamics constraint can make it more difficult to satisfy the observational constraints through nonlinear decoder-based observation coupling. Tensor-Var instead maps observations directly to a latent target, allowing additional observations to provide more information for estimating this target without imposing further reconstruction constraints on the decoder. This difference helps explain why Tensor-Var benefits more consistently from higher observation ratios. More generally, methods with closed-form analysis solutions, including Tensor-Var, DBF, and CGKN, tend to benefit from higher observation ratios, although DMD's limited representational flexibility can restrict these benefits. This dependence on the observation ratio is particularly evident for CGKN, which requires sufficient observed state information to condition its latent dynamics. Sparse observations provide insufficient constraints on the coupled dynamics.

\begin{figure}[htbp]
    \centering
    \includegraphics[width=\textwidth]{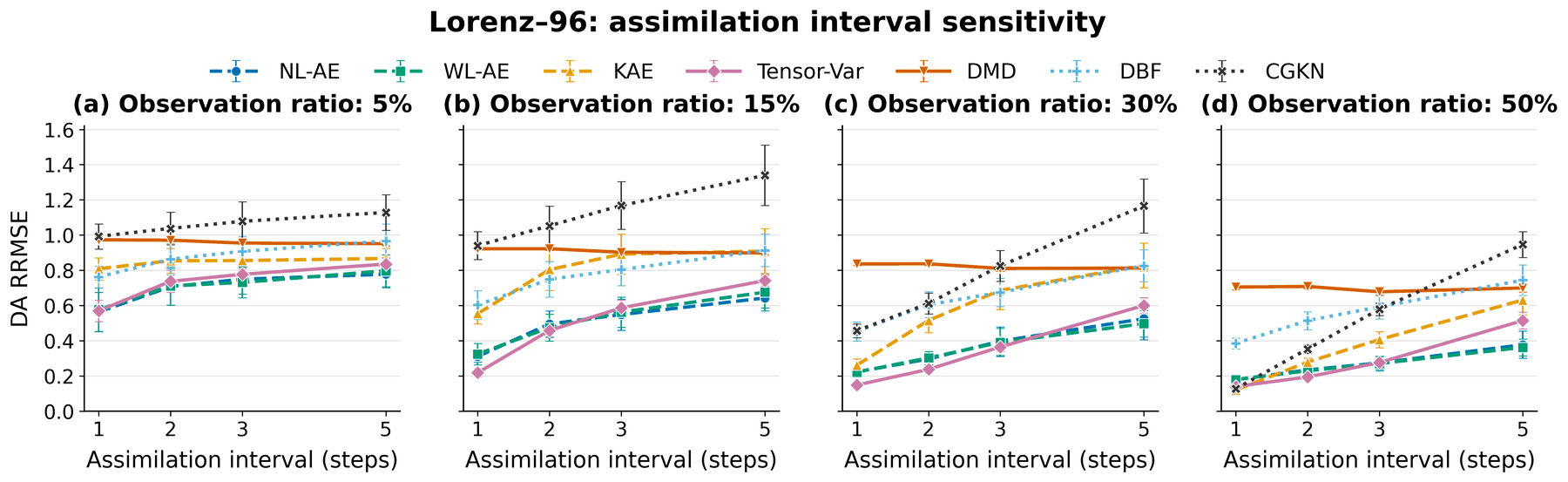}
    \caption{Data assimilation performance on Lorenz--96 across assimilation intervals and observation ratios. Curves show the mean RRMSE, and error bars indicate the standard deviation across evaluation trajectories.}
    \label{fig:lorenz96-assimilation-frequency}
\end{figure}

In the Lorenz--96 experiments, the observation interval of 0.05 time units is short relative to the approximate error-doubling time of 0.4 time units for $F=8$ \citep{lorenz1998optimal}. The short observation interval limits the accumulation of nonlinear forecast errors, allowing a fixed linear latent transition to support accurate assimilation. To examine whether this advantage persists with less frequent observations, we increase the assimilation interval and compare the resulting performance at four observation ratios in Figure~\ref{fig:lorenz96-assimilation-frequency}.

The nonlinear-transition models, such as NL-AE and WL-AE, achieve better assimilation accuracy at longer assimilation intervals, suggesting that a fixed linear latent transition becomes less effective in representing the strongly chaotic dynamics as the time between analysis updates increases. This pattern is evident in Tensor-Var, which achieves the lowest analysis errors in several settings with short intervals but loses this advantage at longer intervals despite incorporating a sequence of observations. For DMD, shortening the assimilation interval does not resolve the underfitting observed above. CGKN is particularly sensitive to the assimilation interval. Between observations, both the observed state and latent state dynamics are used for forecast propagation. Without analysis updates during this period, prediction errors can accumulate in both dynamics, particularly over longer assimilation intervals.

These results show that accurate autoregressive predictions do not consistently yield more accurate analyses, while the benefits of additional observations depend on both the model structure and the assimilation interval. Improvements in analysis accuracy are also not always retained during subsequent forecast propagation.

\subsection{Physical Fields and Spectral Analysis}
\label{sec:physical-fields-spectral-analysis}

To examine how analysis updates recover the physical state from partial observations, we compare representative forecast and analysis fields for ERA5--High, focusing on the reconstruction of spatial structures in a high-dimensional physical state.

\begin{figure*}[htbp]
    \centering
    \includegraphics[width=\textwidth,
    height=0.65\textheight]{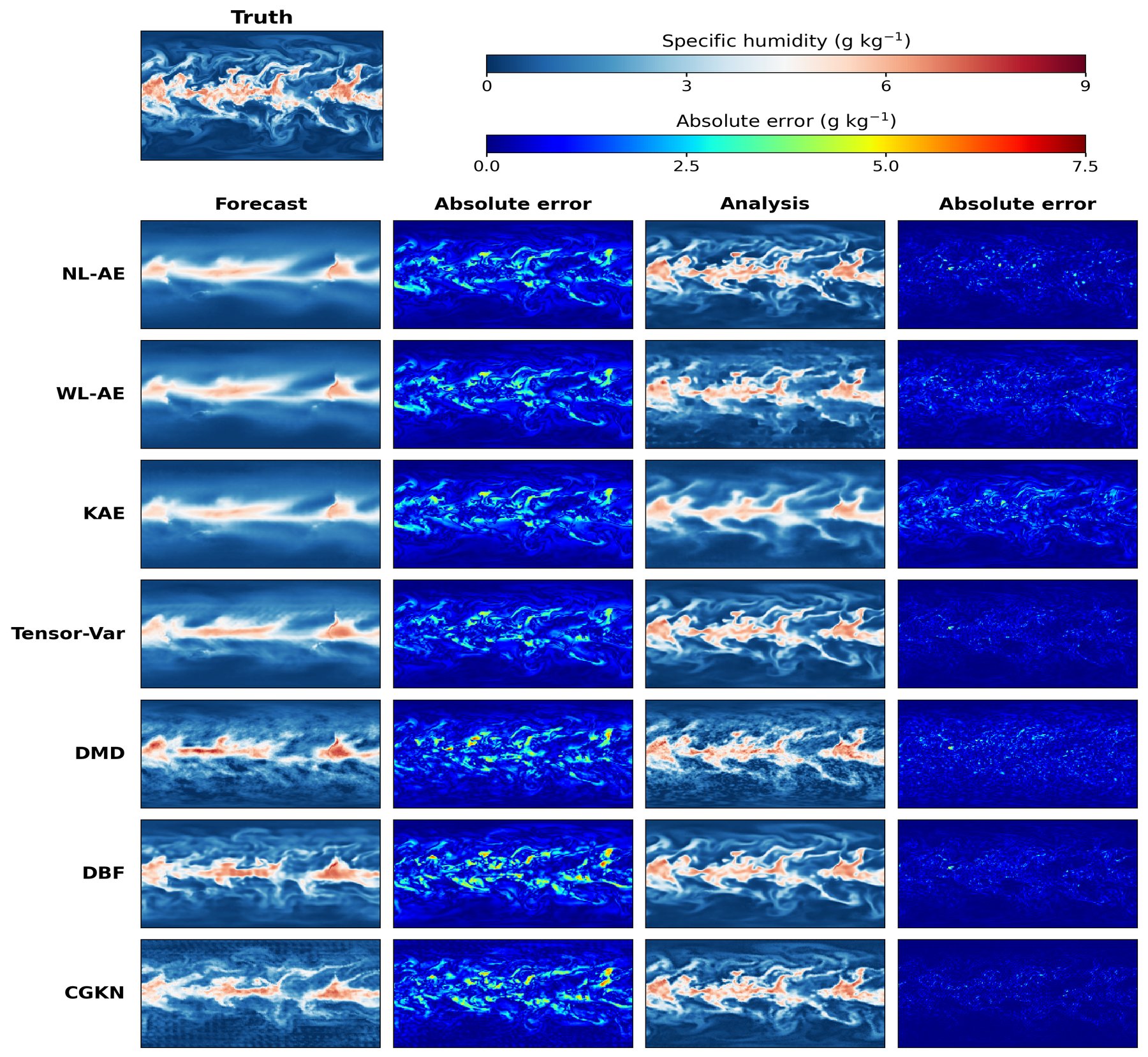}
    \caption{ERA5--High specific humidity fields at 600~hPa, selected as a case whose RRMSE is closest to the test-set median. The forecast and analysis fields are shown together with their absolute errors relative to the true physical state at a 30\% observation ratio.}
    \label{fig:era5-high-fields}
\end{figure*}

Figure~\ref{fig:era5-high-fields} compares the forecast and analysis fields with the true physical state at a 30\% observation ratio. Additional physical-field and spectral comparisons are provided in Appendix~\ref{app:additional-fields-spectra}.

The analysis fields show how partial observations correct the forecast state differently across methods. Among the autoencoder variants, more restrictive constraints on the latent dynamics are associated with more limited analysis corrections. This effect is most apparent for KAE on ERA5--High, where the analysis remains close to the forecast and restores only limited spatial detail. As shown in Table~\ref{tab:da-nmb-vr}, the KAE analysis retains insufficient physical state variability on ERA5--High, consistent with the limited recovery of spatial detail observed here.

The contrast between DMD and Tensor-Var further shows that the effectiveness of a Koopman operator in data assimilation depends on both the information provided by partial observations and the flexibility of the latent representation. On ERA5--High, Tensor-Var restores the dominant spatial structures more effectively, whereas the DMD analysis retains irregular local errors. DMD's observation-derived latent target relies only on the current partial observation and provides limited information about the unobserved state. Tensor-Var incorporates observation history to provide additional temporal information and uses a more flexible learned latent representation to support physical state reconstruction. The effect of observation history is further examined in Appendix~\ref{sec:dmd-tv-history-ablation}. DBF and CGKN also recover the dominant spatial structures, despite large errors in their forecast fields.

\begin{figure*}[htbp]
    \centering
    \includegraphics[width=\textwidth]{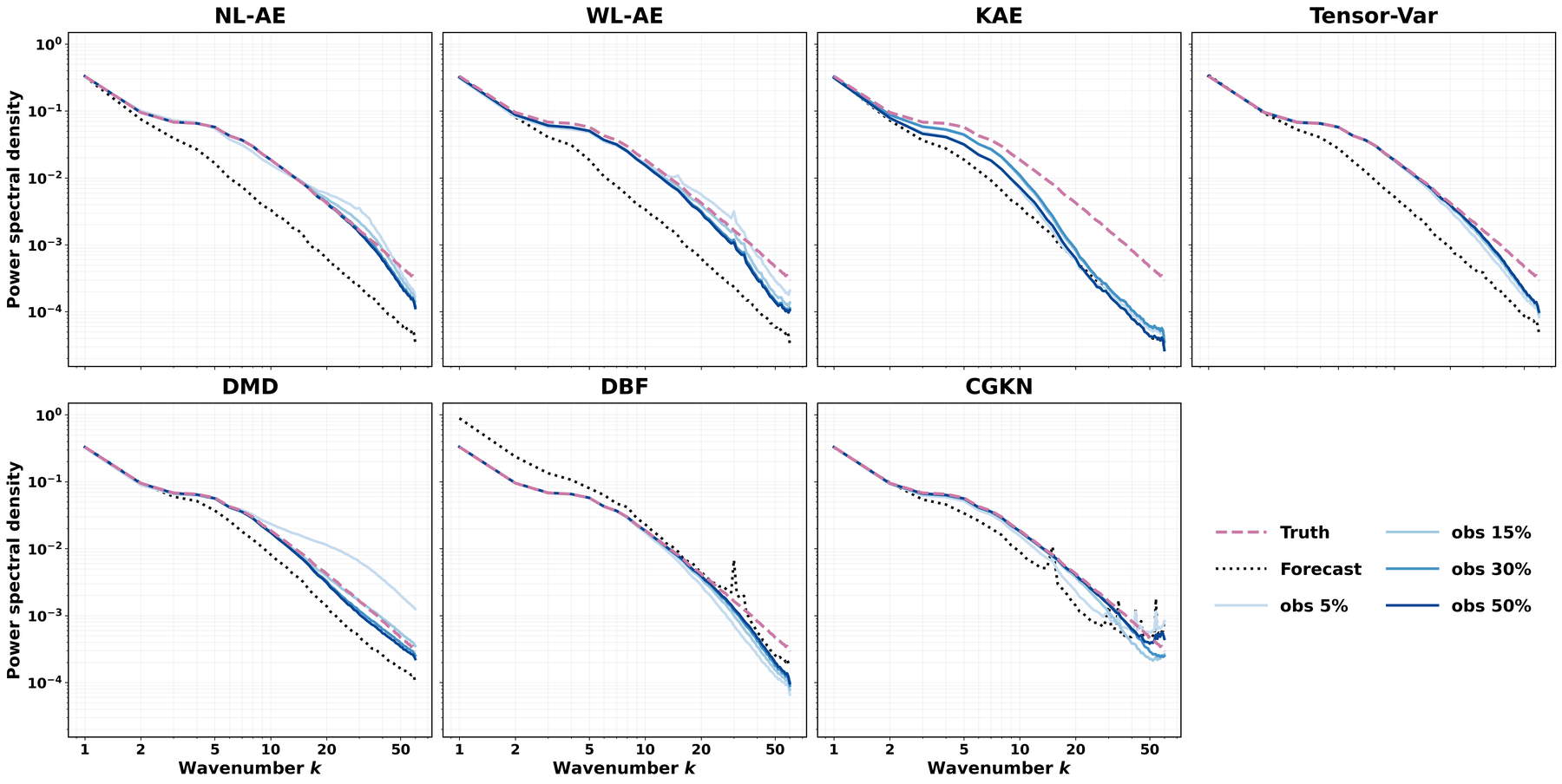}
    \caption{Power spectral density of the ERA5--High specific humidity fields at 600~hPa across observation ratios. Lower wavenumbers represent large-scale spatial structures, whereas higher wavenumbers represent finer-scale variations.}
    \label{fig:era5-high-psd}
\end{figure*}

The physical-field comparisons reveal differences in how the dominant spatial structures and local details are recovered. To quantify how the recovered variability is distributed across spatial scales, we compare the power spectral density (PSD) of the ERA5--High fields at different observation ratios in Figure~\ref{fig:era5-high-psd}. Low wavenumbers correspond to large-scale structures, whereas high wavenumbers correspond to fine-scale variations.

Across methods, the analysis spectra recover low-wavenumber energy more consistently than high-wavenumber energy, indicating more reliable recovery of large-scale structures than fine-scale variations. Increasing the observation ratio does not fully resolve this difference: additional observations generally improve recovery at low and intermediate wavenumbers, while discrepancies persist at the highest wavenumbers. The encoder--decoder can smooth fine-scale variations during physical state reconstruction, leading to underestimated high-wavenumber energy. Under partial observations, large-scale correlations can further suppress these variations by producing overly smooth analysis updates~\citep{Li2015Multiscale}. The additional comparisons in Appendix~\ref{app:additional-fields-spectra} show better high-wavenumber recovery in the other evaluated systems, highlighting the difficulty of recovering fine-scale variations at the higher ERA5 resolution.

Within the autoencoder variants, stronger constraints on the latent transition are associated with more limited analysis corrections at high wavenumbers. KAE shows weaker recovery at intermediate and high wavenumbers, with almost no improvement over the forecast at the highest wavenumbers. This limited spectral recovery is consistent with the physical-field comparison in Figure~\ref{fig:era5-high-fields}, where the KAE analysis remains close to the forecast and fails to recover fine-scale spatial variations. Tensor-Var recovers low- and intermediate-wavenumber energy more accurately than DMD, although discrepancies remain at the highest wavenumbers.

The filtering-based methods show different recovery patterns across spatial scales. DBF consistently recovers low-wavenumber energy across observation ratios but substantially underestimates high-wavenumber energy, suggesting that its Gaussian updates could smooth part of the fine-scale variability. CGKN shows greater improvements as the observation ratio increases, whereas spectral improvements from analysis corrections remain limited at low observation ratios. This dependence on the observation ratio is consistent with the assimilation accuracy trends for CGKN in Figure~\ref{fig:da-performance}.

The physical-field and spectral comparisons show that assimilation recovers dominant spatial structures more consistently than fine-scale variations, particularly at the higher ERA5 resolution. Stronger constraints on the latent transition can limit fine-scale analysis corrections, while higher observation ratios do not consistently improve recovery across all spatial scales.

\subsection{Latent Trajectory Visualization}
\label{sec:latent-trajectory-visualization}

\begin{figure}[htbp]
    \centering
    \includegraphics[width=\textwidth]{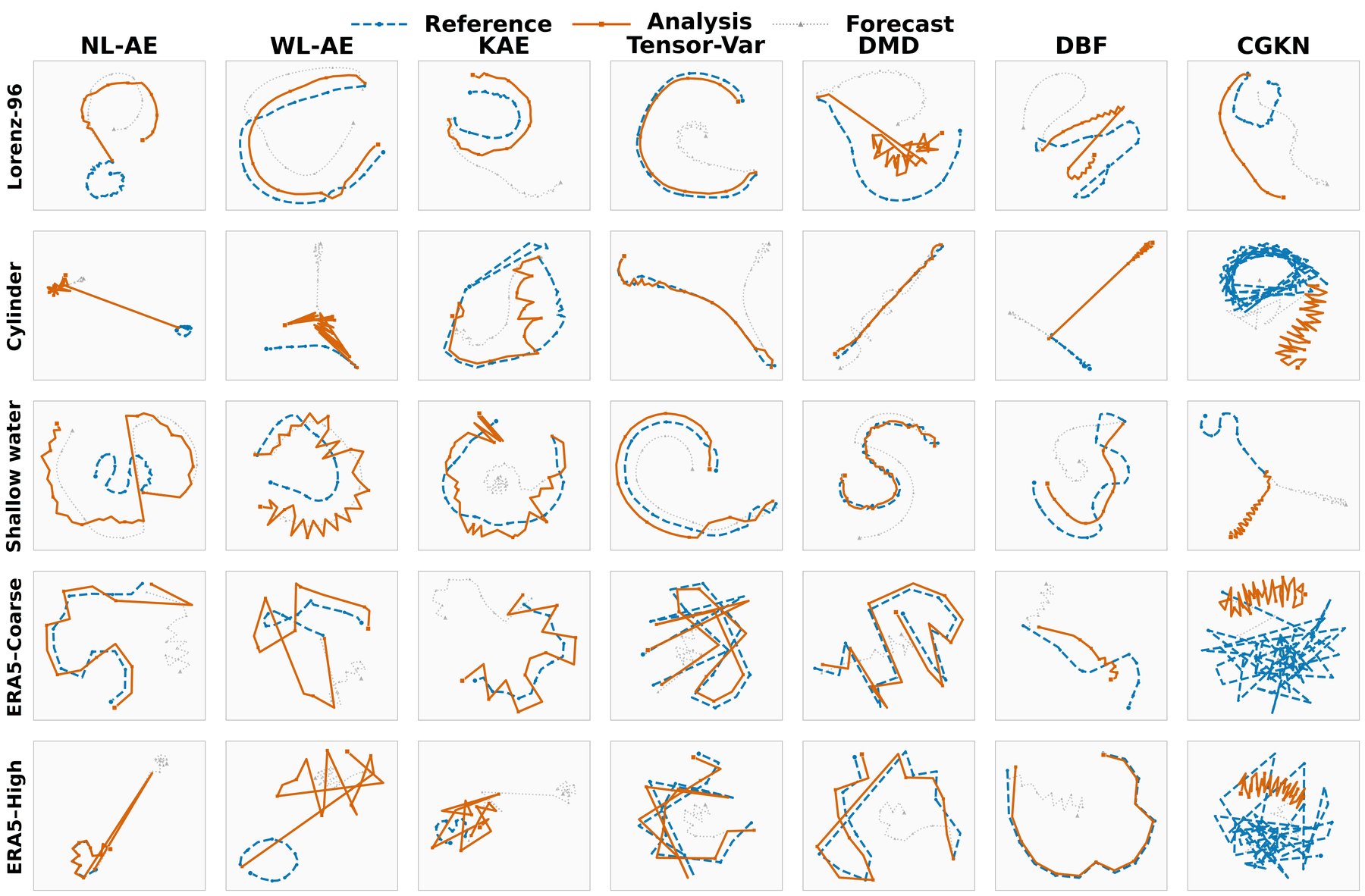}
    \caption{Two-dimensional t-SNE embeddings of the reference, forecast, and analysis latent trajectories. The three trajectories are embedded jointly within each panel.}
    \label{fig:latent-trajectories}
\end{figure}

Having examined the analysis results in physical space, we next investigate how assimilation affects trajectory patterns and evolution in the learned latent space. We compare the forecast and analysis trajectories with a reference trajectory derived from encoding the true physical states. Figure~\ref{fig:latent-trajectories} illustrates their trajectory patterns through two-dimensional t-SNE embeddings~\citep{vanDerMaaten2008}, with the three trajectories embedded jointly within each panel. Table~\ref{tab:latent-reference-distance} reports the median normalized analysis-to-reference distances and their quartiles in each model's original latent space, quantifying alignment with the corresponding reference trajectory. Cross-model differences in these distances also reflect differences in the latent representations. The metric and evaluation procedure are described in Appendix~\ref{app:latent-trajectory-alignment}.

\begin{table}[htbp]
    \centering
    \resizebox{\linewidth}{!}{%
    \begin{tabular}{@{}lccccc@{}}
        \toprule
        Model & Lorenz--96 & Cylinder & Shallow water & ERA5--Coarse & ERA5--High \\
        \midrule
        NL-AE
        & $1.701\;[1.638,\,1.765]$
        & $52.670\;[18.626,\,201.491]$
        & $7.753\;[7.313,\,8.223]$
        & $0.785\;[0.724,\,0.838]$
        & $0.472\;[0.381,\,0.591]$ \\
        WL-AE
        & $0.654\;[0.614,\,0.696]$
        & $0.119\;[0.077,\,0.228]$
        & $1.613\;[1.329,\,1.829]$
        & $0.796\;[0.665,\,0.907]$
        & $6.625\;[6.260,\,7.298]$ \\
        KAE
        & $1.026\;[0.904,\,1.142]$
        & $0.500\;[0.352,\,0.602]$
        & $0.751\;[0.658,\,0.836]$
        & $0.550\;[0.445,\,0.645]$
        & $1.256\;[1.163,\,1.558]$ \\
        Tensor-Var
        & $0.342\;[0.314,\,0.371]$
        & $0.005\;[0.003,\,0.007]$
        & $0.021\;[0.018,\,0.109]$
        & $0.232\;[0.224,\,0.241]$
        & $0.118\;[0.115,\,0.126]$ \\
        DMD
        & $0.848\;[0.802,\,0.891]$
        & $0.022\;[0.017,\,0.029]$
        & $0.033\;[0.028,\,0.043]$
        & $0.460\;[0.437,\,0.481]$
        & $0.167\;[0.162,\,0.170]$ \\
        DBF
        & $0.890\;[0.818,\,0.949]$
        & $0.556\;[0.524,\,0.694]$
        & $0.645\;[0.635,\,0.658]$
        & $0.404\;[0.392,\,0.418]$
        & $0.455\;[0.445,\,0.466]$ \\
        CGKN
        & $1.920\;[1.258,\,2.315]$
        & $3.898\;[3.237,\,4.637]$
        & $1.867\;[1.792,\,1.954]$
        & $1.023\;[0.994,\,1.056]$
        & $1.560\;[1.538,\,1.578]$ \\
        \bottomrule
    \end{tabular}%
    }
    \caption{Normalized analysis-to-reference distance $d_{\mathrm{ref}}$ at a 30\% observation ratio. For each model and data set, distances are normalized by the RMS deviation of the reference latent states from their pooled mean. Each entry reports the median [25th percentile, 75th percentile] over the evaluated trajectory--time pairs after spinup. Smaller values indicate closer alignment within each model's original latent space.}
    \label{tab:latent-reference-distance}
\end{table}

Within the autoencoder variants, stronger constraints on the latent transition can bring the analysis trajectories closer to their respective reference trajectories and produce more similar trajectory patterns, as illustrated in Figure~\ref{fig:latent-trajectories} and quantified in Table~\ref{tab:latent-reference-distance}. However, this benefit does not extend to ERA5--High, where both WL-AE and KAE have larger median analysis-to-reference distances than NL-AE. This suggests that stronger constraints can make the latent transition too restrictive to represent the dynamics of the higher-dimensional physical state, limiting analysis-to-reference alignment.

Beyond the autoencoder variants, Tensor-Var and DMD exhibit analysis trajectory patterns that are more similar to those of their respective reference trajectories, together with smaller analysis-to-reference distances in most comparisons in Table~\ref{tab:latent-reference-distance}. Both methods combine linear latent transitions with linear observation coupling, yielding convex quadratic objectives that remove suboptimal local minima, as examined further in Section~\ref{sec:convexity-experiment}. Despite this shared structure, Tensor-Var has smaller median analysis-to-reference distances than DMD on all evaluated data sets. DMD's limitations are particularly apparent on Lorenz--96, where its analysis trajectory collapses, consistent with the variance collapse observed in its physical state analysis in Section~\ref{sec:data-assimilation-performance}.

The filtering-based methods exhibit different latent-trajectory patterns. DBF produces relatively regular analysis trajectories that differ from the encoded reference trajectories. CGKN instead produces more complex encoded reference trajectories. Its latent transition depends on both the latent state and the corresponding observation, so nearby latent states can evolve differently when conditioned on different observed states. The encoded latent trajectory alone therefore does not fully characterize the coupled dynamics governing the analysis update and subsequent forecast propagation. To examine whether limited observational information contributes to the complex reference-trajectory patterns, we further evaluate CGKN with full observations on ERA5--Coarse in Appendix~\ref{app:cgkn-full-observations}. The encoded reference trajectories exhibit more regular patterns, but the analysis trajectories follow different patterns.

The analysis-to-reference differences observed for DBF and CGKN are consistent with their training objectives, neither of which directly enforces alignment between the analysis mean and the encoded reference state. DBF combines physical state reconstruction with a KL penalty between the analysis posterior and forecast prior, whereas CGKN constrains the observed state and latent state dynamics together with physical state reconstruction, as defined in Eqs.~\eqref{eq:dbf-training-loss} and~\eqref{eq:cgkn-training-loss}, respectively. These objectives can accommodate different latent states that yield similar physical state reconstructions, so accurate reconstruction does not guarantee analysis-to-reference alignment.

These results show that stronger constraints on the latent transition can improve analysis-to-reference alignment and produce more similar trajectory patterns, but these benefits are not consistent across systems, particularly at the higher ERA5 resolution.

\subsection{Consistency with Koopman Evolution}
\label{sec:koopman-invariance-experiment}

The latent-trajectory comparisons show how assimilation changes trajectory patterns and alignment with the encoded reference. This raises a further question: do analysis corrections preserve consistency with Koopman dynamics and the linear evolution of the learned latent representation? To answer this question, we assess dynamical consistency from two complementary perspectives: consistency with the linear evolution of the learned latent representation and, for methods with fixed linear latent transitions, consistency with the model's own Koopman dynamics used for forecast propagation. Both comparisons use the encoded reference trajectories as a baseline. This allows us to assess how assimilation changes dynamical consistency while accounting for the mismatch already present between the learned dynamics and the true system's evolution.

As discussed in Section~\ref{sec:deep-koopman}, an exact finite-dimensional linear representation requires the selected observables to span a Koopman-invariant subspace~\citep{Brunton_2016}. Learned observables do not necessarily satisfy this condition, so the finite-dimensional operator generally only approximates their Koopman evolution~\citep{Lusch2018,otto2019linearly}. To quantify the error in this approximation for the reference and analysis trajectories, we use Residual Dynamic Mode Decomposition (ResDMD)~\citep{colbrook2023residual}. The residual measures a normalized mismatch in the Koopman evolution defined in Eq.~\eqref{eq:koopman-observable-evolution}, allowing us to evaluate how assimilation changes the trajectory's consistency with this evolution. We illustrate the underlying consistency relation schematically and define the analysis-to-reference residual ratio as
\begin{equation}
\operatorname{res}
\sim
\left\|
\mathbf{K}_{\mathrm{eval}}\mathcal{E}_{x}(\mathbf{x}_{t})
-
\mathcal{E}_{x}\!\left(\mathcal{M}(\mathbf{x}_{t})\right)
\right\|,
\qquad
R_{\mathrm{res}}
=
\frac{\operatorname{res}^{\mathrm{a}}}{\operatorname{res}^{\mathrm{ref}}}.
\label{eq:resdmd-ratio-new}
\end{equation}
Here, $\mathbf{K}_{\mathrm{eval}}$ denotes either the separately fitted operator $\widehat{\mathbf{K}}$ or the model's own Koopman operator $\mathbf{K}$. The quantities $\operatorname{res}^{\mathrm{ref}}$ and $\operatorname{res}^{\mathrm{a}}$ are the median spectral residuals of the reference and analysis trajectories, respectively, evaluated using the same latent representation and operator. A ratio below one indicates improved dynamical consistency relative to the reference trajectories, whereas a ratio above one indicates reduced consistency.

\begin{figure*}[htbp]
    \centering
    \includegraphics[width=\textwidth]{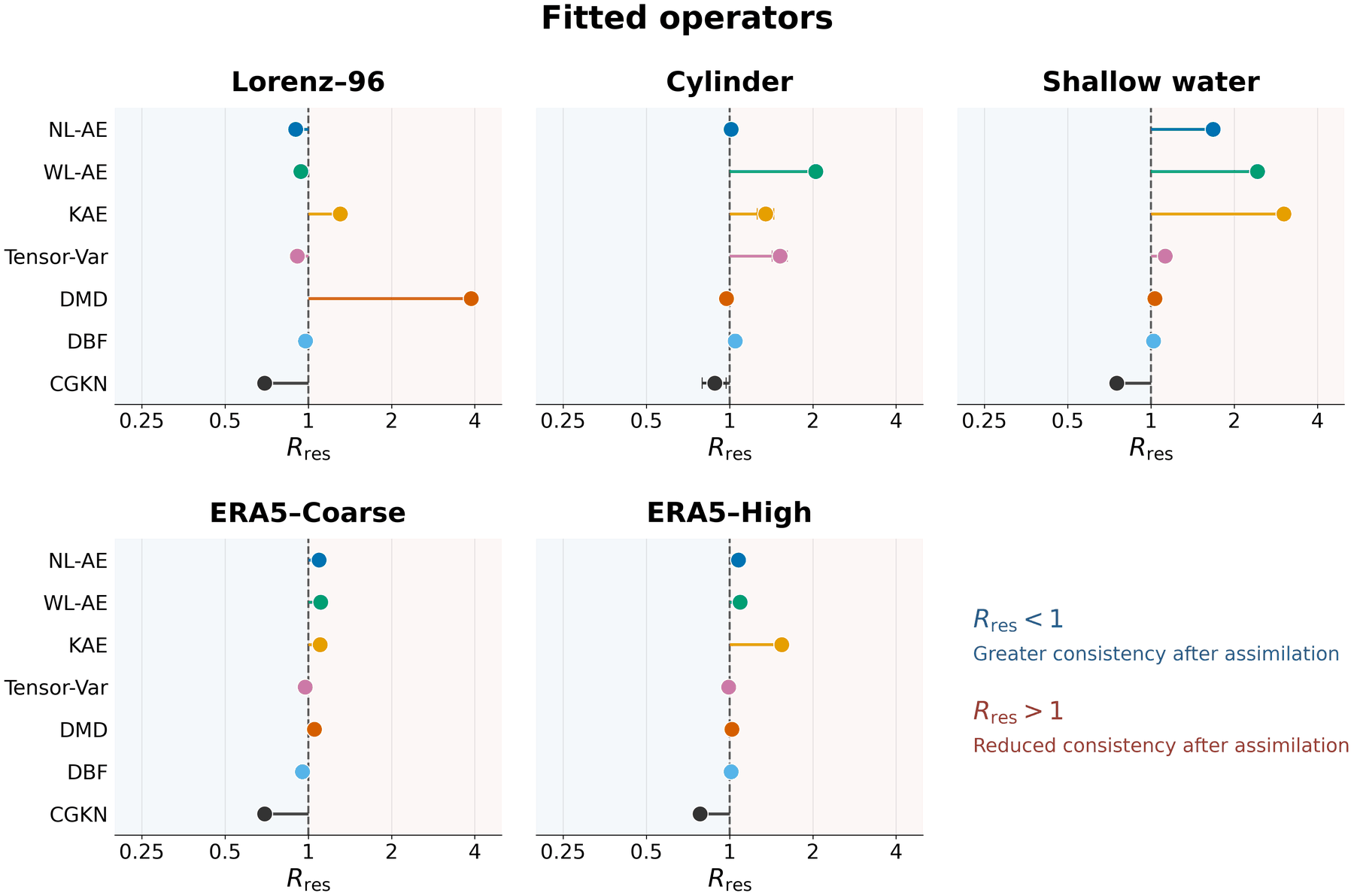}
    \caption{Analysis-to-reference ResDMD residual ratios using data-driven Koopman approximations at a 30\% observation ratio. Points and error bars show the mean and standard deviation of residual ratios obtained by resampling the evaluated trajectory.}
    \label{fig:resdmd-fitted}
\end{figure*}

To characterize the linear evolution of the learned latent representation independently of each model's forecast operator, we fit $\widehat{\mathbf{K}}$ by least squares to consecutive reference latent states constructed from the training data. Figure~\ref{fig:resdmd-fitted} compares the resulting residual ratios across the seven methods at a 30\% observation ratio. Details of the fitting procedure and residual evaluation are provided in Appendix~\ref{app:koopman-invariance-analysis}.

Assimilation generally reduces dynamical consistency, with residual ratios above one in most comparisons. Within the autoencoder variants, stronger constraints on the latent dynamics can further reduce this consistency, particularly on shallow water. KAE also shows a large residual ratio on ERA5--High, indicating that its analysis trajectory deviates further from the linear evolution of the learned latent representation. This is consistent with Figure~\ref{fig:da-obs30-per-step}, where improvements from analysis updates do not consistently persist during subsequent forecast propagation. Stronger constraints on the latent transition alone therefore do not ensure that analysis corrections preserve this linear evolution.

Compared with the autoencoder variants, Tensor-Var and DMD generally show smaller relative residual increases after assimilation. Their linear observation coupling allows observations to constrain the latent state directly, helping analysis corrections remain consistent with the linear evolution of the learned latent trajectories. However, DMD shows a large residual ratio on Lorenz--96, reflecting the limited flexibility of its projected latent representation in describing strongly chaotic dynamics.

The filtering-based methods preserve or improve dynamical consistency after assimilation more often than the variational methods. CGKN has mean residual ratios below one across all evaluated systems, whereas DBF's ratios remain close to one. As previously discussed in Section~\ref{sec:latent-trajectory-visualization}, CGKN's encoded reference trajectory alone does not fully describe its observation-dependent latent dynamics. Its lower residual ratios show that, after incorporating observations, the analysis trajectories more closely follow the linear evolution of the learned latent representation. Both filtering-based methods incorporate analysis updates during training, accounting for how observations modify the latent state and affect subsequent forecast propagation. This can help preserve consistency with the linear evolution of the learned latent trajectories after assimilation.

\begin{figure*}[htbp]
    \centering
    \includegraphics[width=\textwidth]{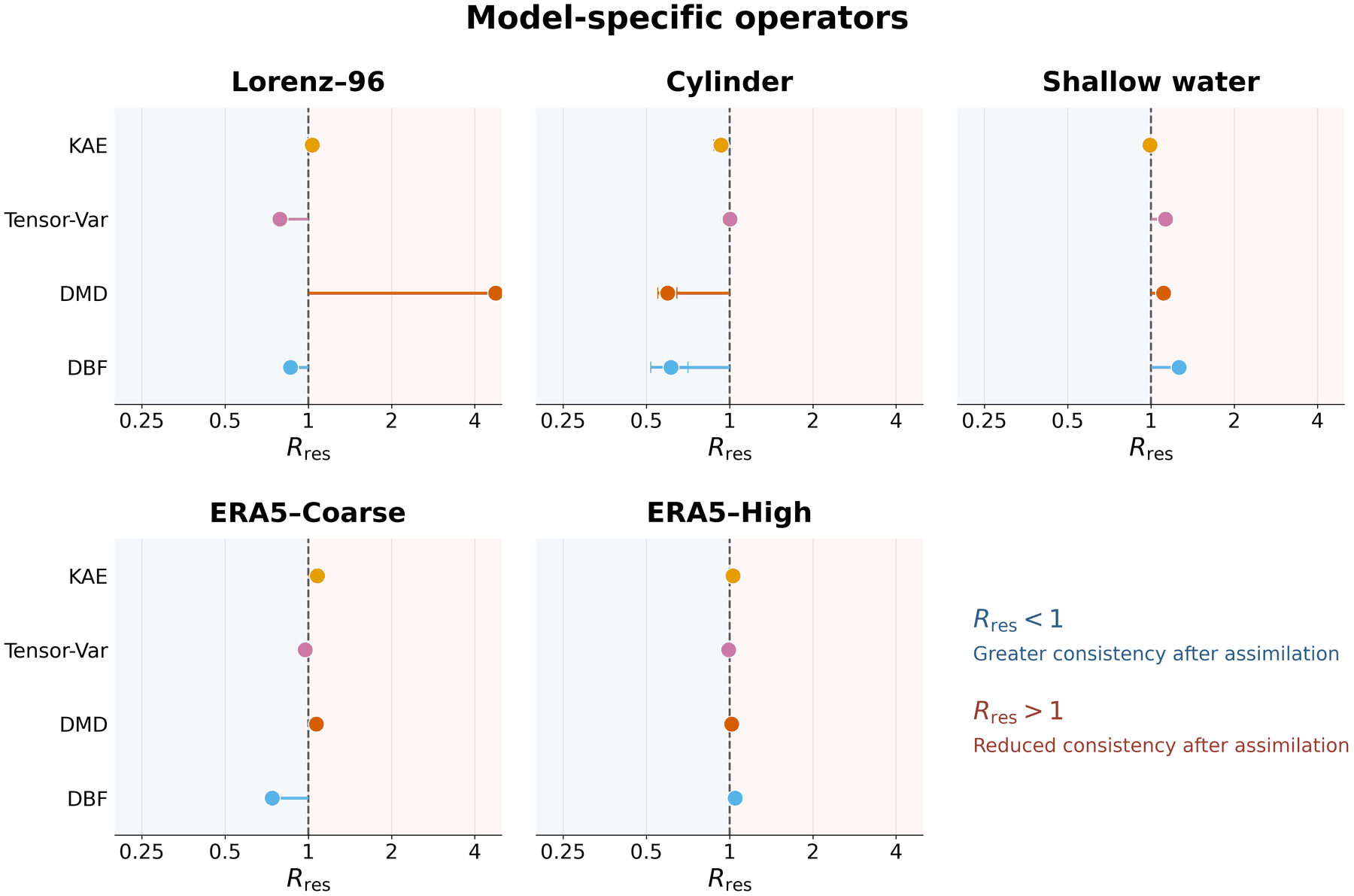}
    \caption{Analysis-to-reference ResDMD residual ratios using model-specific Koopman operators at a 30\% observation ratio. Points and error bars show the mean and standard deviation of residual ratios obtained by resampling the evaluated trajectory.}
    \label{fig:resdmd-model-specific}
\end{figure*}

For methods with fixed linear latent transitions, we also assess whether the analysis trajectories remain consistent with the model's own Koopman dynamics used for subsequent forecast propagation. Figure~\ref{fig:resdmd-model-specific} shows the corresponding residual ratios.

Compared with the fitted-operator results in Figure~\ref{fig:resdmd-fitted}, KAE shows residual ratios closer to one under its own Koopman operator, particularly on shallow water and ERA5--High. DMD, in contrast, shows large residual ratios on Lorenz--96 under both operators, in line with its poor assimilation performance reported earlier. Across the two comparisons, Tensor-Var generally preserves or improves consistency, consistent with the similar analysis and encoded reference trajectory patterns observed in Figure~\ref{fig:latent-trajectories}. DBF shows more data-set-dependent differences, preserving or improving consistency in most settings but reducing it on shallow water under its own Koopman operator. These results complement the latent-trajectory comparisons, where DBF produces relatively regular analysis trajectories that remain distinct from the encoded reference trajectories.

Analysis corrections can improve physical state accuracy while reducing consistency with the learned latent dynamics, providing a possible explanation for why these improvements do not always persist during subsequent forecast propagation. These findings help explain why methods with linear latent transitions exhibit different assimilation performance in strongly chaotic and high-dimensional systems.

\subsection{Gaussianity in the Latent Space}
\label{sec:gaussianity-experiment}
We next assess whether assimilation preserves the Gaussian error model assumed for the latent state in Eq.~\eqref{eq:linear-gaussian-lda}. This property is important because the analysis uncertainty is propagated through the forecast to form the background uncertainty for the next assimilation cycle. If the analysis errors become non-Gaussian, their covariance matrices no longer fully characterize the error distributions, and subsequent Gaussian-based updates as in Eq.~\eqref{eq:kalman-analysis} are no longer exact.

\begin{figure}[htbp]
    \centering
    \includegraphics[width=\textwidth]{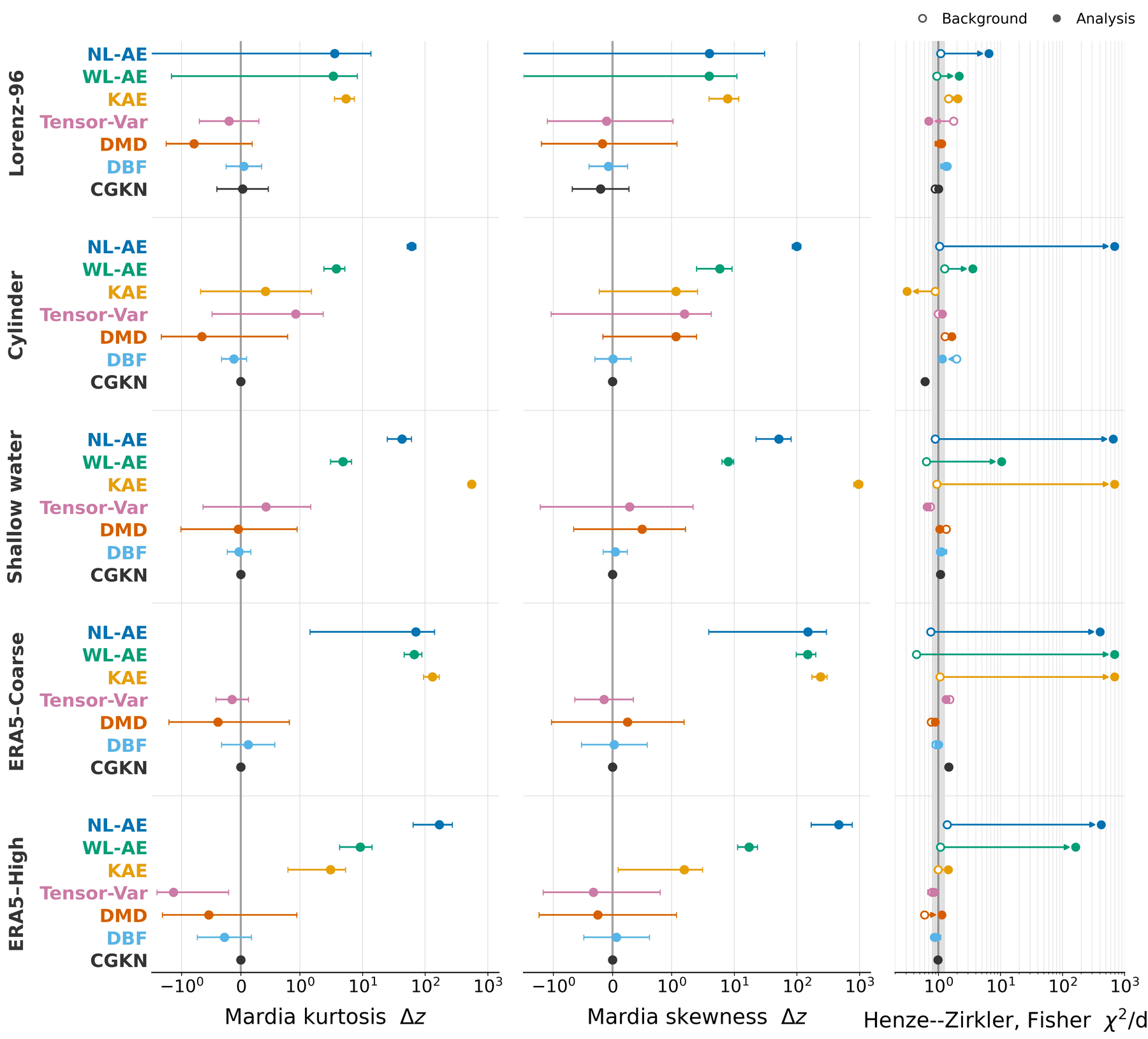}
    \caption{Gaussianity diagnostics in latent space. Left and center: median background-to-analysis changes in standardized Mardia kurtosis and skewness; error bars indicate standard deviations across evaluation starts. Right: Henze--Zirkler Fisher statistics for background and analysis errors.}
    \label{fig:gaussianity}
\end{figure}

Because Gaussianity is a property of the error distribution in DA, it must be assessed using multiple error samples. To obtain these error samples under controlled conditions, we repeat the assimilation using independently sampled Gaussian background errors in latent space while keeping the observation sequence fixed. Each run yields a paired background error and analysis error sample. Figure~\ref{fig:gaussianity} compares the resulting analysis and background error distributions using the Henze--Zirkler test for joint Gaussianity~\citep{henze1990class}, together with Mardia skewness and excess kurtosis~\citep{mardia1970measures} to characterize multivariate asymmetry and heavy- or light-tailed behavior, respectively. The complete sampling procedure and statistical definitions are provided in Appendix~\ref{app:gaussianity-analysis}.

Across the autoencoder variants, the preservation of Gaussianity improves as stronger constraints are imposed on the latent transition. The nonlinear-transition models, such as NL-AE and WL-AE, exhibit larger changes in Mardia skewness and excess kurtosis, particularly for strongly chaotic dynamics. During iterative assimilation, the optimization repeatedly evaluates the nonlinear latent transition and decoder-based observation coupling while updating the latent trajectory. This process creates a highly nonlinear mapping from the initially Gaussian background errors to the analysis errors, producing asymmetric and heavy-tailed distributions. In contrast, KAE shows slightly smaller deviations from Gaussianity on most data sets, consistent with its linear latent transition reducing one source of nonlinear distortion. However, its nonlinear decoder-based observation coupling can still introduce non-Gaussianity.

Compared with iterative variational updates, the closed-form methods produce smaller changes in the Gaussianity diagnostics. For DMD and Tensor-Var, the linear latent transition and observation terms make the analysis solution an affine function of the underlying Gaussian errors. Further details are provided in Appendix~\ref{app:convexity-analysis}. DBF and CGKN instead obtain the analysis distribution through conditional Gaussian updates. Both operations preserve multivariate Gaussianity~\citep{MajumdarMajumdar2019}. Among the filtering methods, DBF exhibits larger data-set-dependent changes because the Gaussian distribution inferred by its learned inverse-observation operator may be less stable. For CGKN, observations affect the latent transition only through its coefficients. For a given observation, the transition remains affine in the latent state and the analysis is obtained through a closed-form Gaussian update, allowing CGKN to preserve Gaussianity more effectively.

A linear latent transition alone does not ensure that analysis errors remain consistent with Gaussian error assumptions. Appropriate observation coupling and analysis updates are therefore needed to maintain consistency with these assumptions.

\subsection{Convexity}
\label{sec:convexity-experiment}

For variational methods, the tractability of the analysis update depends on the convexity of the complete objective. A linear latent transition makes the model error term quadratic, but nonlinear observation coupling can still introduce non-convexity through the observation error term, as in Eq.~\eqref{eq:weak-4d-var}. We first examine the two terms separately to identify the source of non-convexity. We then analyze the complete objective to determine how strongly it constrains the latent-trajectory directions. The theoretical derivation and evaluation procedure are provided in Appendix~\ref{app:convexity-analysis}.

\begin{figure*}[htbp]
    \centering
    \includegraphics[width=\textwidth]{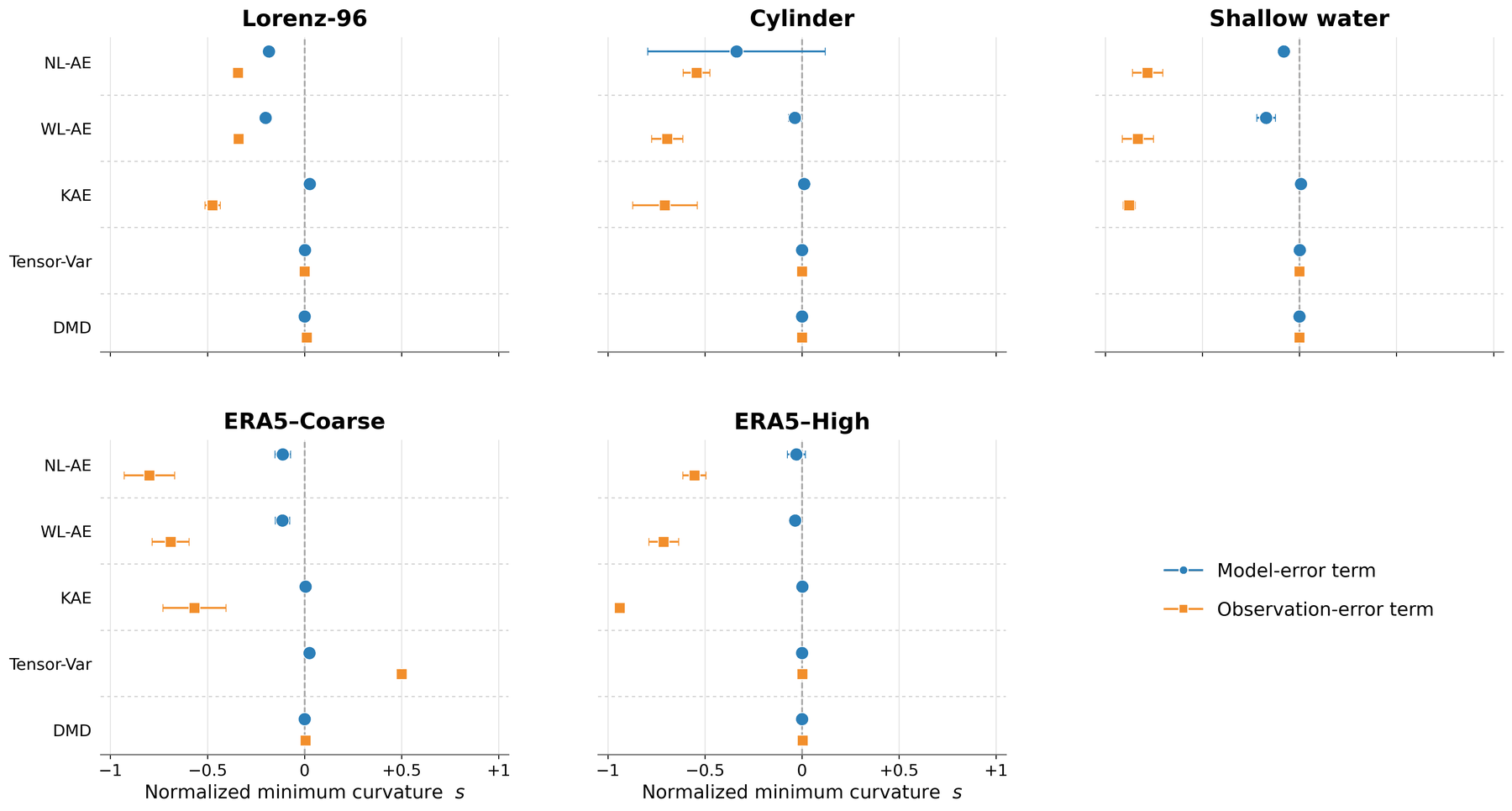}
    \caption{Normalized minimum curvature of the model error and observation error Hessians at a 30\% observation ratio. Points show the mean across evaluation trajectories, and horizontal error bars indicate the corresponding standard deviations. Negative values indicate that the term introduces negative curvature.}
    \label{fig:convexity-terms}
\end{figure*}

To identify the sources of non-convexity, we compare the minimum eigenvalues of the model error and observation error Hessians. Negative values indicate local non-convexity and possible saddle-point behavior that can complicate optimization. Figure~\ref{fig:convexity-terms} compares the normalized minimum eigenvalues of the model error and observation error Hessians at a 30\% observation ratio. 

Across the autoencoder variants, stronger constraints on the latent transition are accompanied by more negative observation term curvature on most benchmarks. This pattern suggests that stronger constraints on the latent dynamics shift more of the nonlinear representation burden toward the decoder, consistent with related discussions in previous work~\citep{udrescu2021symbolic,otto2019linearly}. The observation error term exhibits stronger negative curvature than the model error term and remains the dominant source of non-convexity. Consequently, although KAE removes negative curvature from the model error term, imposing a linear latent transition does not necessarily make the complete assimilation objective more likely to be convex. Tensor-Var and DMD address this remaining nonlinearity through linear observation coupling, yielding convex quadratic objectives by construction, as derived in Appendix~\ref{app:convexity-analysis}.

\begin{figure*}[htbp]
    \centering
    \includegraphics[width=\textwidth]{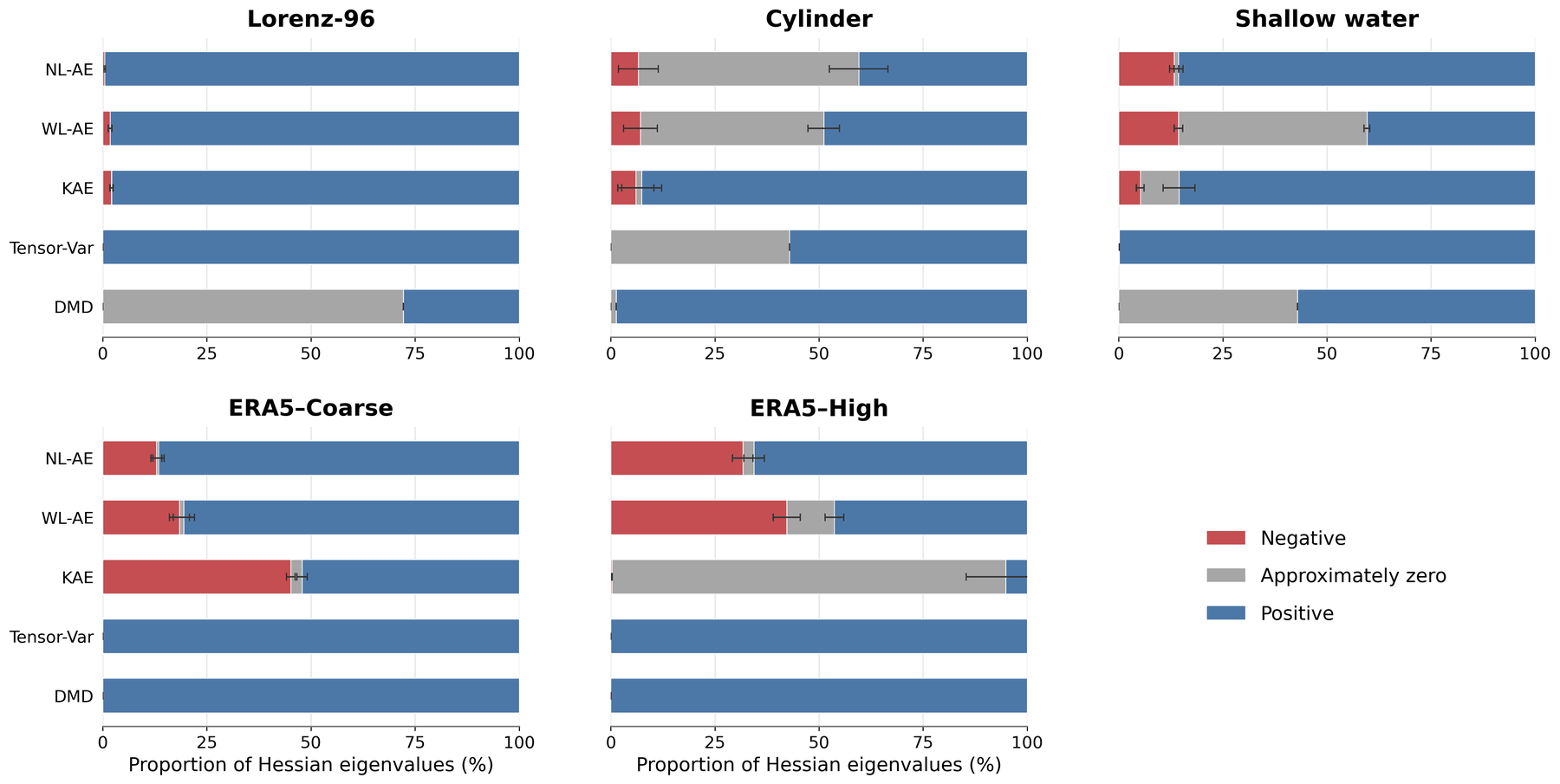}
    \caption{Proportions of negative, approximately zero, and positive eigenvalues of the complete Hessian at a 30\% observation ratio. Stacked bars show the mean proportions across evaluation trajectories, with error bars showing the standard deviations of the segment boundary positions.}
    \label{fig:convexity-spectrum}
\end{figure*}

We next examine how curvature is distributed across the latent-trajectory directions. Figure~\ref{fig:convexity-spectrum} reports the proportions of negative, approximately zero, and positive eigenvalues in the complete Hessian spectrum at a 30\% observation ratio. For DMD and Tensor-Var, the objectives have unique minimizers under positive-definite error covariance assumptions. Their Hessian spectra characterize how strongly different latent-trajectory directions are constrained, with eigenvalues close to zero indicating relatively weak constraints.

The complete Hessian spectra show that increasing the structural constraints on the latent transition does not uniformly improve objective convexity across data sets. Although all autoencoder variants retain negative eigenvalues, stronger constraints increase the proportion of positive eigenvalues for the approximately periodic Cylinder dynamics. This improvement does not extend consistently to strongly chaotic or high-dimensional systems. KAE exhibits a larger proportion of negative Hessian eigenvalues on ERA5--Coarse, while its ERA5--High Hessian spectrum is dominated by approximately zero eigenvalues. DMD also exhibits a large proportion of approximately zero eigenvalues on Lorenz--96. The weak curvature in the latter two settings indicates limited constraints along many latent-trajectory directions, consistent with the limited analysis corrections of KAE and the substantial variance collapse of DMD reported in Section~\ref{sec:data-assimilation-performance}.

Among the methods with convex objectives, Tensor-Var has Hessian eigenvalues above the numerical zero threshold in nearly all latent-trajectory directions on most data sets, indicating positive curvature along these directions. DMD also has predominantly positive eigenvalues above this threshold on Cylinder and both ERA5 settings. Yet it remains less accurate than Tensor-Var, even on Cylinder, where its proportion of positive Hessian eigenvalues is higher. This comparison shows that a higher proportion of positive Hessian eigenvalues does not necessarily translate into more accurate physical state reconstruction.

These experiments show that simplifying the latent transition does not necessarily make the complete objective convex, as decoder-based observation coupling remains a major source of non-convexity. Within convex formulations, the Hessian spectra reveal differences in how strongly the latent trajectory is constrained. However, even when the latent objective admits a unique analysis solution, that solution does not necessarily yield accurate physical state reconstruction.

\subsection{Computational Cost}
\label{sec:computational-cost}

\begin{figure*}[htbp]
    \centering
    \includegraphics[width=\textwidth]{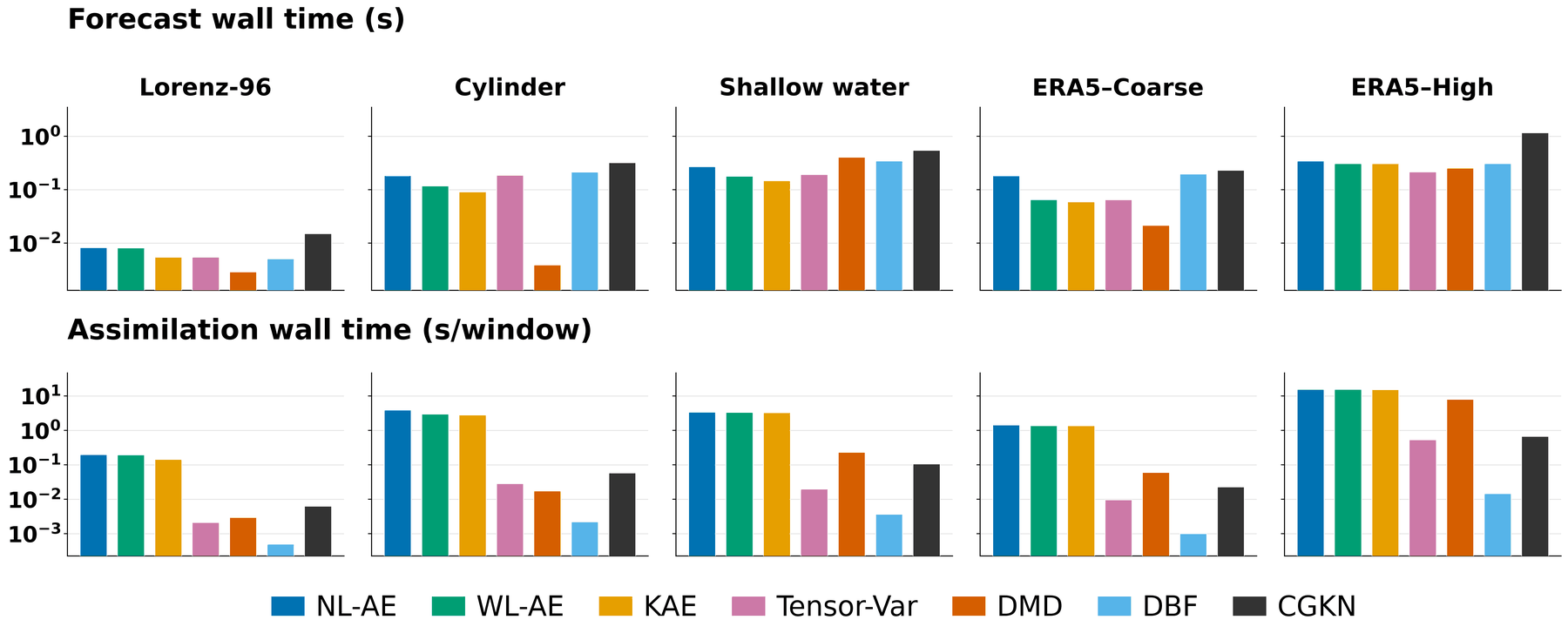}
    \caption{Computational cost of forecasting (top) and data assimilation at a 30\% observation ratio (bottom). Bars show wall time on a logarithmic scale.}
    \label{fig:computational-cost}
\end{figure*}

Repeated forecast--analysis cycles make computational cost an important consideration for practical data assimilation. We therefore conclude the benchmark by comparing the computational cost of forecast propagation and analysis updates. Figure~\ref{fig:computational-cost} reports the total forecast wall time and the wall time per assimilation window at a 30\% observation ratio.

Forecast costs are similar across most methods. CGKN incurs additional costs because it evaluates the coupled observed state and latent state dynamics. DMD is particularly efficient on lower-dimensional systems because its forecasts require only fixed linear operations for propagation and reconstruction. As the physical state and latent dimensions increase, however, these operations involve larger matrices, reducing its computational advantage on higher-dimensional systems.

The models can be divided into three groups according to their assimilation procedure. The autoencoder variants require iterative optimization over the complete latent trajectory and therefore have the longest wall times. In contrast, DMD and Tensor-Var use analytical variational updates that avoid repeated gradient evaluations. DMD becomes more expensive in higher-dimensional systems because the cost of constructing its observation-derived latent target increases with the latent dimension, whereas Tensor-Var obtains this target directly through its learned inverse-observation encoder. The filtering-based methods use recursive filtering updates and are generally less expensive. DBF has the lowest assimilation cost, while CGKN requires additional computation to evaluate the observation-dependent coefficients and propagate the coupled observed state and latent state dynamics.

Even among methods using Koopman operators, computational cost varies substantially with the assimilation and inference procedures. Koopman-type LDA should therefore be designed for the requirements of the target application, balancing assimilation accuracy with computational efficiency and scalability.

\subsection{Discussion}

A linear latent transition simplifies forecast propagation and can improve short-term forecast performance, particularly in approximately periodic systems. These improvements are less consistent in strongly chaotic or high-dimensional systems, where a fixed linear latent transition lacks sufficient expressiveness. The assimilation interval results in Section~\ref{sec:data-assimilation-performance} suggest that frequent analysis updates limit forecast error accumulation, allowing fixed linear latent transitions to remain effective over short intervals. At longer assimilation intervals, the advantages of nonlinear-transition models become more apparent. A Koopman matrix with spectral radius at or below one has no growing modes, so the error growth of a chaotic system cannot be represented within the latent dynamics and can only appear as a residual relative to the encoded trajectory. This connects the assimilation interval results and the ResDMD residuals in Section~\ref{sec:koopman-invariance-experiment} as two views of the same limitation. An observation-dependent latent transition provides another way to increase expressiveness, although its assimilation accuracy remains sensitive to both the amount and temporal availability of observational information.

Linear latent dynamics alone do not ensure consistency with Gaussian error assumptions or convexity of the assimilation objective. More specifically, stronger constraints on the latent transition may shift nonlinearity toward the decoder, leaving nonlinear observation coupling to limit these advantages during analysis updates. Convex variational formulations and conditional Gaussian filtering show that tractable assimilation depends on the structure of the complete analysis update. However, satisfying Gaussian error assumptions or obtaining a unique solution does not by itself ensure accurate physical state reconstruction. The DMD and Tensor-Var comparison illustrates this distinction: stronger constraints on the latent trajectory do not necessarily yield higher assimilation accuracy. Improving the statistical and optimization properties of assimilation must therefore be accompanied by a flexible latent representation and sufficient observational information to support accurate recovery of the physical state.

The consistency experiments show that analysis corrections can improve physical state accuracy while reducing consistency with Koopman dynamics and the linear evolution of the learned latent representation. This provides a possible explanation for why improvements in the current analysis are not always retained during subsequent forecast propagation. At the same time, the learned dynamics are themselves approximate, so preserving consistency does not by itself establish that the corrected trajectory is accurate.

\section{Conclusion and Future Work}
\label{sec:discussion-future-work}

In this work, we proposed a unified evaluation framework to examine how Koopman representations affect latent data assimilation. Our results show that Koopman representations can make latent data assimilation more tractable when combined with appropriate observation coupling and assimilation procedures. A linear latent transition alone does not ensure objective convexity or consistency with Gaussian error assumptions. Even when the latent objective is convex and has a unique minimizer, accurate physical state reconstruction still requires a latent representation that retains the relevant physical state information and observations that effectively constrain the latent state. Although analysis corrections can improve the current state estimate, they may reduce consistency with Koopman dynamics and the linear evolution of the learned latent representation. The improvement in analysis accuracy does not always persist during subsequent forecast propagation.

These conclusions are subject to several limitations. Beyond the controlled comparisons among the autoencoder variants, differences in latent representations, observation coupling, and training objectives limit the attribution of performance differences to individual structural choices. The Hessian analysis of nonlinear variational objectives characterizes local curvature and does not establish global convexity. The Gaussianity diagnostics rely on finite background and analysis error samples and can therefore be sensitive to sample size.

Future work should investigate how Koopman representations can be learned to support the complete forecast--analysis cycle. A central challenge is to combine analysis updates with desirable statistical and optimization properties with a latent representation that supports both accurate physical state reconstruction and subsequent propagation of corrected states. Possible directions include training with analysis states and incorporating dynamical consistency into the assimilation objective while retaining sufficient flexibility for state correction. These approaches should be evaluated across a broader range of observation intervals, noise levels, and non-Gaussian uncertainty. Such investigations could guide the joint design of the latent representation, latent transition, observation coupling, and assimilation procedure.

\section*{Data and Code Availability}

The code, data and pretrained models used in this study are provided as open-access resources: the code is available at \url{https://github.com/KZHIWEI/KoopmanTypeLatentDataAssimilation}, and the data and pretrained models at \url{https://huggingface.co/KZHIwEI/koopmanda-release}.

\acks{S.C. acknowledges the support of Hi!Paris and French state funding managed by ANR under the France 2030
program, reference ANR-23-IACL-0005. N.C., P. B., and C.M. are partially supported by the University of Wisconsin-Madison Office of the Vice Chancellor for Research, with funding from the Wisconsin Alumni Research Foundation. 
The authors used OpenAI's GPT-5.6 Sol to assist with language editing of the manuscript. The authors have reviewed the content of this work and take full responsibility for it.
}

\begin{table}[htbp]
\centering
\small
\setlength{\fboxsep}{8pt}
\setlength{\fboxrule}{0.6pt}
\fbox{%
\begin{minipage}{0.95\linewidth}
\textbf{Main Notations}

\vspace{0.6em}
\renewcommand{\arraystretch}{1.15}
\begin{tabular}{@{}p{0.20\linewidth}p{0.72\linewidth}@{}}

\multicolumn{2}{@{}l}{\textit{States, observations, and latent variables}} \\[0.2em]

$\mathbf{x}_{t}$ 
& physical state at time $t$ \\

$\hat{\mathbf{x}}_{t}$ 
& reconstructed, forecast, or analysis physical state at time $t$ \\

$\mathbf{z}_{t}$ 
& latent state at time $t$ \\

$\mathbf{x}_{0:T}$, $\mathbf{z}_{0:T}$ 
& physical and latent trajectories over the assimilation window \\

$\mathbf{y}_{t}$ 
& observation at time $t$ \\

$\mathcal{T}$ 
& set of observation-time indices \\

$\mathbf{x}_{0}^{\mathrm{b}}$, $\mathbf{z}_{0}^{\mathrm{b}}$ 
& background state in the physical and latent spaces \\[0.4em]

\multicolumn{2}{@{}l}{\textit{Mappings and dynamical models}} \\[0.2em]

$\mathcal{E}_{x}$, $\mathcal{E}_{y}$
& encoders for the physical state and observation \\

$\mathcal{D}_{x}$ 
& decoder from latent space to physical space \\

$\mathcal{H}_t$
& time-dependent observation operator \\

$\mathbf{H}_t$
& linear sampling mask used in the benchmark \\

$\widetilde{\mathbf{H}}_t$
& linear latent-space observation operator \\

$\mathcal{M}_{z}$ 
& latent dynamical model \\

$\mathcal{G}_{\theta}$ 
& nonlinear neural network \\

$\mathcal{K}$ 
& theoretical Koopman operator \\

$\mathbf{K}$ 
& finite-dimensional Koopman matrix defining the linear latent transition \\

$\mathcal{F}_{d}$
& finite-dimensional subspace of observable functions \\[0.4em]

\multicolumn{2}{@{}l}{\textit{Assimilation and uncertainty quantities}} \\[0.2em]

$\mathbf{B}$ 
& latent background error covariance matrix \\

$\mathbf{Q}_{z}$ 
& latent model error covariance matrix \\

$\mathbf{R}$ 
& observation error covariance matrix \\

$\boldsymbol{\mu}_{t}^{f}$, $\boldsymbol{\Sigma}_{t}^{f}$ 
& forecast mean and covariance of the latent state \\

$\boldsymbol{\mu}_{t}^{\mathrm{a}}$, $\boldsymbol{\Sigma}_{t}^{\mathrm{a}}$ 
& analysis mean and covariance of the latent state \\

$J(\mathbf{z}_{0:T})$ 
& weak-constraint latent 4D-Var objective function \\

\end{tabular}
\end{minipage}%
}
\caption{Main notations used in this work.}
\label{tab:main-notations}
\end{table}

\appendix

\section{Model Derivations}
\label{app:model-details}

This appendix provides supporting derivations for the formulations presented in the main text.

\subsection{Dynamic Mode Decomposition (DMD)}
\label{app:dmd}

Dynamic Mode Decomposition (DMD) provides a finite-dimensional linear approximation of the Koopman operator directly from state snapshots, without a neural encoder--decoder~\citep{tu2014dynamic,kutz2016dynamic,rowley2009spectral}. The paired snapshots $\mathbf{x}_{0},\ldots,\mathbf{x}_{n+1}$ define the snapshot matrices and the full DMD operator:
\begin{equation}
\begin{gathered}
    \mathbf{X}=[\mathbf{x}_{0},\ldots,\mathbf{x}_{n}],
    \qquad
    \mathbf{X}^{+}=[\mathbf{x}_{1},\ldots,\mathbf{x}_{n+1}],
    \\
    \mathbf{X}^{+}\approx\mathbf{M}\mathbf{X},
    \qquad
    \mathbf{M}=\mathbf{X}^{+}\mathbf{X}^{\dagger},
\end{gathered}
\end{equation}
where $\mathbf{X}^{\dagger}$ is the Moore--Penrose pseudoinverse. A rank-$r$ truncated SVD defines the reduced representation and reconstruction:
\begin{equation}
    \mathbf{X}\approx
    \mathbf{U}_{r}\boldsymbol{\Sigma}_{r}\mathbf{V}_{r}^{\top},
    \qquad
    \mathbf{z}_{t}=\mathbf{U}_{r}^{\top}\mathbf{x}_{t},
    \qquad
    \widehat{\mathbf{x}}_{t}=\mathbf{U}_{r}\mathbf{z}_{t}.
\end{equation}
Projecting the snapshot dynamics onto the same subspace gives the reduced Koopman approximation and its one-step dynamics:
\begin{equation}
\begin{gathered}
    \mathbf{K}
    =
    \mathbf{U}_{r}^{\top}\mathbf{X}^{+}
    \mathbf{V}_{r}\boldsymbol{\Sigma}_{r}^{-1},
    \\
    \mathbf{z}_{t+1}
    =
    \mathbf{K}\mathbf{z}_{t}+\boldsymbol{\epsilon}_{t}^{z},
    \qquad
    \boldsymbol{\epsilon}_{t}^{z}
    \sim\mathcal{N}(\mathbf{0},\mathbf{Q}_{z}).
\end{gathered}
\end{equation}
The same reduced basis is also used to incorporate partial observations. At each observation time, let $\mathbf{U}_{o,t}$ contain the rows of $\mathbf{U}_{r}$ corresponding to the observed state components, such that $\mathbf{y}_{t}\approx\mathbf{U}_{o,t}\mathbf{z}_{t}$. The observation-derived latent target is then obtained through ridge regression:
\begin{equation}
    \widehat{\mathbf{z}}_{t}
    =
    \left(
        \mathbf{U}_{o,t}^{\top}\mathbf{U}_{o,t}
        +
        \lambda_{\mathrm{DMD}}
        \frac{
            \operatorname{tr}\!\left(
                \mathbf{U}_{o,t}^{\top}\mathbf{U}_{o,t}
            \right)
        }{r}
        \mathbf{I}
    \right)^{-1}
    \mathbf{U}_{o,t}^{\top}\mathbf{y}_{t},
\end{equation}
where $r$ is the reduced dimension and $\lambda_{\mathrm{DMD}}$ is the ridge regularization parameter. Because a partial observation does not sufficiently constrain all latent components, the ridge term stabilizes the inverse projection. Once computed, the resulting target is held fixed and assimilated directly in the reduced space:
\begin{equation}
\begin{aligned}
    J_{\mathrm{DMD}}(\mathbf{z}_{0:T})
    & =
    \frac{1}{2}
    \left\|
        \mathbf{z}_{0}-\mathbf{z}_{0}^{\mathrm{b}}
    \right\|_{\mathbf{B}^{-1}}^{2}
    +
    \frac{1}{2}
    \sum_{t=1}^{T}
    \left\|
        \mathbf{z}_{t}
        -
        \mathbf{K}\mathbf{z}_{t-1}
    \right\|_{\mathbf{Q}_{z}^{-1}}^{2}
    +
    \frac{1}{2}
    \sum_{t\in\mathcal{T}}
    \left\|
        \mathbf{z}_{t}
        -
        \widehat{\mathbf{z}}_{t}
    \right\|_{\mathbf{R}_{z}^{-1}}^{2}.
\end{aligned}
\end{equation}
Here, the background state is projected as $\mathbf{z}_{0}^{\mathrm{b}}=\mathbf{U}_{r}^{\top}\mathbf{x}_{0}^{\mathrm{b}}$, while $\mathbf{B}$ and $\mathbf{R}_{z}$ denote the background error and observation error covariance matrices in the reduced space, respectively. Because the reduced transition is linear and $\widehat{\mathbf{z}}_{t}$ is fixed during assimilation, the resulting objective is convex quadratic in $\mathbf{z}_{0:T}$ when the covariance matrices define positive-definite weights.

\subsection{Closed-Form Weak-Constraint 4D-Var Solutions}
\label{app:closed-form-4dvar}

DMD and Tensor-Var use different forms of observation information, but in both methods, the model and observation residuals depend linearly on the latent trajectory. Their weak-constraint 4D-Var objectives can therefore be expressed as weighted linear least-squares problems. Write the latent trajectory as the stacked vector
\begin{equation}
    \mathbf{z}_{0:T}
    =
    \left[
        \mathbf{z}_{0}^{\top},
        \ldots,
        \mathbf{z}_{T}^{\top}
    \right]^{\top}.
\end{equation}
For an observed interior time $t\in\mathcal{T}$ with $1\leq t<T$, the background term is absent because it depends only on the initial latent state $\mathbf{z}_{0}$. The terms involving $\mathbf{z}_{t}$ are therefore the two adjacent latent model errors and the observation error:
\begin{equation}
    \frac{1}{2}
    \left\|
        \mathbf{z}_{t}-\mathbf{K}\mathbf{z}_{t-1}
    \right\|_{\mathbf{Q}_{z}^{-1}}^{2}
    +
    \frac{1}{2}
    \left\|
        \mathbf{z}_{t+1}-\mathbf{K}\mathbf{z}_{t}
    \right\|_{\mathbf{Q}_{z}^{-1}}^{2}
    +
    \frac{1}{2}
    \left\|
        \mathbf{z}_{t}-\widehat{\mathbf{z}}_{t}
    \right\|_{\mathbf{R}_{z}^{-1}}^{2}.
\end{equation}
Differentiating these terms with respect to $\mathbf{z}_{t}$ and setting their sum to zero gives
\begin{equation}
    \mathbf{Q}_{z}^{-1}
    (\mathbf{z}_{t}-\mathbf{K}\mathbf{z}_{t-1})
    -
    \mathbf{K}^{\top}\mathbf{Q}_{z}^{-1}
    (\mathbf{z}_{t+1}-\mathbf{K}\mathbf{z}_{t})
    +
    \mathbf{R}_{z}^{-1}
    (\mathbf{z}_{t}-\widehat{\mathbf{z}}_{t})
    =
    \mathbf{0}.
\end{equation}
The factor $-\mathbf{K}^{\top}$ in the second term appears because $\mathbf{z}_{t}$ enters the next-step residual through $-\mathbf{K}\mathbf{z}_{t}$. After expanding the products and moving the observation target to the right-hand side, this becomes
\begin{equation}
\begin{aligned}
    -\mathbf{Q}_{z}^{-1}\mathbf{K}\mathbf{z}_{t-1}
    &+
    \left(
        \mathbf{Q}_{z}^{-1}
        +
        \mathbf{K}^{\top}\mathbf{Q}_{z}^{-1}\mathbf{K}
        +
        \mathbf{R}_{z}^{-1}
    \right)
    \mathbf{z}_{t}
    -
    \mathbf{K}^{\top}\mathbf{Q}_{z}^{-1}\mathbf{z}_{t+1}
    =
    \mathbf{R}_{z}^{-1}\widehat{\mathbf{z}}_{t}.
\end{aligned}
\end{equation}
Applying this equation at every time step yields a block-tridiagonal linear system for the latent trajectory, with diagonal blocks $\mathbf{A}_{t}$ and right-hand sides $\mathbf{b}_{t}$. At $t=0$, the first latent model error term is replaced by the background term $\frac{1}{2}\|\mathbf{z}_{0}-\mathbf{z}_{0}^{\mathrm{b}}\|_{\mathbf{B}^{-1}}^{2}$; at $t=T$, no further latent transition is considered. The resulting boundary and interior terms are
\begin{equation}
\begin{aligned}
    \mathbf{A}_{0}
    &=
    \mathbf{B}^{-1}
    +
    \mathbf{K}^{\top}\mathbf{Q}_{z}^{-1}\mathbf{K}
    +
    \mathbf{R}_{z}^{-1},\\
    \mathbf{A}_{t}
    &=
    \mathbf{Q}_{z}^{-1}
    +
    \mathbf{K}^{\top}\mathbf{Q}_{z}^{-1}\mathbf{K}
    +
    \mathbf{R}_{z}^{-1},
    \qquad 1\leq t<T,\\
    \mathbf{A}_{T}
    &=
    \mathbf{Q}_{z}^{-1}
    +
    \mathbf{R}_{z}^{-1},\\
    \mathbf{b}_{0}
    &=
    \mathbf{B}^{-1}\mathbf{z}_{0}^{\mathrm{b}}
    +
    \mathbf{R}_{z}^{-1}\widehat{\mathbf{z}}_{0},
    \qquad
    \mathbf{b}_{t}
    =
    \mathbf{R}_{z}^{-1}\widehat{\mathbf{z}}_{t},
    \quad 1\leq t\leq T.
\end{aligned}
\end{equation}
With these definitions, the full system is
\begin{equation}
\begin{bmatrix}
    \mathbf{A}_{0}
    & -\mathbf{K}^{\top}\mathbf{Q}_{z}^{-1}
    & \mathbf{0}
    & \cdots
    & \mathbf{0}\\
    -\mathbf{Q}_{z}^{-1}\mathbf{K}
    & \mathbf{A}_{1}
    & -\mathbf{K}^{\top}\mathbf{Q}_{z}^{-1}
    & \cdots
    & \mathbf{0}\\
    \mathbf{0}
    & -\mathbf{Q}_{z}^{-1}\mathbf{K}
    & \ddots
    & \ddots
    & \vdots\\
    \vdots
    & \ddots
    & \ddots
    & \mathbf{A}_{T-1}
    & -\mathbf{K}^{\top}\mathbf{Q}_{z}^{-1}\\
    \mathbf{0}
    & \cdots
    & \mathbf{0}
    & -\mathbf{Q}_{z}^{-1}\mathbf{K}
    & \mathbf{A}_{T}
\end{bmatrix}
\begin{bmatrix}
    \mathbf{z}_{0}^{\mathrm{a}}\\
    \mathbf{z}_{1}^{\mathrm{a}}\\
    \vdots\\
    \mathbf{z}_{T-1}^{\mathrm{a}}\\
    \mathbf{z}_{T}^{\mathrm{a}}
\end{bmatrix}
=
\begin{bmatrix}
    \mathbf{b}_{0}\\
    \mathbf{b}_{1}\\
    \vdots\\
    \mathbf{b}_{T-1}\\
    \mathbf{b}_{T}
\end{bmatrix}.
\end{equation}
If no observation is available at time $t$ ($t\notin\mathcal{T}$), the observation terms are omitted from $\mathbf{A}_{t}$ and $\mathbf{b}_{t}$. The analysis trajectory is obtained by solving this block-tridiagonal system, with a unique global minimizer guaranteed when the complete Hessian is positive definite.

\subsection{Derivation of the CGKN Conditional Gaussian Filter}
\label{app:cgkn-filter}

Assume that the latent posterior at time $t-1$ is Gaussian,
\begin{equation}
    p(\mathbf{z}_{t-1}\mid\mathbf{y}_{1:t-1})
    =
    \mathcal{N}\!\left(
        \mathbf{z}_{t-1};
        \boldsymbol{\mu}_{t-1},
        \boldsymbol{\Sigma}_{t-1}
    \right).
\end{equation}
For the transition from $t-1$ to $t$, the deterministic coefficients are generated from the previous observation:
\begin{equation}
    \left[
        \mathbf{d}_{t},
        \mathbf{C}_{t},
        \mathbf{f}_{t},
        \mathbf{K}_{t}
    \right]
    =
    \mathcal{G}_{\theta}(\mathbf{y}_{t-1}).
\end{equation}
Conditioned on $\mathbf{y}_{1:t-1}$, these coefficients are fixed. Equation~\eqref{eq:cgkn-discrete} can therefore be re-indexed as
\begin{equation}
\begin{aligned}
    \mathbf{y}_{t}
    &=
    \mathbf{d}_{t}
    +
    \mathbf{C}_{t}\mathbf{z}_{t-1}
    +
    \boldsymbol{\sigma}_{1}\boldsymbol{\epsilon}_{t-1}^{y}, &
    \mathbf{z}_{t}
    &=
    \mathbf{f}_{t}
    +
    \mathbf{K}_{t}\mathbf{z}_{t-1}
    +
    \boldsymbol{\sigma}_{2}\boldsymbol{\epsilon}_{t-1}^{z},
\end{aligned}
\end{equation}
where $\boldsymbol{\epsilon}_{t-1}^{y}$ and $\boldsymbol{\epsilon}_{t-1}^{z}$ are independent standard Gaussian variables. The pair $(\mathbf{z}_{t},\mathbf{y}_{t})$ is consequently jointly Gaussian given $\mathbf{y}_{1:t-1}$, with means
\begin{equation}
    \mathbb{E}[\mathbf{z}_{t}]
    =
    \mathbf{f}_{t}
    +
    \mathbf{K}_{t}\boldsymbol{\mu}_{t-1},
    \qquad
    \mathbb{E}[\mathbf{y}_{t}]
    =
    \mathbf{d}_{t}
    +
    \mathbf{C}_{t}\boldsymbol{\mu}_{t-1},
\end{equation}
and covariance blocks
\begin{equation}
\begin{aligned}
    \operatorname{Cov}(\mathbf{z}_{t})
    &=
    \mathbf{K}_{t}\boldsymbol{\Sigma}_{t-1}\mathbf{K}_{t}^{\top}
    +
    \boldsymbol{\sigma}_{2}\boldsymbol{\sigma}_{2}^{\top},\\
    \operatorname{Cov}(\mathbf{y}_{t})
    &=
    \mathbf{C}_{t}\boldsymbol{\Sigma}_{t-1}\mathbf{C}_{t}^{\top}
    +
    \boldsymbol{\sigma}_{1}\boldsymbol{\sigma}_{1}^{\top},\\
    \operatorname{Cov}(\mathbf{z}_{t},\mathbf{y}_{t})
    &=
    \mathbf{K}_{t}\boldsymbol{\Sigma}_{t-1}\mathbf{C}_{t}^{\top}.
\end{aligned}
\end{equation}
The cross-covariance contains no noise contribution because the two Gaussian noise terms are independent. Define the innovation covariance and Kalman gain as
\begin{equation}
    \mathbf{S}_{t}
    =
    \boldsymbol{\sigma}_{1}\boldsymbol{\sigma}_{1}^{\top}
    +
    \mathbf{C}_{t}\boldsymbol{\Sigma}_{t-1}\mathbf{C}_{t}^{\top},
    \qquad
    \mathbf{G}_{t}
    =
    \mathbf{K}_{t}\boldsymbol{\Sigma}_{t-1}
    \mathbf{C}_{t}^{\top}\mathbf{S}_{t}^{-1}.
\end{equation}
Provided that $\mathbf{S}_{t}$ is positive definite, conditioning the joint Gaussian on the current observation $\mathbf{y}_{t}$ gives
\begin{equation}
\begin{aligned}
    \boldsymbol{\mu}_{t}
    &=
    \mathbf{f}_{t}
    +
    \mathbf{K}_{t}\boldsymbol{\mu}_{t-1}
    +
    \mathbf{G}_{t}
    \left(
        \mathbf{y}_{t}
        -
        \mathbf{d}_{t}
        -
        \mathbf{C}_{t}\boldsymbol{\mu}_{t-1}
    \right),\\
    \boldsymbol{\Sigma}_{t}
    &=
    \mathbf{K}_{t}\boldsymbol{\Sigma}_{t-1}\mathbf{K}_{t}^{\top}
    +
    \boldsymbol{\sigma}_{2}\boldsymbol{\sigma}_{2}^{\top}
    -
    \mathbf{G}_{t}\mathbf{S}_{t}\mathbf{G}_{t}^{\top}.
\end{aligned}
\end{equation}
These are the updates in Eq.~\eqref{eq:cgkn-filter}. Starting from a Gaussian initial distribution $\mathcal{N}(\boldsymbol{\mu}_{0},\boldsymbol{\Sigma}_{0})$, the same argument applies recursively, so the latent state distribution remains Gaussian conditional on the observation history.

\subsection{Derivation of the DBF Posterior Update}
\label{app:dbf-update}

DBF propagates a Gaussian forecast prior in latent space:
\begin{equation}
    p(\mathbf{z}_{t}\mid\mathbf{y}_{1:t-1})
    =
    \mathcal{N}\!\left(
        \mathbf{z}_{t};
        \boldsymbol{\mu}_{t}^{f},
        \boldsymbol{\Sigma}_{t}^{f}
    \right),
    \qquad
    \boldsymbol{\mu}_{t}^{f}
    =
    \mathbf{K}\boldsymbol{\mu}_{t-1}^{\mathrm{a}},
    \qquad
    \boldsymbol{\Sigma}_{t}^{f}
    =
    \mathbf{K}\boldsymbol{\Sigma}_{t-1}^{\mathrm{a}}\mathbf{K}^{\top}
    +\mathbf{Q}_{z}.
\end{equation}
The inverse-observation distribution and virtual prior are defined as
\begin{equation}
    r(\mathbf{z}_{t}\mid\mathbf{y}_{t})
    =
    \mathcal{N}\!\left(
        \mathbf{z}_{t};
        \boldsymbol{\mu}_{t}^{o},
        \boldsymbol{\Sigma}_{t}^{o}
    \right),
    \qquad
    \rho(\mathbf{z}_{t})
    =
    \mathcal{N}\!\left(
        \mathbf{z}_{t};
        \mathbf{m},
        \mathbf{V}
    \right).
\end{equation}
In practice, DBF uses $\mathbf{m}=\mathbf{0}$ and $\mathbf{V}=10^{8}\mathbf{I}$, making the virtual prior broad relative to the inverse-observation distribution.
The DBF update is
\begin{equation}
    q(\mathbf{z}_{t}\mid \mathbf{y}_{1:t})
    \propto
    \frac{
        r(\mathbf{z}_{t}\mid \mathbf{y}_{t})
    }{
        \rho(\mathbf{z}_{t})
    }
    p(\mathbf{z}_{t}\mid \mathbf{y}_{1:t-1}).
\end{equation}
Taking the logarithm and keeping only the terms depending on $\mathbf{z}_{t}$ gives
\begin{equation}
\begin{aligned}
    \log q(\mathbf{z}_{t}\mid\mathbf{y}_{1:t})
    &=
    -\frac{1}{2}
    (\mathbf{z}_{t}-\boldsymbol{\mu}_{t}^{f})^{\top}
    (\boldsymbol{\Sigma}_{t}^{f})^{-1}
    (\mathbf{z}_{t}-\boldsymbol{\mu}_{t}^{f})
    -\frac{1}{2}
    (\mathbf{z}_{t}-\boldsymbol{\mu}_{t}^{o})^{\top}
    (\boldsymbol{\Sigma}_{t}^{o})^{-1}
    (\mathbf{z}_{t}-\boldsymbol{\mu}_{t}^{o})
    \\
    &\quad
    +\frac{1}{2}
    (\mathbf{z}_{t}-\mathbf{m})^{\top}
    \mathbf{V}^{-1}
    (\mathbf{z}_{t}-\mathbf{m})
    +\mathrm{const}.
\end{aligned}
\end{equation}
The positive sign of the third term comes from the virtual prior appearing in the denominator. Define the resulting posterior precision as
\begin{equation}
    \mathbf{P}_{t}
    =
    \left(\boldsymbol{\Sigma}_{t}^{f}\right)^{-1}
    +
    \left(\boldsymbol{\Sigma}_{t}^{o}\right)^{-1}
    -
    \mathbf{V}^{-1}.
\end{equation}
Provided that $\mathbf{P}_{t}$ is positive definite, the Gaussian quotient is normalizable and the analysis posterior is
\begin{equation}
    q(\mathbf{z}_{t}\mid \mathbf{y}_{1:t})
    =
    \mathcal{N}
    \left(
        \mathbf{z}_{t};
        \boldsymbol{\mu}_{t}^{\mathrm{a}},
        \boldsymbol{\Sigma}_{t}^{\mathrm{a}}
    \right),
\end{equation}
with
\begin{equation}
    \boldsymbol{\mu}_{t}^{\mathrm{a}}
    =
    \boldsymbol{\Sigma}_{t}^{\mathrm{a}}
    \left[
        \left(
            \boldsymbol{\Sigma}_{t}^{f}
        \right)^{-1}
        \boldsymbol{\mu}_{t}^{f}
        +
        \left(
            \boldsymbol{\Sigma}_{t}^{o}
        \right)^{-1}
        \boldsymbol{\mu}_{t}^{o}
        -
        \mathbf{V}^{-1}\mathbf{m}
    \right],
    \qquad
    \boldsymbol{\Sigma}_{t}^{\mathrm{a}}
    =
    \mathbf{P}_{t}^{-1}.
\end{equation}
This is the closed-form DBF update used in the main text.

For a Gaussian reconstruction likelihood with mean $\mathcal{D}_{x}(\mathbf{z}_{t})$ and covariance $\mathbf{R}_{x}$, the training objective in Eq.~\eqref{eq:dbf-training-loss} expands as follows, up to an additive constant independent of the learned parameters:
\begin{equation}
\begin{aligned}
\mathcal{L}_{\mathrm{DBF}}
(\mathbf{K},\mathcal{D}_{x},\mathcal{G}_{\theta})
&=\frac{1}{2}\sum_t \Biggl[
\mathbb{E}_{q(\mathbf{z}_{t}\mid\mathbf{y}_{1:t})}
\left\|\mathbf{x}_{t}-\mathcal{D}_{x}(\mathbf{z}_{t})\right\|_{\mathbf{R}_{x}^{-1}}^{2}
+\log\det\mathbf{R}_{x}
\\
&\quad+
\operatorname{tr}\!\left(
(\boldsymbol{\Sigma}_{t}^{f})^{-1}
\boldsymbol{\Sigma}_{t}^{\mathrm{a}}
\right)
+\left\|
\boldsymbol{\mu}_{t}^{\mathrm{a}}
-\mathbf{K}\boldsymbol{\mu}_{t-1}^{\mathrm{a}}
\right\|_{(\boldsymbol{\Sigma}_{t}^{f})^{-1}}^{2}
-d+
\log\frac{\det\boldsymbol{\Sigma}_{t}^{f}}
{\det\boldsymbol{\Sigma}_{t}^{\mathrm{a}}}
\Biggr].
\end{aligned}
\label{eq:dbf-training-loss-operators}
\end{equation}
Here, $d$ is the latent dimension and $\mathbf{R}_{x}$ is the covariance of the physical state reconstruction likelihood. The forecast prior and inverse-observation operator supply
\begin{equation}
\begin{aligned}
\boldsymbol{\Sigma}_{t}^{f}
&=\mathbf{K}\boldsymbol{\Sigma}_{t-1}^{\mathrm{a}}\mathbf{K}^{\top}
+\mathbf{Q}_{z}, &
(\boldsymbol{\mu}_{t}^{o},\boldsymbol{\Sigma}_{t}^{o})
&=\mathcal{G}_{\theta}(\mathbf{y}_{t}).
\end{aligned}
\end{equation}
The analysis mean $\boldsymbol{\mu}_{t}^{\mathrm{a}}$ and covariance $\boldsymbol{\Sigma}_{t}^{\mathrm{a}}$ are computed from these quantities through the posterior update derived above. Thus, $\mathcal{D}_{x}$ enters the reconstruction term directly, $\mathbf{K}$ determines the forecast prior, and $\mathcal{G}_{\theta}$ affects both loss terms through the analysis posterior. Gradients are propagated through these recursive updates to jointly train the three operators.

\section{Experimental Configuration}
\label{app:experimental-details}

This appendix provides details of the data preparation and model training, including the estimation and calibration of error covariances.

\subsection{Data Set Preparation and Data Splits}
\label{app:dataset-preparation}

The benchmark data sets are partitioned according to their trajectory structure, physical parameters, or temporal coverage, as described below.

\textit{Lorenz--96 and shallow water.}
The training, validation, and test sets contain separate trajectories. Each trajectory is assigned to a single set, so temporal segments from the same trajectory are not shared across the three sets.

\textit{Cylinder.}
We use 49 trajectories from the cylinder-flow BC subset of CFDBench. The trajectories are partitioned at the case level, with training, validation, and test cases distributed across the Reynolds-number range.

\textit{ERA5--Coarse and ERA5--High.}
The ERA5 data sets are partitioned chronologically: records from 1979--2015 are used for training, 2016 for validation, and 2018 for testing. The same temporal partition is used for both spatial resolutions. Training sequences are constructed using sliding windows, whereas validation and test sequences are extracted using fixed windows.

\subsection{Training Hyperparameters}
\label{app:hyperparameters}

For Tensor-Var, we set $\lambda_K=0.01\,\operatorname{tr}(\mathbf{Z}\mathbf{Z}^{\top})/d$, where $d$ is the latent dimension. The Koopman operator is fitted using at most $2\times10^4$ consecutive latent state pairs from the training split.

\begin{table}[htbp]
\centering
\setlength{\tabcolsep}{4pt}
\renewcommand{\arraystretch}{0.95}
\resizebox{\textwidth}{!}{%
\begin{tabular}{@{}llrllrrr@{}}
\toprule
Data set & Model & Parameters & Optimizer & Learning rate(s) & Batch & Epochs & Seeds \\
\midrule
Lorenz--96 & DMD & -- & -- & -- & -- & -- & -- \\
Lorenz--96 & NL-AE & 380,440 & AdamW & $3\times10^{-5}/10^{-3}$ & 16 & 100 & 1, 2, 3 \\
Lorenz--96 & WL-AE & 380,440 & AdamW & $3\times10^{-5}/10^{-3}$ & 16 & 100 & 1, 2, 3 \\
Lorenz--96 & KAE & 380,640 & AdamW & $3\times10^{-5}/10^{-3}$ & 16 & 100 & 1, 2, 3 \\
Lorenz--96 & Tensor-Var & 16,401,622 & AdamW & $3\times10^{-4}$ & 64 & 60 & 1, 2, 3 \\
Lorenz--96 & DBF & 549,520 & Adam & $3\times10^{-4}$ & 16 & 100 & 1, 2, 3 \\
Lorenz--96 & CGKN$^{*}$ & 26,098 & Adam & $10^{-3}/10^{-3}$ & 200 & 500/2500 & 1, 2, 3 \\
\midrule
Cylinder & DMD & -- & -- & -- & -- & -- & -- \\
Cylinder & NL-AE & 16,245,022 & AdamW & $3\times10^{-5}/10^{-3}$ & 16 & 100 & 1, 2, 3 \\
Cylinder & WL-AE & 16,245,022 & AdamW & $3\times10^{-5}/10^{-3}$ & 16 & 100 & 1, 2, 3 \\
Cylinder & KAE & 11,252,510 & AdamW & $3\times10^{-5}/10^{-3}$ & 16 & 100 & 1, 2, 3 \\
Cylinder & Tensor-Var & 25,601,440 & AdamW & $3\times10^{-4}$ & 16 & 60 & 1, 2, 3 \\
Cylinder & DBF & 16,473,952 & Adam & $3\times10^{-4}$ & 16 & 100 & 1, 2, 3 \\
Cylinder & CGKN$^{*}$ & 11,224,192 & Adam & $10^{-3}/3\times10^{-4}$ & 64 & 10/400 & 1, 2, 3 \\
\midrule
Shallow water & DMD & -- & -- & -- & -- & -- & -- \\
Shallow water & NL-AE & 17,193,607 & AdamW & $3\times10^{-5}/10^{-3}$ & 16 & 100 & 1, 2, 3 \\
Shallow water & WL-AE & 17,193,607 & AdamW & $3\times10^{-5}/10^{-3}$ & 16 & 100 & 1, 2, 3 \\
Shallow water & KAE & 12,610,055 & AdamW & $3\times10^{-5}/10^{-3}$ & 16 & 100 & 1, 2, 3 \\
Shallow water & Tensor-Var & 27,312,746 & AdamW & $3\times10^{-4}$ & 16 & 60 & 1, 2, 3 \\
Shallow water & DBF & 16,912,337 & Adam & $3\times10^{-4}$ & 16 & 100 & 1, 2, 3 \\
Shallow water & CGKN$^{*}$ & 3,467,224 & Adam & $10^{-3}/3\times10^{-4}$ & 64 & 10/400 & 1, 2, 3 \\
\midrule
ERA5--Coarse & DMD & -- & -- & -- & -- & -- & -- \\
ERA5--Coarse & NL-AE & 16,871,705 & AdamW & $3\times10^{-5}/10^{-3}$ & 64 & 100 & 1, 2, 3 \\
ERA5--Coarse & WL-AE & 16,871,705 & AdamW & $3\times10^{-5}/10^{-3}$ & 64 & 100 & 1, 2, 3 \\
ERA5--Coarse & KAE & 12,140,825 & AdamW & $3\times10^{-5}/10^{-3}$ & 64 & 100 & 1, 2, 3 \\
ERA5--Coarse & Tensor-Var & 26,469,694 & AdamW & $3\times10^{-4}$ & 64 & 60 & 1, 2, 3 \\
ERA5--Coarse & DBF & 16,624,755 & Adam & $3\times10^{-4}$ & 64 & 100 & 1, 2, 3 \\
ERA5--Coarse & CGKN$^{*}$ & 20,268,832 & Adam & $10^{-3}/3\times10^{-4}$ & 64 & 10/400 & 1, 2, 3 \\
\midrule
ERA5--High & DMD & -- & -- & -- & -- & -- & -- \\
ERA5--High & NL-AE & 21,348,125 & AdamW & $3\times10^{-5}/10^{-3}$ & 8 & 100 & 1, 2, 3 \\
ERA5--High & WL-AE & 21,348,125 & AdamW & $3\times10^{-5}/10^{-3}$ & 8 & 100 & 1, 2, 3 \\
ERA5--High & KAE & 40,243,269 & AdamW & $3\times10^{-5}/10^{-3}$ & 8 & 100 & 1, 2, 3 \\
ERA5--High & Tensor-Var & 55,214,186 & AdamW & $3\times10^{-4}$ & 4 & 30 & 1, 2, 3 \\
ERA5--High & DBF & 16,615,651 & Adam & $3\times10^{-4}$ & 8 & 100 & 1, 2, 3 \\
ERA5--High & CGKN$^{*}$ & 7,418,016 & Adam & $10^{-3}/3\times10^{-4}$ & 64 & 10/400 & 1, 2, 3 \\
\bottomrule
\end{tabular}%
}
\caption{Training configurations of the benchmarked methods. Parameter counts are computed from the released checkpoint tensors. For NL-AE, WL-AE, and KAE, learning rates are listed in encoder--decoder and latent transition order. Tensor-Var settings refer to inverse-observation encoder training. DBF settings follow the implementation defaults.}
\label{tab:training-hyperparameters}
\par\smallskip
\begin{minipage}{\textwidth}
\footnotesize
$^{*}$CGKN uses two-stage training; slash-separated values indicate the settings for Stage 1 and Stage 2, respectively. Stage 2 batch sizes are not specified here.
\end{minipage}
\end{table}

Tensor-Var uses a weight decay of $10^{-4}$ for inverse-observation encoder training. Its pretrained state encoder--decoder is frozen, and its Koopman operator is fitted by ridge regression. DBF uses cosine learning rate decay.

\subsection{Training Loss Weights}
\label{app:training-loss-weights}

Table~\ref{tab:training-loss-weights} summarizes the weights used in the training objectives described in Section~\ref{sec:koopman-type-latent-da}. Tensor-Var trains the state autoencoder and inverse-observation encoder separately. In our experiments, the autoencoder is trained using only the reconstruction loss ($\lambda_{\mathrm{pred}}=0$), then frozen before fitting $\mathbf{K}$ to encoded training state pairs. For the field data sets, the inverse-observation training additionally includes an MSE loss with unit weight between the reconstructed physical field before state encoding and the ground-truth state. CGKN uses stage-dependent weights, with zero indicating that a loss term is excluded from the corresponding training stage. DMD is fitted analytically and therefore has no neural network training loss weights.

\begin{table}[htbp]
    \centering
    \small
    \setlength{\tabcolsep}{4pt}
    \renewcommand{\arraystretch}{1.2}
    \begin{tabularx}{\textwidth}{
        @{}l
        >{\raggedright\arraybackslash}p{0.27\textwidth}
        >{\raggedright\arraybackslash}X@{}
    }
        \toprule
        Method & Data set or training stage & Loss weights \\
        \midrule

        NL-AE
        & All systems
        & $\lambda_{\mathrm{rec}}=\lambda_{\mathrm{pred}}=1$;
          $\lambda_{\mathrm{latent}}=0$ \\
        \addlinespace

        WL-AE
        & All systems
        & $\lambda_{\mathrm{rec}}=\lambda_{\mathrm{pred}}=1$;
          $\lambda_{\mathrm{latent}}
          =\lambda_{\mathrm{straight}}=0.1$ \\
        \addlinespace

        KAE
        & All systems
        & $\lambda_{\mathrm{rec}}=\lambda_{\mathrm{pred}}=1$;
          $\lambda_{\mathrm{latent}}=0.1$;
          $\lambda_{\mathrm{spec}}=1$ for shallow water
          and $10$ for the other systems \\
        \addlinespace

        Tensor-Var
        & State autoencoder
        & $\lambda_{\mathrm{rec}}=1$;
          $\lambda_{\mathrm{pred}}=0$ \\

        Tensor-Var
        & Inverse-observation encoder
        & Latent state alignment weight: $1$;
          physical state reconstruction weight:
          $\lambda_{\mathrm{rec}}=1$ \\
        \addlinespace

        DMD
        & All systems
        & Not applicable \\
        \addlinespace

        DBF
        & All systems
        & Reconstruction and KL weights: $1$ each \\
        \addlinespace

        CGKN
        & Lorenz--96
        & Each active loss term has weight $1$ \\

        CGKN
        & Other systems: Stage 1
        & $\lambda_{\mathrm{rec}}=\lambda_x=\lambda_z=1$;
          $\lambda_{\mathrm{DA}}=0$ \\

        CGKN
        & Other systems: Stage 2
        & $\lambda_{\mathrm{rec}}=0.2$;
          $\lambda_x=\lambda_z=0$;
          $\lambda_{\mathrm{DA}}=1$ \\

        \bottomrule
    \end{tabularx}
    \caption{Training loss weights. Unless otherwise specified, the settings apply to all evaluated systems.}
    \label{tab:training-loss-weights}
\end{table}

\subsection{Hardware Setup}
\label{app:hardware-setup}

Most experiments were conducted using a single NVIDIA GeForce RTX~5090 GPU with 32\,GB of video memory. The local system was equipped with an AMD Ryzen AI~9 HX~370 processor with 12 cores and 24 threads, and 64\,GB of system memory. The ERA5 experiments were conducted on a cloud-computing instance equipped with an NVIDIA RTX PRO~6000 GPU and a 25-core CPU. All models were trained and evaluated under Ubuntu~24.04 LTS.

\subsection{Covariance Construction and Calibration}
\label{app:covariance-calibration}

For the variational methods, a proxy for the latent background error covariance $\mathbf{B}$ is constructed using the NMC method~\citep{NMC}. Specifically, latent forecasts with different lead times are compared at the same physical time. Across all five data sets, we randomly sample 300 forecast pairs with replacement using a fixed seed. The short- and long-forecast lead times are summarized in Table~\ref{tab:nmc-lead-times}.

\begin{table}[htbp]
    \centering
    \begin{tabular}{lcc}
        \toprule
        Data set & Short forecast lead & Long forecast lead \\
        \midrule
        Lorenz--96    & 0.20 MTU  & 0.40 MTU \\
        Cylinder      & 20 frames & 40 frames \\
        Shallow water & 8 frames  & 16 frames \\
        ERA5--Coarse          & 4 days    & 8 days \\
        ERA5--High    & 4 days    & 8 days \\
        \bottomrule
    \end{tabular}
    \caption{Short- and long-forecast leads used for NMC covariance estimation. Frames refer to the stored data before subsampling.}
    \label{tab:nmc-lead-times}
\end{table}

Following the NMC method~\citep{NMC}, the covariance proxy is constructed as
\begin{equation}
    \widehat{\mathbf{B}}
    =
    \operatorname{Cov}\!\left(
        \mathbf{z}_{t}^{f,\mathrm{long}}
        -
        \mathbf{z}_{t}^{f,\mathrm{short}}
    \right),
\end{equation}
where the two forecasts have different lead times but share the same valid time $t$. The covariance is estimated using the empirical second moment under the assumed zero-mean background error model.

For the autoencoder variants, the observation error and latent model error covariances are parameterized as
\begin{equation}
    \mathbf{R}=\sigma_r^2\mathbf{I},
    \qquad
    \mathbf{Q}_{z}=\sigma_q^2\mathbf{I},
\end{equation}
where $\sigma_r$ and $\sigma_q$ control the observation error and latent model error scales, respectively. We first select $\sigma_r$ by grid search on the validation set under the strong-constraint formulation, then hold it fixed while selecting $\sigma_q$ under the weak-constraint formulation. In each stage, we select the value that yields the lowest mean assimilation RRMSE.

For Tensor-Var and DMD, the latent model error and observation error covariances are approximated using empirical second moments of residuals computed on the training set:
\begin{equation}
    \widehat{\mathbf{Q}}_{z}
    =
    \frac{1}{N_q}\mathbf{E}_{q}^{\top}\mathbf{E}_{q},
    \qquad
    \widehat{\mathbf{R}}_{z}
    =
    \frac{1}{N_r}\mathbf{E}_{r}^{\top}\mathbf{E}_{r},
\end{equation}
where the rows of $\mathbf{E}_{q}$ and $\mathbf{E}_{r}$ contain the residuals $\mathbf{z}_{t}-\mathbf{K}\mathbf{z}_{t-1}$ and $\mathbf{z}_{t}-\widehat{\mathbf{z}}_{t}$, respectively, and $N_q$ and $N_r$ denote their sample counts. Here, $\mathbf{z}_{t}$ is the encoded or projected physical state and $\widehat{\mathbf{z}}_{t}$ is the observation-derived latent state. These empirical second moments approximate the covariances under the assumed zero-mean error model.

To stabilize covariance inversion when the estimates are rank deficient or poorly conditioned, we regularize the covariance matrices as
\begin{equation}
    \mathbf{B}
    =
    \widehat{\mathbf{B}}
    + 0.05\,\frac{\operatorname{tr}(\widehat{\mathbf{B}})}{d}\mathbf{I},
    \qquad
    \mathbf{Q}_{z}
    =
    \widehat{\mathbf{Q}}_{z}
    + \epsilon\,\frac{\operatorname{tr}(\widehat{\mathbf{Q}}_{z})}{d}\mathbf{I},
    \qquad
    \mathbf{R}_{z}
    =
    \widehat{\mathbf{R}}_{z}
    + \epsilon\,\frac{\operatorname{tr}(\widehat{\mathbf{R}}_{z})}{d}\mathbf{I}.
\end{equation}
Here, $d$ is the latent dimension. The coefficient for $\mathbf{B}$ is $0.05$ across all data sets and variational methods. For Tensor-Var, we use $\epsilon=0.01$ on Lorenz--96 and $\epsilon=0.1$ on the remaining data sets. For DMD, we use $\epsilon=0.1$ across all data sets. In each setting, the same $\epsilon$ is applied to both $\widehat{\mathbf{Q}}_{z}$ and $\widehat{\mathbf{R}}_{z}$. For Tensor-Var and DMD, the assimilation objective uses the scaled covariances $\sigma_q^2\mathbf{Q}_{z}$ and $\sigma_r^2\mathbf{R}_{z}$, with the scaling factors selected by grid search to minimize mean assimilation RRMSE on the validation set and held fixed during testing.

Table~\ref{tab:nmc-conditioning} reports the condition numbers of the regularized background error covariance matrices for the five variational methods. ERA5--High exhibits the largest condition number for each method, indicating greater sensitivity to numerical errors in covariance inversion, which can affect the weighting of the background error term in the assimilation objective.

\begin{table}[htbp]
    \centering
    \begin{tabular}{lccccc}
        \toprule
        Data set & NL-AE & WL-AE & KAE & Tensor-Var & DMD \\
        \midrule
        Lorenz--96
        & $2.03\times10^{2}$
        & $2.41\times10^{2}$
        & $2.35\times10^{2}$
        & $1.04\times10^{2}$
        & $1.24\times10^{2}$ \\
        Cylinder
        & $9.68\times10^{3}$
        & $6.31\times10^{3}$
        & $4.93\times10^{3}$
        & $4.99\times10^{3}$
        & $3.74\times10^{3}$ \\
        Shallow water
        & $1.69\times10^{3}$
        & $1.83\times10^{3}$
        & $6.42\times10^{3}$
        & $7.10\times10^{2}$
        & $2.84\times10^{3}$ \\
        ERA5--Coarse
        & $3.52\times10^{3}$
        & $1.32\times10^{4}$
        & $2.69\times10^{3}$
        & $1.48\times10^{3}$
        & $1.56\times10^{3}$ \\
        ERA5--High
        & $6.88\times10^{4}$
        & $5.08\times10^{4}$
        & $8.94\times10^{4}$
        & $6.83\times10^{3}$
        & $4.96\times10^{3}$ \\
        \bottomrule
    \end{tabular}
    \caption{Condition numbers of the regularized NMC background error covariance matrices for the five variational methods.}
    \label{tab:nmc-conditioning}
\end{table}
 
For the filtering-based methods, we follow the covariance parameterizations and training procedures of the original implementations. DBF combines a predefined latent model error covariance $\mathbf{Q}_{z}$ with an observation-derived covariance $\boldsymbol{\Sigma}_{t}^{o}$ learned by the inverse-observation network from the current observation~\citep{Tarumi2025}. We set $\mathbf{Q}_{z}=e^{-4}\mathbf{I}$ across all data sets and hold it fixed during training and assimilation. For CGKN~\citep{Chen_2025}, the observed state noise coefficient on Lorenz--96 is estimated from the one-step prediction residuals of a pretrained model, while the latent state noise coefficient is fixed at $\boldsymbol{\sigma}_{2}=\mathbf{I}$. On the remaining data sets, we set $\boldsymbol{\sigma}_{1}=\sqrt{0.05}\mathbf{I}$ and $\boldsymbol{\sigma}_{2}=\mathbf{I}$. These noise settings remain fixed during assimilation-stage training and evaluation. For DBF, the initial latent state covariance is obtained through a Gaussian update combining the prescribed prior with the inverse-observation output for the initial physical state. For CGKN, the initial latent state covariance is set to $0.01\mathbf{I}$.

For the variational methods, the error covariance matrices for the background state, latent dynamics, and observations are estimated or calibrated offline and held fixed during assimilation. These flow-independent covariances do not adapt to changes in error statistics during assimilation. Further improvements in covariance modeling for individual variational methods may therefore improve assimilation performance. In contrast, DBF and CGKN recursively propagate and update the latent state forecast and analysis covariances, allowing the latent state uncertainty to evolve during assimilation even when the model error covariances or noise coefficients remain fixed.

\section{Evaluation Methods}
\label{app:evaluation-methods}

This appendix details the evaluation metrics and diagnostic procedures used in the experiments.

\subsection{Evaluation Metrics}
\label{app:evaluation-metrics}

We report forecasting and data assimilation performance using the relative root-mean-square error (RRMSE) and the structural similarity index (SSIM). The RRMSE measures the magnitude of the state error relative to the ground-truth state and therefore quantifies pointwise numerical accuracy. At prediction step $t$, it is defined as
\begin{equation}
    \mathrm{RRMSE}_{t}
    =
    \frac{1}{N}
    \sum_{n=1}^{N}
    \frac{
        \left\|
            \widehat{\mathbf{x}}_{n,t}
            -
            \mathbf{x}_{n,t}
        \right\|_{2}
    }{
        \left\|
            \mathbf{x}_{n,t}
        \right\|_{2}
    },
\end{equation}
where $\mathbf{x}_{n,t}$ and $\widehat{\mathbf{x}}_{n,t}$ denote the ground-truth and predicted states for trajectory $n$, respectively, and $N$ is the number of evaluation trajectories.

RRMSE measures the relative magnitude of the state error. For spatial states, we additionally report the structural similarity index (SSIM) to assess whether the predicted field preserves the spatial structure of the ground-truth field~\citep{Wang2004}. SSIM is computed locally and then averaged over the domain:
\begin{equation}
    \operatorname{SSIM}(\mathbf{x},\widehat{\mathbf{x}})
    =
    \frac{1}{|\Omega|}
    \sum_{p\in\Omega}
    \frac{
        (2\mu_{\mathbf{x},p}\mu_{\widehat{\mathbf{x}},p}+C_1)
        (2\sigma_{\mathbf{x}\widehat{\mathbf{x}},p}+C_2)
    }{
        (\mu_{\mathbf{x},p}^{2}+\mu_{\widehat{\mathbf{x}},p}^{2}+C_1)
        (\sigma_{\mathbf{x},p}^{2}+\sigma_{\widehat{\mathbf{x}},p}^{2}+C_2)
    },
\end{equation}
where $\mu$, $\sigma^2$, and $\sigma$ denote local means, variances, and covariance, respectively, and $\Omega$ contains the window centers used for averaging. We use $7\times7$ windows, or 7-point periodic windows for Lorenz--96, and average SSIM across channels. The constants $C_1=(0.01L)^2$ and $C_2=(0.03L)^2$ ensure numerical stability, where $L$ is the ground-truth range.

To distinguish systematic bias from errors in state variability, we additionally report the normalized mean bias (NMB) and variance ratio (VR):
\begin{equation}
\begin{aligned}
    \mathrm{NMB}
    &=
    \frac{
        \operatorname{mean}\!\left(\widehat{\mathbf{x}}-\mathbf{x}\right)
    }{
        \sqrt{\operatorname{mean}\!\left(\mathbf{x}^{2}\right)}
    },
    \qquad
    \mathrm{VR}
    &=
    \frac{
        \operatorname{Var}\!\left(\widehat{\mathbf{x}}\right)
    }{
        \operatorname{Var}\!\left(\mathbf{x}\right)
    }.
\end{aligned}
\label{eq:nmb-vr}
\end{equation}
For each evaluation trajectory, the means and variances are pooled over the evaluated time steps, channels, and spatial locations. NMB is normalized by the root mean square of the true physical state because the standardized anomaly fields can have means close to zero. Positive and negative NMB values indicate positive and negative systematic offsets, respectively. VR values below one indicate insufficient state variability or smoothing, whereas values above one indicate excessive variability. A VR value near one does not imply an accurate spatial reconstruction and is therefore interpreted together with RRMSE and SSIM.

\subsection{Latent-Trajectory Visualization and Distance Evaluation}
\label{app:latent-trajectory-alignment}

\textit{t-SNE visualization.}
For each panel in Figure~\ref{fig:latent-trajectories}, the reference, forecast, and analysis latent states are concatenated and embedded jointly using the scikit-learn implementation of t-SNE. The parameter settings are summarized in Table~\ref{tab:tsne-parameters}. Each panel is fitted independently, so the embedding coordinates are not directly comparable across panels.

\begin{table}[htbp]
    \centering
    \small
    \begin{tabular}{@{}ll@{}}
        \toprule
        Parameter & Setting \\
        \midrule
        Embedding dimension & 2 \\
        Perplexity & $\min(30,\max(5,\lfloor n/4\rfloor))$ \\
        Initialization & PCA \\
        Learning rate & \texttt{auto} \\
        Early exaggeration & 12 \\
        Maximum optimization iterations & 1,000 \\
        Distance metric & Euclidean \\
        Method & Barnes--Hut \\
        Barnes--Hut angle & 0.5 \\
        Random seed & 0 \\
        \bottomrule
    \end{tabular}
    \caption{Parameter settings for the t-SNE visualizations. Here, $n$ denotes the total number of latent states jointly embedded in each panel.}
    \label{tab:tsne-parameters}
\end{table}

Before embedding, all three trajectories are centered using the reference-trajectory mean and clipped at ten times the maximum reference-state distance from this mean. Any trajectory containing nonfinite values is truncated at its first nonfinite state. These preprocessing steps apply only to the visualization; the analysis-to-reference distances below are computed from the original latent states without this clipping.

\textit{Analysis-to-reference distance.}
The normalized analysis-to-reference distances reported in Table~\ref{tab:latent-reference-distance} complement the qualitative visualizations in Section~\ref{sec:latent-trajectory-visualization}. Distances are computed between the analysis and reference states at corresponding times in each model's original latent space, before dimensionality reduction.

For each model and data set, we compute the distances between analysis and reference states at the same observation times after spinup, using the same test trajectories across models. Let $N$ denote the total number of evaluated analysis--reference pairs, and let $\mathbf{z}_{i}^{\mathrm{a}}$ and $\mathbf{z}_{i}^{\mathrm{ref}}$ denote the corresponding analysis and reference latent states. The reference mean, normalization scale, and normalized distance for each pair are defined as
\begin{equation}
\bar{\mathbf{z}}^{\mathrm{ref}}
=
\frac{1}{N}\sum_{i=1}^{N}\mathbf{z}_{i}^{\mathrm{ref}},
\qquad
s_{\mathrm{ref}}
=
\left(
\frac{1}{N}\sum_{i=1}^{N}
\left\|
\mathbf{z}_{i}^{\mathrm{ref}}
-
\bar{\mathbf{z}}^{\mathrm{ref}}
\right\|_{2}^{2}
\right)^{1/2},
\qquad
d_{\mathrm{ref},i}
=
\frac{
\left\|
\mathbf{z}_{i}^{\mathrm{a}}
-
\mathbf{z}_{i}^{\mathrm{ref}}
\right\|_{2}
}{
s_{\mathrm{ref}}
}.
\label{eq:latent-reference-distance}
\end{equation}

A single normalization scale is shared across all evaluation runs within each model and data set. This definition is invariant to a common translation, uniform rescaling, or orthogonal transformation of the latent coordinates. The analysis and reference states are not centered separately, so systematic offsets between them remain part of the measured distance.

Table~\ref{tab:latent-reference-distance} reports the median and the 25th and 75th percentiles of $d_{\mathrm{ref},i}$ over the pooled trajectory--time pairs. We use the median because the distance distributions can be strongly skewed, with a small number of large distances dominating the mean. The median therefore summarizes typical analysis-to-reference alignment, while the quartiles describe its variability across trajectories and time steps. Smaller values indicate closer alignment within each model's latent space.

\subsection{ResDMD Residual Evaluation}
\label{app:koopman-invariance-analysis}

As defined in Section~\ref{sec:deep-koopman}, a finite-dimensional subspace of observable functions $\mathcal{F}_{d}$ is Koopman invariant when $\mathcal{K}\mathcal{F}_{d}\subseteq\mathcal{F}_{d}$~\citep{Brunton_2016}. Under this condition, the evolution of these observables admits an exact finite-dimensional linear representation. In practice, the learned linear operator generally only approximates the evolution of the learned observables, and analysis corrections may further change the latent trajectory's consistency with the corresponding Koopman evolution. We use the residual from Residual Dynamic Mode Decomposition (ResDMD)~\citep{colbrook2024rigorous,colbrook2023residual} to quantify this change.

Computing the reference and analysis residuals requires a common set of eigenpairs associated with a linear operator. For the data-driven comparison, a Koopman approximation $\widehat{\mathbf{K}}$ is fitted separately for each method by least squares using reference snapshot pairs constructed from the training data. Each method uses its corresponding latent representation, and the fitted operator is held fixed when evaluating the reference and analysis trajectories.

For the model-specific comparison, methods with fixed linear latent transitions use their trained Koopman operators, except for DMD, whose operator is reconstructed through its fitting procedure. The residuals are evaluated on the full available trajectory set for each method. For simplicity, we use $\mathbf{K}$ to denote either the fitted or model-specific operator throughout the following derivation.

Let $\lambda_{i}$ and $\mathbf{v}_{i}$ be an eigenvalue and corresponding eigenvector of $\mathbf{K}^{\top}$:
\begin{equation}
    \mathbf{K}^{\top}\mathbf{v}_{i}
    =
    \lambda_{i}\mathbf{v}_{i}.
\end{equation}
If a latent trajectory evolves exactly under $\mathbf{K}$, such that $\mathbf{z}_{t+1}=\mathbf{K}\mathbf{z}_{t}$, then
\begin{equation}
    \mathbf{z}_{t+1}^{\top}\mathbf{v}_{i}
    =
    \left(\mathbf{K}\mathbf{z}_{t}\right)^{\top}\mathbf{v}_{i}
    =
    \mathbf{z}_{t}^{\top}\mathbf{K}^{\top}\mathbf{v}_{i}
    =
    \lambda_{i}\mathbf{z}_{t}^{\top}\mathbf{v}_{i}.
\end{equation}
For each evaluated trajectory, $\mathbf{Z}$ and $\mathbf{Z}^{+}$ are constructed from consecutive latent states as defined in Eq.~\eqref{eq:latent-snapshot-matrices}. Under this exact linear evolution, the latent snapshots satisfy
$(\mathbf{Z}^{+})^{\top}\mathbf{v}_{i}
=\lambda_{i}\mathbf{Z}^{\top}\mathbf{v}_{i}$.
ResDMD quantifies the normalized mismatch from this relation as
\begin{equation}
\begin{aligned}
    \operatorname{res}(\lambda_{i},\mathbf{v}_{i})^{2}
    &=
    \frac{
        \left\|
            \left(\mathbf{Z}^{+}\right)^{\top}\mathbf{v}_{i}
            -
            \lambda_{i}\mathbf{Z}^{\top}\mathbf{v}_{i}
        \right\|^{2}
    }{
        \left\|\mathbf{Z}^{\top}\mathbf{v}_{i}\right\|^{2}
    } \\
    &=
    \frac{
        \mathbf{v}_{i}^{*}
        \left[
            \mathbf{Z}^{+}\left(\mathbf{Z}^{+}\right)^{\top}
            -\lambda_{i}\mathbf{Z}^{+}\mathbf{Z}^{\top}
            -\overline{\lambda}_{i}\mathbf{Z}\left(\mathbf{Z}^{+}\right)^{\top}
            +|\lambda_{i}|^{2}\mathbf{Z}\mathbf{Z}^{\top}
        \right]
        \mathbf{v}_{i}
    }{
        \mathbf{v}_{i}^{*}\mathbf{Z}\mathbf{Z}^{\top}\mathbf{v}_{i}
    },
\end{aligned}
\end{equation}
where $(\cdot)^{*}$ denotes the conjugate transpose and $\overline{\lambda}_{i}$ is the complex conjugate of $\lambda_{i}$. A smaller residual indicates that the trajectory more closely follows the corresponding Koopman evolution. An eigenpair with negligible snapshot energy $\|\mathbf{Z}^{\top}\mathbf{v}_{i}\|^{2}$ makes the denominator of the normalized residual close to zero and therefore leads to numerical instability. To avoid near-zero denominators, we retain eigenpairs with snapshot energy above $10^{-8}$ of the maximum in both the reference and analysis data.

To evaluate the effect of assimilation, we compare the analysis trajectory with a reference trajectory derived from the ground-truth physical states, following the construction described in Section~\ref{sec:latent-trajectory-visualization}. For $\operatorname{res}_{i}^{\mathrm{ref}}$, $\mathbf{Z}$ and $\mathbf{Z}^{+}$ are constructed from consecutive states of the reference trajectory, whereas for $\operatorname{res}_{i}^{\mathrm{a}}$ they are constructed from consecutive states of the analysis trajectory. Both residuals use the same eigenpair $(\lambda_i,\mathbf{v}_i)$. We summarize the evaluated spectral residuals using the median, which is less sensitive to a small number of large residuals than the mean. The reference and analysis medians are computed separately, and their ratio defines $R_{\mathrm{res}}$ in Eq.~\eqref{eq:resdmd-ratio-new}. Values below and above one indicate improved and reduced dynamical consistency after assimilation, respectively.

\subsection{Gaussianity Evaluation Procedure}
\label{app:gaussianity-analysis}

Gaussianity is evaluated in latent space by repeating assimilation from multiple perturbed background states and examining the resulting error distribution. Because the background error distribution varies across evaluation trajectories, we control the initial error distribution to isolate the effect of the assimilation procedure. All methods therefore use the same Gaussian perturbation scheme. For each evaluation start, let $\mathbf{z}_{0}^{\mathrm{true}}$ denote the latent representation of the ground-truth initial state. We generate $N_{\mathrm{ens}}=512$ background latent states by adding independent perturbations sampled from $\mathcal{N}(\mathbf{0},\mathbf{B})$, where $\mathbf{B}$ is the same for all evaluation starts. For this Gaussianity experiment, $\mathbf{B}$ is the regularized NMC estimate for the five variational methods and the prescribed initial prior covariance for DBF and CGKN. All methods are evaluated at a 30\% observation ratio using the same evaluation starts within each data set. The random seed for background perturbation sampling is set to zero. The background and analysis errors are defined as
\begin{equation}
\begin{aligned}
    \mathbf{e}_{\mathrm{b}}^{(i)}
    &\sim
    \mathcal{N}(\mathbf{0},\mathbf{B}),
    &
    \mathbf{z}_{0}^{\mathrm{b},(i)}
    &=
    \mathbf{z}_{0}^{\mathrm{true}}
    +\mathbf{e}_{\mathrm{b}}^{(i)},\\
    \mathbf{e}_{\mathrm{a}}^{(i)}
    &=
    \mathbf{z}_{t_{\mathrm a}}^{\mathrm{a},(i)}
    -\mathbf{z}_{t_{\mathrm a}}^{\mathrm{true}},
    &
    i
    &=1,\ldots,N_{\mathrm{ens}}.
\end{aligned}
\end{equation}
Here, $i$ indexes the ensemble members, and $t_{\mathrm a}$ denotes the analysis evaluation time. For the variational methods, $t_{\mathrm a}=0$, and the analysis is obtained by assimilating the full observation sequence. For DBF and CGKN, the analysis is evaluated after the first observation update, relative to the reference state at the corresponding time. All ensemble members assimilate the same observations. The Gaussian background error ensemble provides the reference for identifying distributional changes introduced by assimilation.

When the latent state dimensionality is large relative to the ensemble size $N_{\mathrm{ens}}$, the full-space sample covariance may be poorly conditioned~\citep{ledoit2004well}, and many conventional multivariate Gaussianity tests are not designed for this high-dimensional setting~\citep{mansoor2017using}. We therefore project the background and analysis errors onto the leading $k$-dimensional eigenspace of $\mathbf{B}$, with $k=25$ for all methods and systems. Compared with testing in the full latent space, this projection requires estimating only a $k$-dimensional covariance matrix, reducing the risk of ill-conditioning in finite-sample settings~\citep{ledoit2004well,mansoor2017using}. The leading eigenspace retains the largest background error variance among all $k$-dimensional linear projections~\citep{jolliffe2016principal} and therefore focuses the diagnostics on the dominant directions of background uncertainty~\citep{lermusiaux1999data}. Because $\mathbf{U}_{k}$ is computed from $\mathbf{B}$ before assimilation and applied to both error ensembles, the projection is independent of the resulting analysis distribution and provides a common basis for comparison. The eigendecomposition of $\mathbf{B}$ and the resulting projected errors are given by
\begin{equation}
    \mathbf{B}\mathbf{U}_{k}
    =
    \mathbf{U}_{k}\boldsymbol{\Lambda}_{k},
    \qquad
    \widetilde{\mathbf{e}}_{\mathrm{b}}^{(i)}
    =
    \mathbf{U}_{k}^{\top}\mathbf{e}_{\mathrm{b}}^{(i)},
    \qquad
    \widetilde{\mathbf{e}}_{\mathrm{a}}^{(i)}
    =
    \mathbf{U}_{k}^{\top}\mathbf{e}_{\mathrm{a}}^{(i)},
\end{equation}
where $\boldsymbol{\Lambda}_{k}$ contains the $k$ largest eigenvalues of $\mathbf{B}$, $\mathbf{U}_{k}=[\mathbf{u}_{1},\ldots,\mathbf{u}_{k}]$ contains the corresponding eigenvectors, and $\widetilde{\mathbf{e}}_{\mathrm{b}}^{(i)}$ and $\widetilde{\mathbf{e}}_{\mathrm{a}}^{(i)}$ denote the projected background and analysis errors, respectively. If Gaussianity is rejected in the projected subspace, the full latent error distribution is also non-Gaussian. If it is not rejected, the result does not establish Gaussianity outside the examined directions.

Within this dominant error subspace, the Henze--Zirkler test assesses joint multivariate Gaussianity~\citep{henze1990class}, while Mardia skewness and excess kurtosis characterize multivariate asymmetry and heavy- or light-tailed behavior, respectively~\citep{mardia1970measures}. Because finite ensemble size can produce nonzero standardized Mardia statistics even for the Gaussian background error ensemble, we use the background values as finite-sample references and report
\begin{equation}
    \Delta s_{\mathrm{skew}}
    =
    s_{\mathrm{skew}}^{\mathrm{a}}
    -
    s_{\mathrm{skew}}^{\mathrm{b}},
    \qquad
    \Delta s_{\mathrm{kurt}}
    =
    s_{\mathrm{kurt}}^{\mathrm{a}}
    -
    s_{\mathrm{kurt}}^{\mathrm{b}},
\end{equation}
where $s_{\mathrm{skew}}^{\mathrm{a}}$ and $s_{\mathrm{kurt}}^{\mathrm{a}}$ are the standardized Mardia statistics of the analysis errors, and $s_{\mathrm{skew}}^{\mathrm{b}}$ and $s_{\mathrm{kurt}}^{\mathrm{b}}$ are their background counterparts.
\begin{figure}[htbp]
    \centering
    \includegraphics[width=\linewidth]{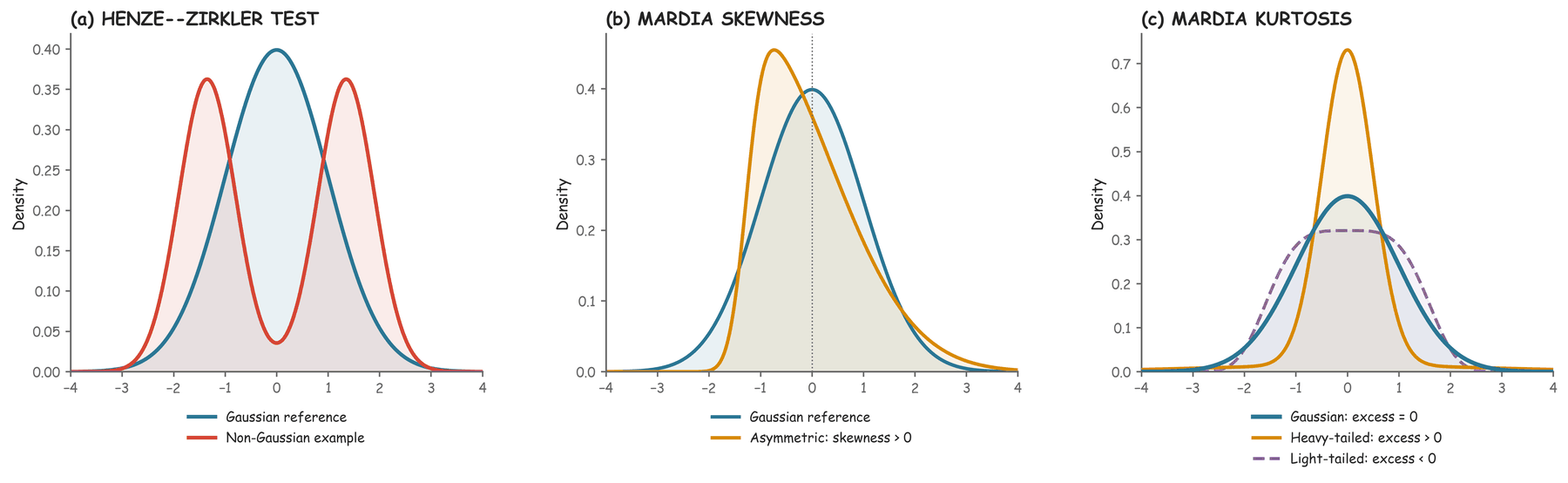}
    \caption{Schematic illustration of the Gaussianity diagnostics. The Henze--Zirkler test assesses joint multivariate Gaussianity, Mardia skewness measures the magnitude of multivariate asymmetry, and Mardia excess kurtosis characterizes heavy- or light-tailed departures from the Gaussian reference.}
    \label{fig:gaussianity-diagnostics}
\end{figure}

\subsection{Hessian-Based Convexity Analysis}
\label{app:convexity-analysis}
\begin{figure}[htbp]
    \centering
    \includegraphics[width=\linewidth]{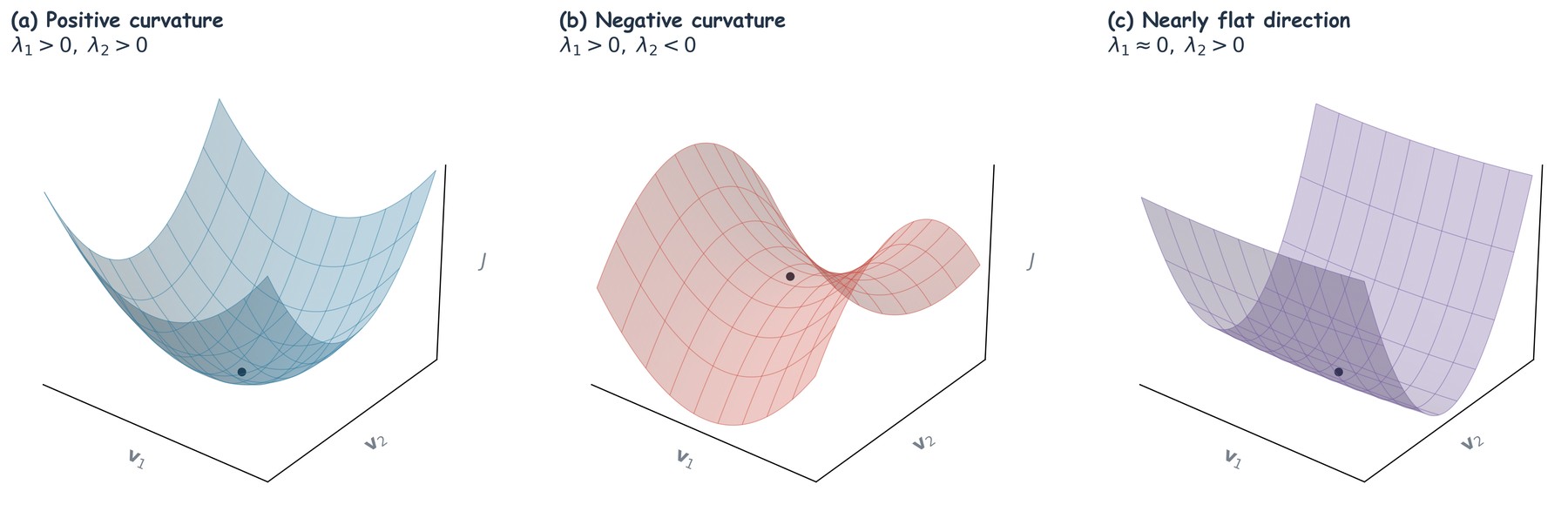}
    \caption{Schematic illustration of the objective curvature along latent-trajectory directions defined by Hessian eigenvectors. Positive eigenvalues indicate positive curvature, negative eigenvalues indicate locally non-convex directions, and approximately zero eigenvalues indicate weakly constrained directions.}
    \label{fig:hessian-curvature}
\end{figure}

Figure~\ref{fig:hessian-curvature} illustrates the interpretation of the Hessian eigenvalues used in Section~\ref{sec:convexity-experiment}. We first separate the curvature contributions of the three terms in the assimilation objective and then explain how the complete Hessian is analyzed for the linear and nonlinear variational models.

Let $\mathbf{z}_{0:T}$ denote the stacked latent trajectory. The complete Hessian $\mathbf{H}_{J}$ of the weak-constraint assimilation objective can be decomposed as
\begin{equation}
    \mathbf{H}_{J}(\mathbf{z}_{0:T})
    =
    \mathbf{H}_{\mathrm{b}}
    +
    \mathbf{H}_{\mathrm{model}}(\mathbf{z}_{0:T})
    +
    \mathbf{H}_{\mathrm{obs}}(\mathbf{z}_{0:T}),
\end{equation}
where $\mathbf{H}_{\mathrm{b}}$, $\mathbf{H}_{\mathrm{model}}$, and $\mathbf{H}_{\mathrm{obs}}$ are the background, model error, and observation error Hessians, respectively. Let $\lambda$ and $\mathbf{v}=[\mathbf{v}_{0}^{\top},\ldots,\mathbf{v}_{T}^{\top}]^{\top}$ denote an eigenvalue and its corresponding eigenvector of $\mathbf{H}_{J}(\mathbf{z}_{0:T})$. Along the latent-trajectory direction defined by $\mathbf{v}$, the background Hessian contributes
\begin{equation}
    \mathbf{H}_{\mathrm{b}}
    =
    \operatorname{diag}\!\left(\mathbf{B}^{-1},\mathbf{0},\ldots,\mathbf{0}\right),
    \qquad
    \mathbf{v}^{\top}\mathbf{H}_{\mathrm{b}}\mathbf{v}
    =
    \mathbf{v}_{0}^{\top}\mathbf{B}^{-1}\mathbf{v}_{0}\geq 0.
\end{equation}
The zero blocks correspond to $\mathbf{z}_{1:T}$ because the background term depends only on $\mathbf{z}_{0}$. Since $\mathbf{B}$ is positive definite, $\mathbf{H}_{\mathrm{b}}$ is positive semidefinite, and its contribution along $\mathbf{v}$ is non-negative. The background Hessian therefore cannot introduce negative curvature.

For DMD and Tensor-Var, the model and observation residuals are linear functions of the latent trajectory, yielding a quadratic objective whose complete Hessian is constant. A displacement along the latent-trajectory direction defined by $\mathbf{v}$ is parameterized as
\begin{equation}
    \mathbf{H}_{J}(\mathbf{z}_{0:T})\mathbf{v}
    =
    \lambda\mathbf{v},
    \qquad
    \mathbf{z}_{0:T}(\alpha)
    =
    \mathbf{z}_{0:T}+\alpha\mathbf{v},
\end{equation}
where $\alpha$ controls the displacement along $\mathbf{v}$. Substituting this trajectory into the objective gives
\begin{equation}
\begin{aligned}
J\!\left(\mathbf{z}_{0:T}(\alpha)\right)
&=
\frac{1}{2}
\left\|
\mathbf{z}_{0}-\mathbf{z}_{0}^{\mathrm{b}}+\alpha\mathbf{v}_{0}
\right\|_{\mathbf{B}^{-1}}^{2}
+
\frac{1}{2}\sum_{t\in\mathcal{T}}
\left\|
\mathbf{z}_{t}-\widehat{\mathbf{z}}_{t}+\alpha\mathbf{v}_{t}
\right\|_{\mathbf{R}_{z}^{-1}}^{2}
\\
&\quad+
\frac{1}{2}\sum_{t=1}^{T}
\left\|
\mathbf{z}_{t}-\mathbf{K}\mathbf{z}_{t-1}
+\alpha\left(\mathbf{v}_{t}-\mathbf{K}\mathbf{v}_{t-1}\right)
\right\|_{\mathbf{Q}_{z}^{-1}}^{2}.
\end{aligned}
\end{equation}
Taking the second derivative with respect to $\alpha$ gives the curvature along $\mathbf{v}$:
\begin{equation}
\begin{aligned}
\frac{\mathrm{d}^{2}}{\mathrm{d}\alpha^{2}}
J\!\left(\mathbf{z}_{0:T}(\alpha)\right)
&=
\mathbf{v}^{\top}\mathbf{H}_{J}(\mathbf{z}_{0:T})\mathbf{v}
=
\lambda\mathbf{v}^{\top}\mathbf{v}
\\
&=
\left\|\mathbf{v}_{0}\right\|_{\mathbf{B}^{-1}}^{2}
+
\sum_{t=1}^{T}
\left\|\mathbf{v}_{t}-\mathbf{K}\mathbf{v}_{t-1}\right\|_{\mathbf{Q}_{z}^{-1}}^{2}
+
\sum_{t\in\mathcal{T}}
\left\|\mathbf{v}_{t}\right\|_{\mathbf{R}_{z}^{-1}}^{2}.
\end{aligned}
\end{equation}
The right-hand side is strictly positive for every nonzero $\mathbf{v}$ when $\mathbf{B}$ and $\mathbf{Q}_z$ are positive definite: if the background and model contributions all vanish, then $\mathbf{v}_0=\mathbf{0}$ and recursively $\mathbf{v}_t=\mathbf{K}\mathbf{v}_{t-1}=\mathbf{0}$. The DMD and Tensor-Var objectives therefore have unique minimizers under these assumptions. Numerically small eigenvalues indicate weakly constrained directions rather than establishing non-uniqueness.

For the decoder-coupled models (NL-AE, WL-AE, and KAE), define the observation mapping $g_t(\mathbf{z}_{0:T})=\mathcal{H}_t\circ\mathcal{D}_x(\mathbf{z}_t)$ and the observation and model residuals
\begin{equation}
\begin{aligned}
\mathbf{r}_t(\mathbf{z}_{0:T})
&=\mathbf{y}_t-g_t(\mathbf{z}_{0:T}),
&
\mathbf{e}_t(\mathbf{z}_{0:T})
&=\mathbf{z}_t-\mathcal{M}_z(\mathbf{z}_{t-1}).
\end{aligned}
\end{equation}
Let $\mathbf{J}_{g_t}$ and $\mathbf{J}_{e_t}$ denote the Jacobians of $g_t$ and $\mathbf{e}_t$, respectively, with respect to the full trajectory $\mathbf{z}_{0:T}$. Differentiating the complete objective $J(\mathbf{z}_{0:T})$ in Eq.~\eqref{eq:weak-4d-var} twice gives the standard Gauss--Newton decomposition~\citep{nocedal2006numerical}:
\begin{equation}
\begin{aligned}
\mathbf{H}_J
&=\mathbf{A}+\mathbf{C}_{\mathrm{model}}-\mathbf{C}_{\mathrm{obs}},
&
\mathbf{A}
&=\mathbf{H}_{\mathrm{b}}
+\sum_{t=1}^{T}\mathbf{J}_{e_t}^{\top}\mathbf{Q}_z^{-1}\mathbf{J}_{e_t}
+\sum_{t\in\mathcal{T}}\mathbf{J}_{g_t}^{\top}\mathbf{R}^{-1}\mathbf{J}_{g_t},
\end{aligned}
\label{eq:full-gauss-newton}
\end{equation}
where the residual-weighted second-derivative terms are
\begin{equation}
\begin{aligned}
\mathbf{C}_{\mathrm{model}}
&=\sum_{t=1}^{T}\sum_{j=1}^{d}
\left[\mathbf{Q}_z^{-1}\mathbf{e}_t\right]_j
\nabla_{\mathbf{z}_{0:T}}^2 e_{t,j},
&
\mathbf{C}_{\mathrm{obs}}
&=\sum_{t\in\mathcal{T}}\sum_{j=1}^{m_t}
\left[\mathbf{R}^{-1}\mathbf{r}_t\right]_j
\nabla_{\mathbf{z}_{0:T}}^2 g_{t,j}.
\end{aligned}
\end{equation}
Here, $d$ and $m_t$ are the latent and observation dimensions, respectively, and all quantities are evaluated at $\mathbf{z}_{0:T}$. For positive-definite $\mathbf{B}$, $\mathbf{Q}_z$, and $\mathbf{R}$, $\mathbf{A}$ is positive definite: the background term constrains the initial state, and the sequential model residuals constrain all subsequent states. Weyl's inequality~\citep{hornjohnson2012} therefore gives
\begin{equation}
\lambda_{\min}(\mathbf{H}_J)
\geq
\lambda_{\min}(\mathbf{A})
-\|\mathbf{C}_{\mathrm{model}}\|_{\mathrm{op}}
-\|\mathbf{C}_{\mathrm{obs}}\|_{\mathrm{op}},
\end{equation}
where $\|\cdot\|_{\mathrm{op}}$ denotes the operator norm. Thus, negative curvature can arise only through the residual-weighted curvature of the model or observation mapping. For KAE, the linear latent transition gives $\mathbf{C}_{\mathrm{model}}=\mathbf{0}$, so any negative curvature must originate in the observation term. For NL-AE and WL-AE, both terms can contribute.

For DMD and Tensor-Var, both the model residual and the latent observation residual $\mathbf{z}_t-\widehat{\mathbf{z}}_t$ are affine, so both second-derivative corrections vanish. With fixed positive-definite covariances, the complete objective is strictly convex and quadratic. Under the assumed Gaussian error model, it defines a Gaussian posterior over the latent trajectory. For a single analysis step, the update reduces to the Kalman form; over the full assimilation window, the solution is equivalently obtained by Kalman smoothing. Moreover, with observations held fixed, the analysis solution is an affine function of the background state, so Gaussian background perturbations yield Gaussian analysis errors, as examined in Section~\ref{sec:gaussianity-experiment}.

For the nonlinear models, evaluating the Hessian at a finite number of trajectories can reveal local non-convexity but cannot establish global convexity. We therefore examine the local curvature near the background trajectories from which assimilation is initialized.

For each evaluation case, the model first generates a background trajectory from $\mathbf{z}_{0}^{\mathrm{b}}$ without assimilation:
\begin{equation}
    \mathbf{z}_{t}^{\mathrm{b}}
    =
    \mathcal{M}_{z}\!\left(\mathbf{z}_{t-1}^{\mathrm{b}}\right),
    \qquad t=1,\ldots,T.
\end{equation}
Perturbed trajectories are then sampled around this background trajectory as
\begin{equation}
    \mathbf{z}_{t}^{(i)}
    =
    \mathbf{z}_{t}^{\mathrm{b}}
    +
    \boldsymbol{\sigma}_{z}^{\mathrm{b}}
    \odot
    \boldsymbol{\epsilon}_{t}^{(i)},
    \qquad
    \boldsymbol{\epsilon}_{t}^{(i)}
    \sim
    \mathcal{N}(\mathbf{0},\mathbf{I}),
\end{equation}
where $\boldsymbol{\sigma}_{z}^{\mathrm{b}}$ denotes the component-wise standard deviation of the background trajectory and $\odot$ denotes element-wise multiplication. This scaling keeps the perturbations consistent with the typical variation of each latent component and evaluates the local curvature near the model-generated trajectory. The complete Hessian spectrum is evaluated at the unperturbed background trajectory to characterize the local curvature at initialization. At each perturbed trajectory, the curvature of the model error and observation error terms is measured by
\begin{equation}
    \lambda_{\min}\!\left(\mathbf{H}\!\left(\mathbf{z}_{0:T}^{(i)}\right)\right),
    \qquad
    \mathbf{H}\in\left\{\mathbf{H}_{\mathrm{model}},\mathbf{H}_{\mathrm{obs}}\right\}.
\end{equation}
To compare objectives with different Hessian scales, the minimum curvature is normalized as
\begin{equation}
    r
    =
    \frac{\lambda_{\min}}{|\lambda_{\max}|},
    \qquad
    s
    =
    \frac{r}{|r|+1}.
    \label{eq:normalized-min-curvature}
\end{equation}
This transformation preserves the sign of the minimum eigenvalue while mapping its normalized magnitude to a bounded range.

Although $\mathbf{H}_{J}$, $\mathbf{H}_{\mathrm{model}}$, and $\mathbf{H}_{\mathrm{obs}}$ are theoretically symmetric, automatic differentiation and floating-point computation may produce slightly asymmetric numerical estimates. We therefore symmetrize all three Hessians before eigen-decomposition, preserving their curvature information while improving the stability of the eigenvalue calculation. For the reported experiments, the dimension- and precision-adjusted zero band is defined as
\begin{equation}
    \tau_{H}
    =
    N\epsilon_{\mathrm{dtype}}
    \max_{i}|\lambda_{i}|,
    \label{eq:hessian-zero-band}
\end{equation}
where $N$ is the Hessian dimension and $\epsilon_{\mathrm{dtype}}$ is the machine precision of the computation. Eigenvalues below $-\tau_{H}$ are classified as negative, those satisfying $|\lambda_{i}|\leq\tau_{H}$ as approximately zero, and those above $\tau_{H}$ as positive.

\section{Supplementary Results}
\label{app:supplementary-results}

This appendix presents additional results that complement the main experiments.

\subsection{Koopman Latent-Dimension Sensitivity}
\label{app:koopman-latent-dimension}

Although the Lorenz--96 physical state has a dimension of 40, the neural models use a latent dimension of 120. For Koopman models, the latent dimension is not determined solely by the physical state dimension. The latent representation must also provide learned observables whose evolution can be approximated by a single linear operator ~\citep{Brunton_2016,Lusch2018,otto2019linearly}. To examine the effect of the latent dimension, we train KAE with latent dimensions of 20, 40, 80, and 120 using the same setting and evaluate each variant over 400 trajectory windows.

\begin{figure}[htbp]
    \centering
    \includegraphics[width=0.5\linewidth]
    {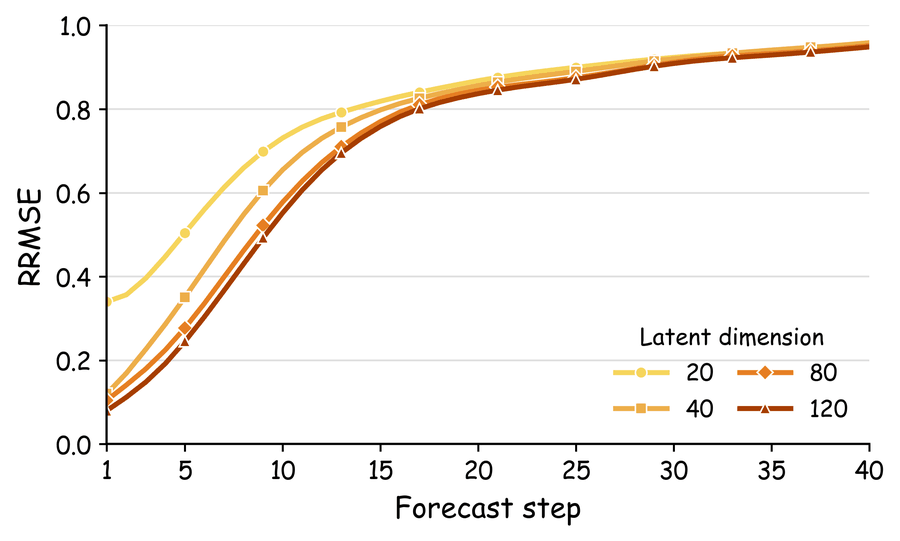}
    \caption{KAE prediction RRMSE on Lorenz--96 for latent dimensions of
    20, 40, 80, and 120.}
    \label{fig:koopman-latent-dimension}
\end{figure}

Figure~\ref{fig:koopman-latent-dimension} shows that the forecast error decreases as the latent dimension increases. The improvement continues beyond the 40-dimensional physical state, indicating that the additional latent dimensions provide learned observables that improve the finite-dimensional Koopman approximation. The 120-dimensional representation consequently produces the most stable predictions, supporting its use in the main benchmark. The smaller improvement from 80 to 120 dimensions indicates that the benefit of increasing the latent dimension begins to saturate. Nevertheless, all configurations eventually approach a similar high-error range. A higher-dimensional latent representation therefore delays but does not eliminate long-horizon forecast degradation.

\subsection{Additional Physical-Field and Spectral Comparisons}
\label{app:additional-fields-spectra}

This subsection extends the comparisons in Section~\ref{sec:physical-fields-spectral-analysis} to Lorenz--96, Cylinder, shallow water, and ERA5--Coarse, showing representative forecast and analysis states at a 30\% observation ratio. For Cylinder, shallow water, and ERA5--Coarse, absolute errors relative to the true physical state are also shown. Within each field comparison, the fields share a common color scale, and the absolute errors share a separate scale. Cylinder and shallow-water fields are shown in standardized units; the ERA5--Coarse field shows specific humidity at 600~hPa in g/kg.

The PSD comparisons show how forecast and analysis variability is distributed across spatial scales. The free-forecast spectra, obtained without incorporating observations, are compared with the analysis spectra at observation ratios of 5\%, 15\%, 30\%, and 50\%. The spectra are computed from standardized fields and averaged over evaluation runs and time steps, rather than obtained from the individual snapshots shown here.

\begin{figure*}[htbp]
    \centering
    \includegraphics[width=\textwidth]{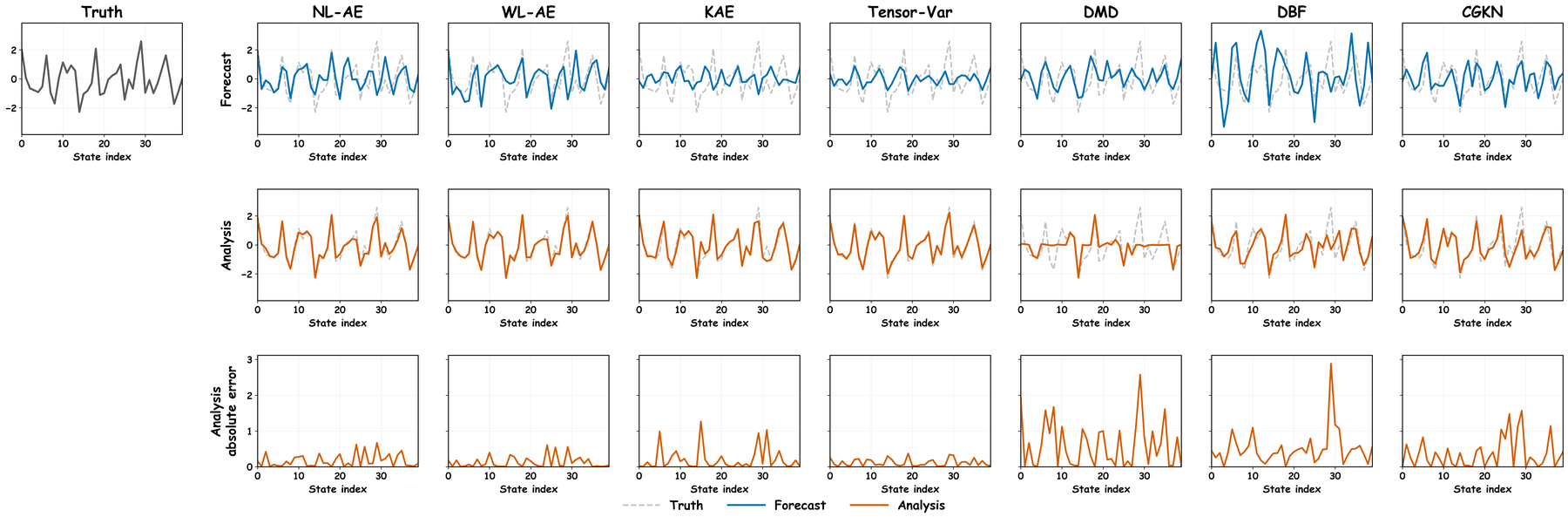}
    \caption{Lorenz--96 forecast and analysis states at a 30\% observation ratio, selected as a case whose RRMSE is closest to the test-set median. The dashed gray curves denote the true physical state.}
    \label{fig:lorenz96-fields}
\end{figure*}

\begin{figure}[htbp]
    \centering
    \includegraphics[width=\textwidth]{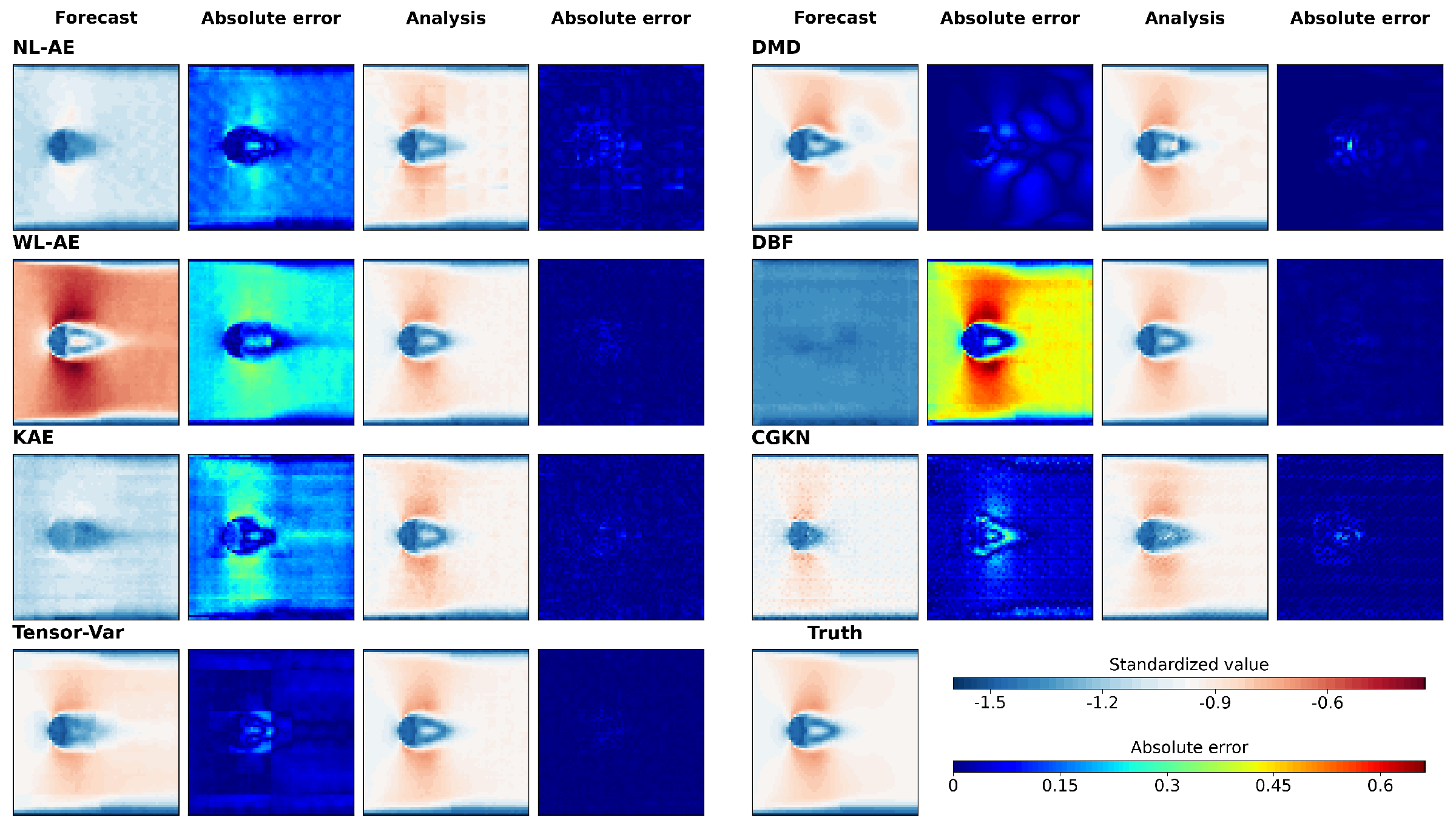}
    \caption{Representative Cylinder forecast and analysis fields at a 30\% observation ratio, with absolute errors relative to the true physical state. Field values and errors are shown in standardized units.}
    \label{fig:app-cylinder-fields}
\end{figure}

\begin{figure}[htbp]
    \centering
    \includegraphics[width=\textwidth]{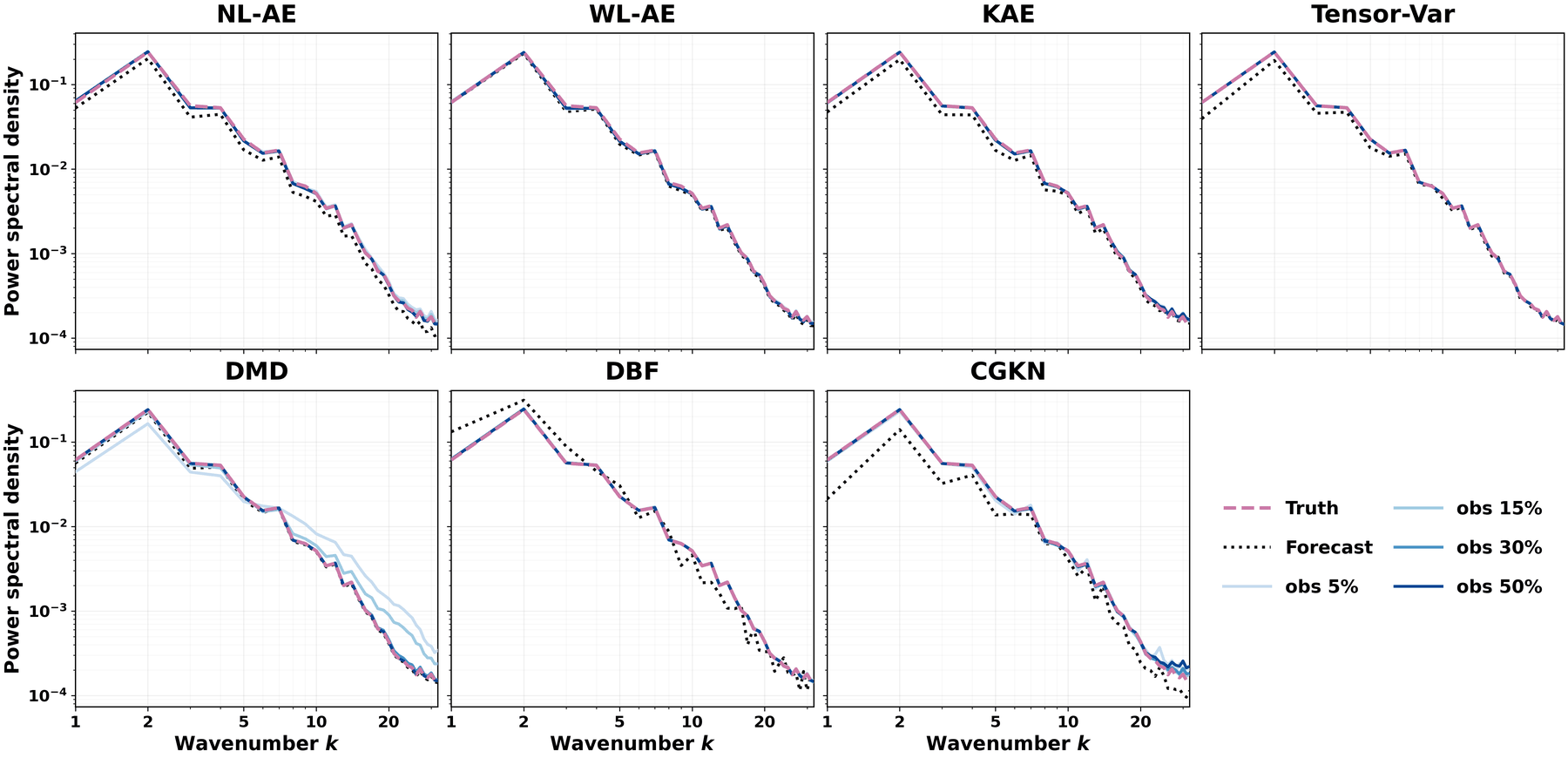}
    \caption{Cylinder power spectral density across observation ratios. The true-state and forecast spectra provide reference curves for the analysis spectra. Lower and higher wavenumbers correspond to larger and finer spatial scales, respectively.}
    \label{fig:app-cylinder-psd}
\end{figure}

\begin{figure}[htbp]
    \centering
    \includegraphics[width=\textwidth,
    height=0.65\textheight]{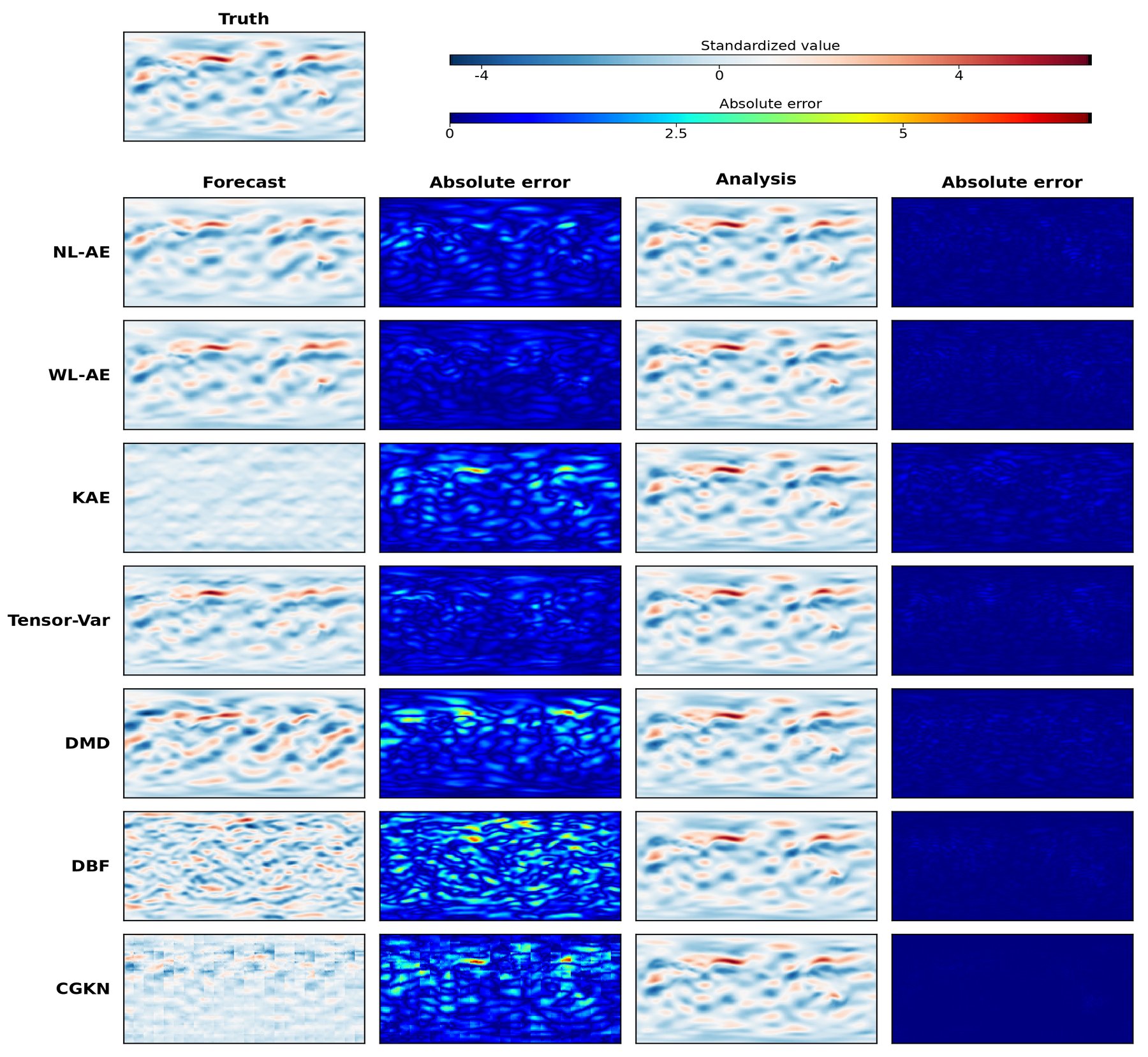}
    \caption{Representative shallow-water forecast and analysis fields at a 30\% observation ratio, with absolute errors relative to the true physical state. Field values and errors are shown in standardized units.}
    \label{fig:app-shallow-water-fields}
\end{figure}

\begin{figure}[htbp]
    \centering
    \includegraphics[width=\textwidth]{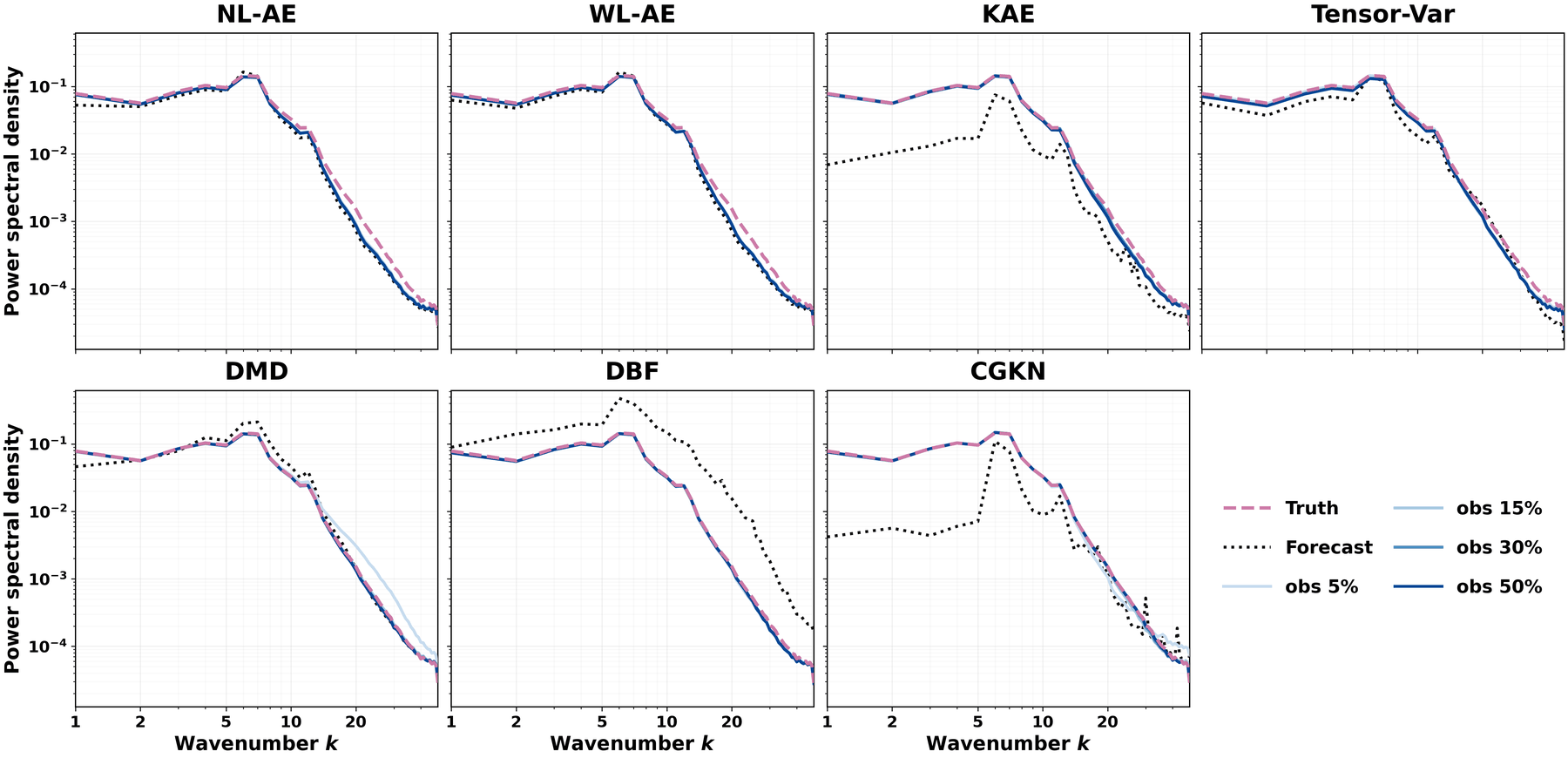}
    \caption{Shallow-water power spectral density across observation ratios, comparing the analysis spectra with the true-state and forecast spectra.}
    \label{fig:app-shallow-water-psd}
\end{figure}

\begin{figure}[htbp]
    \centering
    \includegraphics[width=\textwidth,
    height=0.65\textheight]{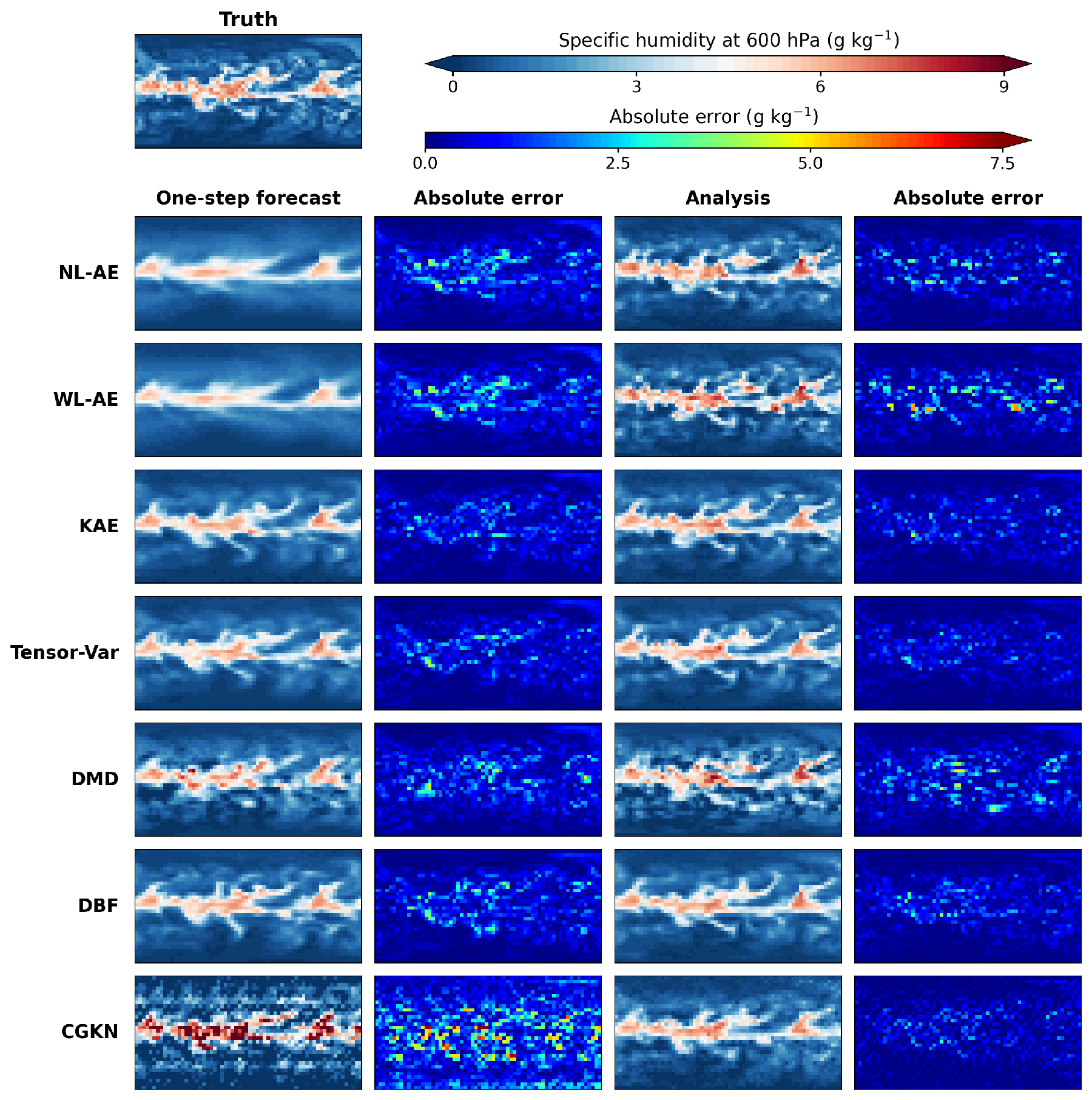}
    \caption{Representative ERA5--Coarse specific humidity fields at 600~hPa and a 30\% observation ratio. Forecast and analysis fields are shown together with their absolute errors relative to the true physical state, all in g/kg.}
    \label{fig:app-era5-fields}
\end{figure}

\begin{figure}[htbp]
    \centering
    \includegraphics[width=\textwidth]{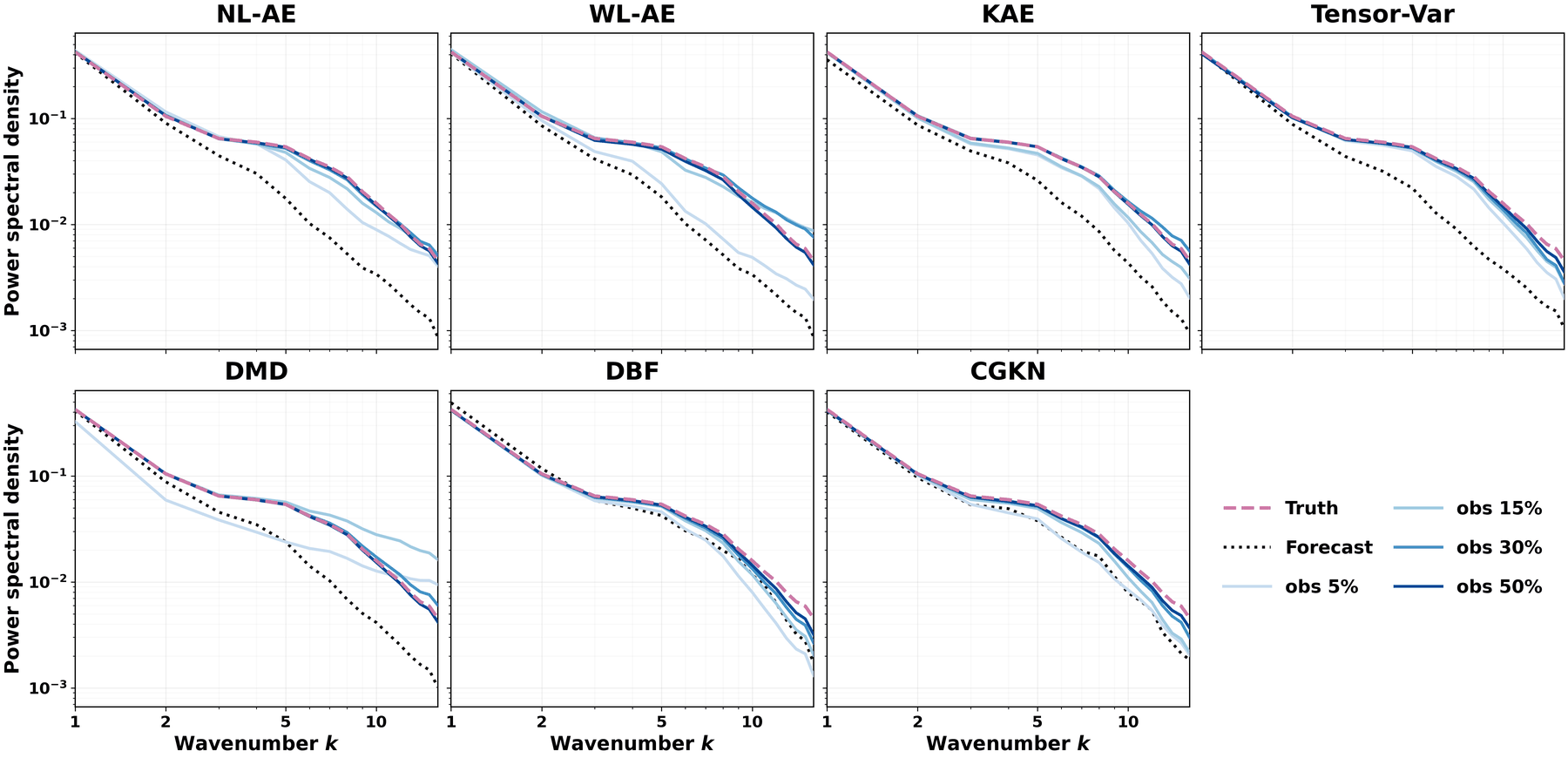}
    \caption{ERA5--Coarse power spectral density across observation ratios, comparing the analysis spectra with the true-state and forecast spectra.}
    \label{fig:app-era5-psd}
\end{figure}

The analysis spectra for Cylinder and shallow water generally follow the true-state spectra across the resolved wavenumber range, including the higher wavenumbers associated with fine-scale variations. ERA5--Coarse shows larger discrepancies at high wavenumbers, particularly under sparse observations, but its recovery of high-wavenumber energy is substantially better than that of ERA5--High. As shown in Section~\ref{sec:physical-fields-spectral-analysis}, substantial high-wavenumber discrepancies remain for ERA5--High even as the observation ratio increases. This contrast further supports the finding that recovering fine-scale variations remains challenging in high-dimensional systems with spatial variations across a wide range of scales.

\subsection{CGKN with Full Observations}
\label{app:cgkn-full-observations}

\begin{figure}[htbp]
    \centering
    \includegraphics[width=\textwidth]{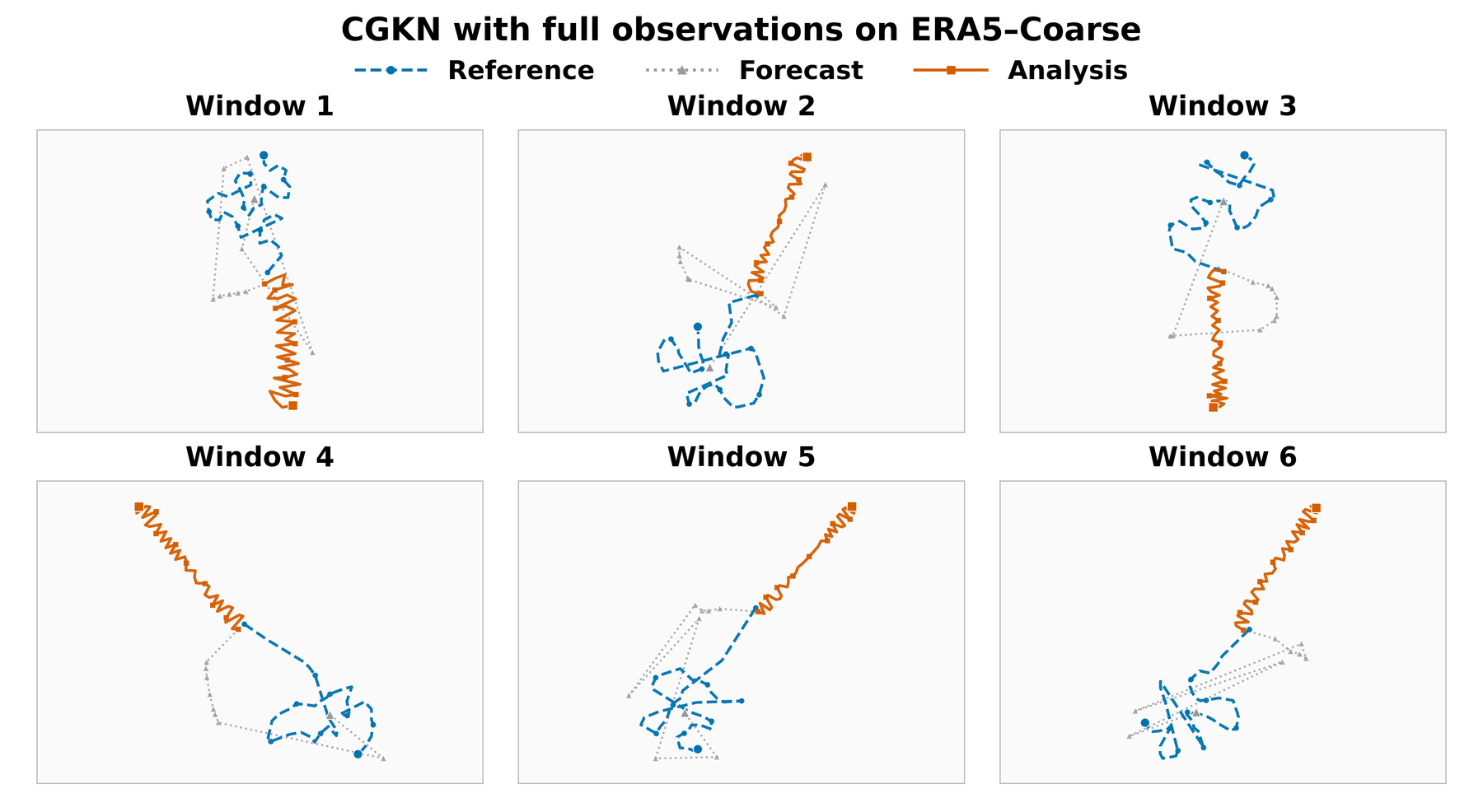}
    \caption{Two-dimensional t-SNE embeddings of the reference, free-forecast, and analysis latent trajectories for CGKN with full observations on ERA5--Coarse. Each panel shows one test-sequence window, with the three trajectories embedded jointly within that window.}
    \label{fig:cgkn-full-obs-trajectories}
\end{figure}

To examine whether limited observational information contributes to the complex encoded reference-trajectory patterns discussed in Section~\ref{sec:latent-trajectory-visualization}, we evaluate a separately trained CGKN model with full observations on ERA5--Coarse. Figure~\ref{fig:cgkn-full-obs-trajectories} shows six trajectories from the test sequence. The encoded reference trajectories exhibit more regular patterns, while the analysis trajectories exhibit patterns similar to those under sparse observations.

With full observations, the coupled dynamics receive complete observed state information at each observation time. This reduces the missing information in the conditioning state and may allow the learned latent dynamics to be more consistently constrained, contributing to the less complex patterns in the encoded reference trajectories. The analysis trajectories retain similar patterns because their evolution remains constrained by the recursive forecast--analysis cycle, even with full observations.

\subsection{Effect of Observation History}
\label{sec:dmd-tv-history-ablation}

\begin{figure}[htbp]
    \centering
    \includegraphics[width=\linewidth]{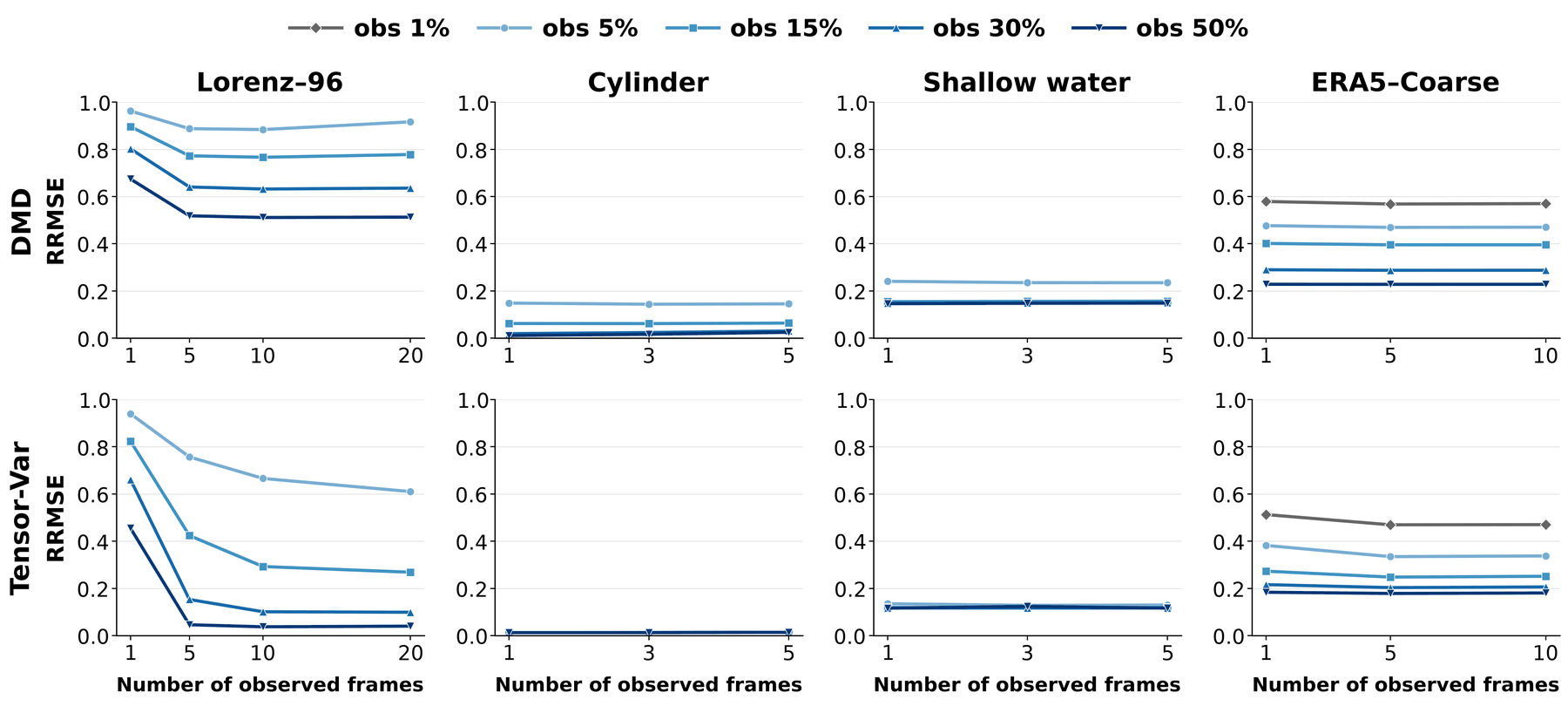}
    \caption{Effect of observation history on assimilation accuracy. Columns correspond to data sets, and rows correspond to DMD and Tensor-Var. Colors indicate observation ratios. The horizontal axis reports the number of observed frames, including the current observation.}
    \label{fig:observation-history}
\end{figure}

To examine the contribution of observation history, we evaluate DMD and Tensor-Var with different numbers of observed frames on Lorenz--96, Cylinder, shallow water, and ERA5--Coarse. The observation sequences include the current observation and the available past observations. Tensor-Var maps these sequences to a latent target through its inverse-observation encoder $\mathcal{E}_{y}$. For DMD, each observation is first mapped to a latent estimate $\widetilde{\mathbf{z}}_{t_j}$ using the regularized inverse projection defined in Appendix~\ref{app:dmd}. These estimates are then combined to construct the current observation-derived latent target:
\begin{equation}
    \widehat{\mathbf{z}}_{t}^{\mathrm{hist}}
    =
    \sum_{j=0}^{n_h-1}
    \mathbf{A}_{j}\widetilde{\mathbf{z}}_{t_j},
    \qquad
    \mathbf{A}_{j}
    =
    \operatorname{diag}(a_{1,j},\ldots,a_{r,j}),
    \label{eq:dmd-history-target}
\end{equation}
where $n_h$ is the number of observed frames, including the current observation, $t_0=t$, and $t_j$ for $j>0$ denotes a past observation time. Here, $r$ is the reduced dimension, and the diagonal weights $\mathbf{A}_{j}$ are fitted independently for each latent component by ridge regression on the training data and held fixed during assimilation. The resulting target replaces the original single-frame target in the variational objective. Figure~\ref{fig:observation-history} compares the resulting assimilation errors across observation ratios.

Observation history substantially improves assimilation accuracy on Lorenz--96. This improvement is consistent with delay-coordinate embedding: under suitable conditions, a sufficiently long sequence of observations provides information for reconstructing states on the attractor~\citep{takens2006detecting}. From a data assimilation perspective, observation history also provides additional temporal information to compensate for representation error. Under this interpretation, historical observations are expected to be more beneficial when the encoder--decoder provides less accurate physical state reconstruction. On ERA5--Coarse, the greater improvements at lower observation ratios further show that observation history provides more useful supplementary information when current observations are sparse.

\subsection{Performance Across Training Seeds}
\label{app:seed-variability}

This subsection reports the performance of RRMSE and SSIM across training seeds. For each seed, scores are averaged over evaluation trajectories. We then report the mean and sample standard deviation across seeds. Lower RRMSE and higher SSIM indicate better performance.

Tables~\ref{tab:seed-da-lorenz96}--\ref{tab:seed-da-era5-high} report performance at the four observation ratios.
\begin{table}[htbp]
\centering
\small
\setlength{\tabcolsep}{4pt}
\begin{tabular}{@{}llcccc@{}}
\toprule
Model & Metric & 5\% & 15\% & 30\% & 50\% \\
\midrule
NL-AE & RRMSE & $0.5916\,\pm\,0.0228$ & $0.3345\,\pm\,0.0405$ & $0.2393\,\pm\,0.0273$ & $0.1873\,\pm\,0.0166$ \\
 & SSIM & $0.4259\,\pm\,0.0335$ & $0.6378\,\pm\,0.0459$ & $0.7542\,\pm\,0.0333$ & $0.8225\,\pm\,0.0206$ \\
\addlinespace[2pt]
WL-AE & RRMSE & $0.6041\,\pm\,0.0271$ & $0.3429\,\pm\,0.0176$ & $0.2348\,\pm\,0.0199$ & $0.1855\,\pm\,0.0100$ \\
 & SSIM & $0.4192\,\pm\,0.0456$ & $0.6380\,\pm\,0.0357$ & $0.7561\,\pm\,0.0252$ & $0.8217\,\pm\,0.0159$ \\
\addlinespace[2pt]
KAE & RRMSE & $0.7955\,\pm\,0.0117$ & $0.5050\,\pm\,0.0127$ & $0.2626\,\pm\,0.0056$ & $0.1190\,\pm\,0.0014$ \\
 & SSIM & $0.2243\,\pm\,0.0080$ & $0.5020\,\pm\,0.0129$ & $0.7421\,\pm\,0.0029$ & $0.9085\,\pm\,0.0021$ \\
\addlinespace[2pt]
Tensor-Var & RRMSE & $0.5746\,\pm\,0.0054$ & $0.2200\,\pm\,0.0017$ & $0.1477\,\pm\,0.0008$ & $0.1398\,\pm\,0.0017$ \\
 & SSIM & $0.4194\,\pm\,0.0082$ & $0.8152\,\pm\,0.0027$ & $0.8948\,\pm\,0.0049$ & $0.9068\,\pm\,0.0025$ \\
\addlinespace[2pt]
DMD & RRMSE & $0.9692$ & $0.9134$ & $0.8224$ & $0.6903$ \\
 & SSIM & $0.0161$ & $0.0513$ & $0.1275$ & $0.2628$ \\
\addlinespace[2pt]
DBF & RRMSE & $0.7753\,\pm\,0.0116$ & $0.6051\,\pm\,0.0025$ & $0.4537\,\pm\,0.0015$ & $0.3896\,\pm\,0.0045$ \\
 & SSIM & $0.2679\,\pm\,0.0154$ & $0.4008\,\pm\,0.0079$ & $0.5282\,\pm\,0.0074$ & $0.6013\,\pm\,0.0038$ \\
\addlinespace[2pt]
CGKN & RRMSE & $0.9615\,\pm\,0.0552$ & $0.8849\,\pm\,0.0508$ & $0.4570\,\pm\,0.0022$ & $0.1327\,\pm\,0.0112$ \\
 & SSIM & $0.0929\,\pm\,0.0022$ & $0.1543\,\pm\,0.0165$ & $0.5016\,\pm\,0.0152$ & $0.8921\,\pm\,0.0118$ \\
\bottomrule
\end{tabular}
\caption{Data assimilation performance across training seeds on Lorenz--96. Entries are mean $\pm$ sample standard deviation; DMD is reported as a single score.}
\label{tab:seed-da-lorenz96}
\end{table}

\begin{table}[htbp]
\centering
\small
\setlength{\tabcolsep}{4pt}
\begin{tabular}{@{}llcccc@{}}
\toprule
Model & Metric & 5\% & 15\% & 30\% & 50\% \\
\midrule
NL-AE & RRMSE & $0.1790\,\pm\,0.0514$ & $0.1644\,\pm\,0.0528$ & $0.1606\,\pm\,0.0570$ & $0.1595\,\pm\,0.0577$ \\
 & SSIM & $0.8185\,\pm\,0.0680$ & $0.8263\,\pm\,0.0717$ & $0.8329\,\pm\,0.0767$ & $0.8352\,\pm\,0.0767$ \\
\addlinespace[2pt]
WL-AE & RRMSE & $0.1049\,\pm\,0.0013$ & $0.0818\,\pm\,0.0017$ & $0.0764\,\pm\,0.0029$ & $0.0756\,\pm\,0.0024$ \\
 & SSIM & $0.9122\,\pm\,0.0026$ & $0.9352\,\pm\,0.0019$ & $0.9405\,\pm\,0.0025$ & $0.9415\,\pm\,0.0023$ \\
\addlinespace[2pt]
KAE & RRMSE & $0.0347\,\pm\,0.0151$ & $0.0311\,\pm\,0.0125$ & $0.0288\,\pm\,0.0138$ & $0.0282\,\pm\,0.0134$ \\
 & SSIM & $0.9658\,\pm\,0.0284$ & $0.9713\,\pm\,0.0218$ & $0.9717\,\pm\,0.0240$ & $0.9725\,\pm\,0.0230$ \\
\addlinespace[2pt]
Tensor-Var & RRMSE & $0.0148\,\pm\,0.0000$ & $0.0152\,\pm\,0.0000$ & $0.0152\,\pm\,0.0001$ & $0.0151\,\pm\,0.0001$ \\
 & SSIM & $0.9936\,\pm\,0.0000$ & $0.9934\,\pm\,0.0000$ & $0.9933\,\pm\,0.0000$ & $0.9934\,\pm\,0.0000$ \\
\addlinespace[2pt]
DMD & RRMSE & $0.2990$ & $0.1407$ & $0.0289$ & $0.0064$ \\
 & SSIM & $0.6740$ & $0.8730$ & $0.9786$ & $0.9981$ \\
\addlinespace[2pt]
DBF & RRMSE & $0.0825\,\pm\,0.0145$ & $0.0572\,\pm\,0.0019$ & $0.0478\,\pm\,0.0061$ & $0.0447\,\pm\,0.0021$ \\
 & SSIM & $0.9591\,\pm\,0.0040$ & $0.9688\,\pm\,0.0016$ & $0.9721\,\pm\,0.0010$ & $0.9728\,\pm\,0.0029$ \\
\addlinespace[2pt]
CGKN & RRMSE & $0.1020\,\pm\,0.0034$ & $0.0708\,\pm\,0.0025$ & $0.0695\,\pm\,0.0087$ & $0.0650\,\pm\,0.0065$ \\
 & SSIM & $0.8952\,\pm\,0.0034$ & $0.9207\,\pm\,0.0196$ & $0.9079\,\pm\,0.0424$ & $0.9074\,\pm\,0.0333$ \\
\bottomrule
\end{tabular}
\caption{Data assimilation performance across training seeds on Cylinder. Entries are mean $\pm$ sample standard deviation; DMD is reported as a single score.}
\label{tab:seed-da-cylinder}
\end{table}

\begin{table}[htbp]
\centering
\small
\setlength{\tabcolsep}{4pt}
\begin{tabular}{@{}llcccc@{}}
\toprule
Model & Metric & 5\% & 15\% & 30\% & 50\% \\
\midrule
NL-AE & RRMSE & $0.2763\,\pm\,0.0013$ & $0.2599\,\pm\,0.0039$ & $0.2548\,\pm\,0.0056$ & $0.2528\,\pm\,0.0041$ \\
 & SSIM & $0.7677\,\pm\,0.0024$ & $0.7811\,\pm\,0.0045$ & $0.7870\,\pm\,0.0068$ & $0.7883\,\pm\,0.0045$ \\
\addlinespace[2pt]
WL-AE & RRMSE & $0.2948\,\pm\,0.0102$ & $0.2804\,\pm\,0.0092$ & $0.2789\,\pm\,0.0098$ & $0.2793\,\pm\,0.0098$ \\
 & SSIM & $0.7540\,\pm\,0.0099$ & $0.7654\,\pm\,0.0093$ & $0.7664\,\pm\,0.0089$ & $0.7657\,\pm\,0.0094$ \\
\addlinespace[2pt]
KAE & RRMSE & $0.1285\,\pm\,0.0027$ & $0.1076\,\pm\,0.0026$ & $0.1037\,\pm\,0.0042$ & $0.1235\,\pm\,0.0114$ \\
 & SSIM & $0.9288\,\pm\,0.0036$ & $0.9470\,\pm\,0.0022$ & $0.9499\,\pm\,0.0026$ & $0.9318\,\pm\,0.0105$ \\
\addlinespace[2pt]
Tensor-Var & RRMSE & $0.1296\,\pm\,0.0002$ & $0.1194\,\pm\,0.0002$ & $0.1175\,\pm\,0.0002$ & $0.1152\,\pm\,0.0000$ \\
 & SSIM & $0.9315\,\pm\,0.0002$ & $0.9410\,\pm\,0.0001$ & $0.9423\,\pm\,0.0002$ & $0.9444\,\pm\,0.0000$ \\
\addlinespace[2pt]
DMD & RRMSE & $0.2776$ & $0.1553$ & $0.1479$ & $0.1458$ \\
 & SSIM & $0.7909$ & $0.8785$ & $0.8844$ & $0.8863$ \\
\addlinespace[2pt]
DBF & RRMSE & $0.1982\,\pm\,0.0032$ & $0.1892\,\pm\,0.0169$ & $0.1708\,\pm\,0.0064$ & $0.1643\,\pm\,0.0113$ \\
 & SSIM & $0.8509\,\pm\,0.0045$ & $0.8567\,\pm\,0.0159$ & $0.8759\,\pm\,0.0044$ & $0.8809\,\pm\,0.0082$ \\
\addlinespace[2pt]
CGKN & RRMSE & $0.3056\,\pm\,0.0062$ & $0.2882\,\pm\,0.0229$ & $0.2513\,\pm\,0.0161$ & $0.2284\,\pm\,0.0185$ \\
 & SSIM & $0.7613\,\pm\,0.0046$ & $0.7804\,\pm\,0.0123$ & $0.7927\,\pm\,0.0062$ & $0.8010\,\pm\,0.0120$ \\
\bottomrule
\end{tabular}
\caption{Data assimilation performance across training seeds on Shallow water. Entries are mean $\pm$ sample standard deviation; DMD is reported as a single score.}
\label{tab:seed-da-shallow-water}
\end{table}

\begin{table}[htbp]
\centering
\small
\setlength{\tabcolsep}{4pt}
\begin{tabular}{@{}llcccc@{}}
\toprule
Model & Metric & 5\% & 15\% & 30\% & 50\% \\
\midrule
NL-AE & RRMSE & $0.5291\,\pm\,0.0019$ & $0.4241\,\pm\,0.0013$ & $0.3000\,\pm\,0.0023$ & $0.2231\,\pm\,0.0016$ \\
 & SSIM & $0.6096\,\pm\,0.0016$ & $0.7335\,\pm\,0.0015$ & $0.8405\,\pm\,0.0023$ & $0.8923\,\pm\,0.0018$ \\
\addlinespace[2pt]
WL-AE & RRMSE & $0.5345\,\pm\,0.0064$ & $0.4909\,\pm\,0.0100$ & $0.3799\,\pm\,0.0035$ & $0.2409\,\pm\,0.0026$ \\
 & SSIM & $0.5782\,\pm\,0.0037$ & $0.6631\,\pm\,0.0104$ & $0.8052\,\pm\,0.0049$ & $0.8811\,\pm\,0.0025$ \\
\addlinespace[2pt]
KAE & RRMSE & $0.4122\,\pm\,0.0025$ & $0.3174\,\pm\,0.0132$ & $0.2569\,\pm\,0.0115$ & $0.2101\,\pm\,0.0125$ \\
 & SSIM & $0.7227\,\pm\,0.0014$ & $0.8202\,\pm\,0.0128$ & $0.8715\,\pm\,0.0110$ & $0.9050\,\pm\,0.0110$ \\
\addlinespace[2pt]
Tensor-Var & RRMSE & $0.3384\,\pm\,0.0002$ & $0.2589\,\pm\,0.0007$ & $0.2143\,\pm\,0.0005$ & $0.1882\,\pm\,0.0007$ \\
 & SSIM & $0.7978\,\pm\,0.0003$ & $0.8691\,\pm\,0.0004$ & $0.9031\,\pm\,0.0004$ & $0.9196\,\pm\,0.0019$ \\
\addlinespace[2pt]
DMD & RRMSE & $0.6573$ & $0.6576$ & $0.3400$ & $0.2359$ \\
 & SSIM & $0.4847$ & $0.6341$ & $0.8232$ & $0.8864$ \\
\addlinespace[2pt]
DBF & RRMSE & $0.3854\,\pm\,0.0008$ & $0.3034\,\pm\,0.0002$ & $0.2507\,\pm\,0.0008$ & $0.2104\,\pm\,0.0010$ \\
 & SSIM & $0.7431\,\pm\,0.0024$ & $0.8262\,\pm\,0.0004$ & $0.8741\,\pm\,0.0010$ & $0.9075\,\pm\,0.0013$ \\
\addlinespace[2pt]
CGKN & RRMSE & $0.3912\,\pm\,0.0005$ & $0.2941\,\pm\,0.0009$ & $0.2177\,\pm\,0.0008$ & $0.1950\,\pm\,0.0010$ \\
 & SSIM & $0.7259\,\pm\,0.0007$ & $0.8293\,\pm\,0.0003$ & $0.9020\,\pm\,0.0013$ & $0.9228\,\pm\,0.0015$ \\
\bottomrule
\end{tabular}
\caption{Data assimilation performance across training seeds on ERA5--Coarse. Entries are mean $\pm$ sample standard deviation; DMD is reported as a single score.}
\label{tab:seed-da-era5}
\end{table}

\begin{table}[htbp]
\centering
\small
\setlength{\tabcolsep}{4pt}
\begin{tabular}{@{}llcccc@{}}
\toprule
Model & Metric & 5\% & 15\% & 30\% & 50\% \\
\midrule
NL-AE & RRMSE & $0.4336\,\pm\,0.0154$ & $0.2369\,\pm\,0.0027$ & $0.1609\,\pm\,0.0001$ & $0.1412\,\pm\,0.0006$ \\
 & SSIM & $0.6991\,\pm\,0.0109$ & $0.8564\,\pm\,0.0006$ & $0.8984\,\pm\,0.0004$ & $0.9106\,\pm\,0.0015$ \\
\addlinespace[2pt]
WL-AE & RRMSE & $0.4116\,\pm\,0.0254$ & $0.2775\,\pm\,0.0242$ & $0.2295\,\pm\,0.0258$ & $0.2156\,\pm\,0.0268$ \\
 & SSIM & $0.6160\,\pm\,0.0110$ & $0.7439\,\pm\,0.0318$ & $0.7862\,\pm\,0.0330$ & $0.8003\,\pm\,0.0346$ \\
\addlinespace[2pt]
KAE & RRMSE & $0.4389\,\pm\,0.0088$ & $0.3491\,\pm\,0.0180$ & $0.3557\,\pm\,0.0518$ & $0.3960\,\pm\,0.0691$ \\
 & SSIM & $0.5574\,\pm\,0.0232$ & $0.6584\,\pm\,0.0039$ & $0.6724\,\pm\,0.0091$ & $0.5953\,\pm\,0.0650$ \\
\addlinespace[2pt]
Tensor-Var & RRMSE & $0.1968\,\pm\,0.0008$ & $0.1554\,\pm\,0.0004$ & $0.1386\,\pm\,0.0005$ & $0.1315\,\pm\,0.0003$ \\
 & SSIM & $0.8636\,\pm\,0.0010$ & $0.9027\,\pm\,0.0008$ & $0.9163\,\pm\,0.0006$ & $0.9212\,\pm\,0.0003$ \\
\addlinespace[2pt]
DMD & RRMSE & $0.6043$ & $0.2797$ & $0.2274$ & $0.2116$ \\
 & SSIM & $0.5202$ & $0.7468$ & $0.7962$ & $0.8126$ \\
\addlinespace[2pt]
DBF & RRMSE & $0.2211\,\pm\,0.0008$ & $0.1690\,\pm\,0.0043$ & $0.1441\,\pm\,0.0041$ & $0.1325\,\pm\,0.0034$ \\
 & SSIM & $0.8393\,\pm\,0.0018$ & $0.8949\,\pm\,0.0057$ & $0.9161\,\pm\,0.0046$ & $0.9245\,\pm\,0.0034$ \\
\addlinespace[2pt]
CGKN & RRMSE & $0.3873\,\pm\,0.0134$ & $0.2294\,\pm\,0.0098$ & $0.1784\,\pm\,0.0072$ & $0.1564\,\pm\,0.0026$ \\
 & SSIM & $0.4768\,\pm\,0.0148$ & $0.6921\,\pm\,0.0171$ & $0.7593\,\pm\,0.0121$ & $0.8017\,\pm\,0.0049$ \\
\bottomrule
\end{tabular}
\caption{Data assimilation performance across training seeds on ERA5--High. Entries are mean $\pm$ sample standard deviation; DMD is reported as a single score.}
\label{tab:seed-da-era5-high}
\end{table}

\clearpage

\bibliography{main}

\end{document}